\documentclass[nohyperref]{article}

\usepackage{microtype}
\usepackage{graphicx}
\usepackage{subfigure}
\usepackage{booktabs} 

\usepackage{hyperref}
\usepackage[hyphenbreaks]{breakurl}
\usepackage[algoruled,linesnumbered,algo2e]{algorithm2e}

\usepackage[accepted]{icml2023}

\usepackage{mathtools}
\usepackage{amsthm}
\usepackage{calc,amsfonts,amssymb,amsmath,bm,url,color,cite,epstopdf,nicefrac,stmaryrd}
\usepackage{psfrag,float,epstopdf,bbold,mathrsfs}
\usepackage{multirow,comment}
\usepackage{algorithmic,booktabs}
\usepackage[normalem]{ulem}
\usepackage{mdframed}
\usepackage{footnote}
\makesavenoteenv{tabular}
\usepackage{xcolor}
\definecolor{orange}{RGB}{255,107,0}

\usepackage{threeparttable}

\usepackage{tikz}
\usetikzlibrary{bayesnet}
\usetikzlibrary{arrows}
\usepackage{color}
\usepackage{graphicx}

\usepackage[capitalize,noabbrev]{cleveref}

\theoremstyle{plain}
\newtheorem{theorem}{Theorem}[section]

\newtheorem{lemma}[theorem]{Lemma}

\newtheorem{fact}[theorem]{Fact}
\theoremstyle{definition}
\newtheorem{definition}[theorem]{Definition}
\newtheorem{assumption}[theorem]{Assumption}
\theoremstyle{remark}

\usepackage[textsize=tiny]{todonotes}

\icmltitlerunning{Under-Counted Tensor Completion with Neural Incorporation of Attributes}

\newcommand{\W}{\boldsymbol{W}}

\newcommand{\X}{\boldsymbol{X}}

\newcommand{\U}{\boldsymbol{U}}

\newcommand{\x}{\boldsymbol{x}}
\newcommand{\f}{\boldsymbol{f}}
\newcommand{\z}{\boldsymbol{z}}

\renewcommand{\v}{\boldsymbol{v}}
\renewcommand{\u}{\boldsymbol{u}}
\newcommand{\w}{\boldsymbol{w}}

\renewcommand{\b}{\boldsymbol{b}}

\newcommand{\T}{{\!\top\!}}

\newcommand{\bOmega}{\bm{\varOmega}}
\newcommand{\bTheta}{\bm{\varTheta}}

\newcommand{\bXi}{\bm{\varXi}}

\newcommand{\bi}{\bm{i}}
\newcommand{\bZ}{\bm{Z}}

\newcommand{\tX}{\underline{\bm X}}
\newcommand{\tY}{\underline{\bm Y}}

\newcommand{\tP}{\underline{\bm P}}
\newcommand{\tM}{\underline{\bm M}}

\newcommand{\tlambda}{\underline{\bm \Lambda}}

\DeclareMathOperator*{\minimize}{\textrm{minimize}}

 \newtheorem{Theorem}{Theorem}

\definecolor{orange}{RGB}{255,107,0}

\begin{document}

\twocolumn[
\icmltitle{Under-Counted Tensor Completion with Neural Incorporation of Attributes}



\icmlsetsymbol{equal}{*}

\begin{icmlauthorlist}
\icmlauthor{Shahana Ibrahim}{yyy}
\icmlauthor{Xiao Fu}{yyy}
\icmlauthor{Rebecca Hutchinson}{yyy}
\icmlauthor{Eugene Seo}{comp}
\end{icmlauthorlist}

\icmlaffiliation{yyy}{School of Electrical Engineering and Computer Science, Oregon State University, Corvallis, OR 97331, USA}
\icmlaffiliation{comp}{Department of Applied Mathematics, Brown University, Providence, RI 02912, USA}

\icmlcorrespondingauthor{Xiao Fu}{xiao.fu@oregonstate.edu}

\icmlkeywords{Machine Learning, ICML}

\vskip 0.3in
]



\printAffiliationsAndNotice{}  

\begin{abstract}
Systematic under-counting effects are observed in data collected across many disciplines, e.g., epidemiology and ecology. {\it Under-counted tensor completion} (UC-TC) is well-motivated for many data analytics tasks, e.g., inferring the case numbers of infectious diseases at unobserved locations from under-counted case numbers in neighboring regions.  
{However, existing methods for similar problems often lack supports in theory, making it hard to understand the underlying principles and conditions beyond empirical successes.}
{In this work,} a low-rank Poisson tensor model with an expressive unknown {\it nonlinear} side information extractor is proposed for  under-counted multi-aspect data.
A joint low-rank tensor completion and neural network learning algorithm is designed to {recover} the model.
Moreover, the UC-TC formulation is supported by theoretical analysis showing that the fully counted entries of the tensor and each entry's under-counting probability can be provably recovered from partial observations{---under reasonable conditions}. To our best knowledge, the result {is the} first to offer theoretical supports for under-counted multi-aspect data completion. Simulations and real-data experiments corroborate the theoretical claims.
\end{abstract}

\section{Introduction}\label{sec:introduction}
{\it Tensor completion} (TC) has been a workhorse for a large variety of data analytics tasks, e.g., image/video inpainting \citep{liu2012tensor}, hyperspectral denoising {\citep{wang2018hyperspectral}}, sampling and recovery in magnetic resonance imaging (MRI) \citep{kanatsoulis2019regular}, and {multimodal data mining  \citep{papalexakis2016tensor}}---just to name a few.
In the past decade, theory and methods of TC progressed significantly---aspects such as recoverability, sample complexity, optimal algorithm design were extensively studied under various tensor models, e.g., {\it Tucker decomposition} \citep{mu2014square}, {\it canonical polyadic decomposition} (CPD) \citep{kanatsoulis2019regular,sorensen2019fiber}, {and {\it block-term decomposition} (BTD) \citep{zhang2020spectrum,ding2020hyperspectral}.}
The vast majority of classic TC methods were proposed for handling real-valued data, but TC algorithms designed for integer data, e.g., binary data \citep{ghadermarzy2018learning} and count data \citep{chi2012tensors}, also exist.
{This is because} such data often arise in real-world problems, e.g., movie recommendation \citep{karatzoglou2010multiverse}, {knowledge graph completion  \citep{ivana2019tucker}}, 
and social network analysis \citep{kashima2009link}.

However, less attention has been paid to integer data TC problems where the observed entries suffer from
systematical {\it under-counting}---but (strong) under-counting effects are encountered in many disciplines,
e.g., epidemiology and ecology. 
In epidemiology, the cases of an infectious disease (e.g., COVID-19) may be under-counted due to the existence of symptom-free patients and the lack of testing \citep{eric2020failing}. In ecology, when an observer sees a species at a certain location, the observer  likely just observes a small portion of this species' {population} \citep{MacKenzie2006,Tylianakis2010}.
{Hence,}
taking under-counting effects into consideration for TC is meaningful in these domains. For example, recovering the ``true'' (fully counted) number of COVID-19 cases at a certain location and time helps estimate/understand the infection situation. 
The task of {\it under-counted tensor completion} (UC-TC) is naturally more challenging than the conventional (integer data) TC, as it implicitly involves an extra task of compensating the {\it unknown} under-counting effect.

\subsection{Prior Work and Challenges}
In the literature, there exist some limited efforts towards dealing with under-counted data completion/prediction. For example, the line of work for user exposure matrix completion considered link prediction under undetected links and unobserved links simultaneously \citep{liang2016modeling}.
The work on ``zero-inflated'' matrix/tensor factorization and related models \citep{billio2017bayesian,simchowitz2013zero,durif2019probabilistic} considered a similar scenario, where the datasets' observed entries are zeros with a certain non-zero probability.
The recent work by \citep{fu2019link} proposed an approach to tackle under-counted matrix completion (UC-MC) using a Poisson-Binomial matrix factorization model
---which was inspired by the species' abundance modeling in ecology \citep{joseph2009modeling,royle2004n,hutchinson2011incorporating}.

\paragraph{Challenges.}
The existing works by  \citep{liang2016modeling,billio2017bayesian,simchowitz2013zero,durif2019probabilistic,fu2019link} were shown effective on many UC-MC tasks. With proper modifications, these computational frameworks may be generalized to handle tensor data. However, some notable challenges remain.
{First, how to effectively model the under-counting effect has been an open question. 
For example, the works by \citep{liang2016modeling,billio2017bayesian,simchowitz2013zero,durif2019probabilistic} all considered zero-inflated MC models, which dealt with a special case of under-counting, i.e., the observed matrix entries are either the true counts or zeros. 
For more general under-counted data (i.e., the observed counts are smaller than or equal to the true counts),} the recent work by \citep{fu2019link} proposed to model the probability of missing a count (e.g., an interaction of a pollinator and a plant) as a {\it linear function} of side attributes, e.g.,
the location, temperature, and humidity{---which is reminiscent of the classic species distribution models (SDMs) \citep{hutchinson2011incorporating}.}
However, using simple linear functions is a compromise {between} model expressiveness and {computational convenience}.
Second, the existing works demonstrated their effectiveness only {\it empirically}. 
It has been unclear under what conditions the underlying true counts and some key model parameters (e.g., the detection probabilities of the counts) could be {\it provably} recovered.

\subsection{Contributions}
In this work, we propose a UC-TC framework based on the Poisson-Binomial model \citep{joseph2009modeling,royle2004n,hutchinson2011incorporating} as in the MC work by \citep{fu2019link}.
Unlike \citep{fu2019link} that used a simple side information model and was largely empirical, we address a number of key challenges in modeling, computation, and theory. Our contribution is twofold:

\noindent
{\bf UC-TC with Nonlinear Incorporation of Side Information.} 
To model and learn the under-counting effect in tensor data, we propose a UC-TC framework with neural network-based incorporation of attributes.
Specifically, we extend the UC-MC idea in \citep{fu2019link} to the higher-order tensor regime and model the underlying fully counted data as a Poisson tensor that admits a low-rank {\it canonical polyadic decomposition} (CPD) \citep{harshman1970foundations1};
the Poisson tensor is then under-counted via a Binomial detection procedure.
Different from the existing work that uses a linear function to model the relationship between the attributes and the detection probability of a count,
we represent the relationship as an {\it unknown nonlinear function}, which substantially generalizes the modeling capacity.
We recast the UC-TC problem as a {\it maximum likelihood estimation} (MLE) {criterion}, where the unknown nonlinear function is represented by a neural network (NN) for side information incorporation. We also design a \textit{block coordinate descent} (BCD) algorithm \citep{razaviyayn2013unified} {to tackle the formulated} MLE effectively.

\noindent
{\bf Recoverability Analysis of UC-TC.} We present theoretical characterizations of the proposed UC-TC criterion under the Poisson-Binomial framework.
We show that, under reasonable conditions, the fully counted tensor is recoverable, and the associated detection probabilities of all the entries are identifiable---both are up to a global scaling/counter-scaling ambiguity. 
Our analysis reveals interesting performance-deciding factors such as the similarity of the under-counting effects {across some entries}.
The result also shows that the nonlinear incorporation of side information imposes an intuitive trade-off between model complexity and recovery accuracy, given a fixed amount of samples---which also justifies our proposal of using neural nonlinear models.
To our best knowledge, this is the first provable learning criterion for under-counted factor analysis models.

\smallskip

We conduct extensive evaluation of the proposed approach on synthetic data to validate our theoretical findings. We also test our approach on real-world prediction tasks in epidemiology \citep{zhang2020interactive} and ecology \citep{Jones2017,Daly1957}.

\section{Background}
\subsection{UC-TC: Problem Statement}
A $K$th-order tensor $\tY\in\mathbb{R}^{I_1\times \ldots \times I_K}$ is a multi-way array whose entries are indexed by $K$ coordinates.
An entry of such a tensor can be expressed as follows: 
\begin{equation}\label{eq:tensor_y}
    y_{\bm i} = [\tY]_{\bm i},~\bm i=(i_1,\ldots,i_K)  \in [I_1]\times \ldots \times [I_K].
\end{equation}
Tensors are widely used to represent multi-aspect data.
For example, 
in a four-aspect epidemic data,
$y_{\bi}$ {can represent the number of} recorded infected cases in city $i_1$, during week $i_2$, among age group $i_3$ and ethnicity group $i_4$.

TC and MC tasks naturally arise in many applications, where multi-aspect data are only observed partially. A classic use case is recommender systems \citep{Hu2008, karatzoglou2010multiverse}, where the ratings that users provide to products are used to predict the unseen ratings---which can be done via completing the rating matrix/tensor.
TC problems were primarily considered for continuous data {\citep{gandy2011tensor,montanari2016spectral,yuan2016tensor,zhang2017exact,sorensen2019fiber,zhang2020spectrum}}, but TC for other data formats (e.g., binary and integer data) were also studied in the literature \citep{ghadermarzy2018learning,chi2012tensors,lee2020tensor}---as the latter found many applications in real data analytics problems, e.g., {5-star rating}-based recommender systems \citep{karatzoglou2010multiverse}, adjacency network analysis \citep{acar2009link} and traffic flow prediction \citep{tan2016shortterm}.

In this work, our interest lies in integer TC problems where the partially observed tensor data suffer from systematical under-counting. 
To be specific, we consider integer tensor data as in { \citep{chi2012tensors,lee2020tensor}}.
Like in conventional integer TC problems, the observed tensor $\tY$ is highly incomplete; i.e.,
$\bm \varOmega\subseteq [I_1]\times \ldots \times [I_K]$ are the indices of the observed entries,
where $|\bm \varOmega|\ll \prod_{k=1}^K I_k$.  
The {key difference} from conventional integer TC is that in our setting, the observed entries are smaller than or equal to their actual values, i.e.,
$y_{\bm i} \leq n_{\bm i},~{\bm i}\in\bm \varOmega$,
where $n_{\bm i}=[\underline{\bm N}]_{\bm i}$, $\underline{\bm N}\in\mathbb{Z}_{+}^{I_1\times \ldots \times I_K}$ represents the (unobserved) true-count tensor, and $y_{\bm i}\in\mathbb{Z}_{+}$ is the observed and under-counted version of $n_{\bm i}$.  

We assume that for each entry of $\tY$, there is an associated attribute/feature vector $\bm z_{\bm i}\in\mathbb{R}^D$, capturing side information regarding the observation. The attributes reflect the {\it detection probability} of each count in $n_{\bm i}$. We note that such entry attributes are often available in real-world applications. Two examples are as follows:

{$\bullet$ \bf Epidemiology Data}: Consider tensor data in epidemiology that records infected case counts.  The count tensor could be indexed by `city'$\times$`age group'$\times$`profession'. Then, $\bm z_{\bm i}$ may contain observation-affecting features such as the testing capacity of city $i_1$, population of age group $i_2$, and the level of exposure to the virus for profession $i_3$. 
    
     {$\bullet$ \bf Ecology Data}: 
In ecology, the counts of interactions between pollinators and plants at different times could be represented by a tensor (i.e., a `pollinator'$\times$`plant'$\times$`time' tensor). There, $\bm z_{\bm i}$ could contain the properties of the pollinator $i_1$ and plant $i_2$, the temperature, humidity and other observation-affecting factors of site $i_3$.

 In this work, we assume that the entry features $\z_{
\bi}$ are available for a subset of entries indexed by
$
\bm \varXi\subseteq [I_1]\times \ldots \times [I_K].   
$ {Note that $\bm \varXi$ and $\bm \varOmega$ need not be identical.}

\subsection{Existing Work: The Poisson-Binomial MC Model}
The work in \citep{fu2019link} proposed an MC framework with the following generative model for the under-counted entries of a pollinator-plant interaction count matrix:
\begin{subequations} \label{eq:poisson_n_mixture}
\begin{align} 
	n_{i,j} &\sim {\sf Poisson}(\lambda_{i,j}),~\lambda_{i,j}=\bm u_i^\T\bm v_j,~\u_i,\bm v_j\geq\bm 0,\\
	y_{i,j} &\sim {\sf Binomial}(n_{i,j},p_{i,j}),~p_{i,j}=\bm z_{i,j}^\T \bm \theta, \label{eq:linear_feature}
	\end{align}
\end{subequations} 
where $\bm u_i\in\mathbb{R}^F, \bm v_j\in\mathbb{R}^F$ and $\bm \theta \in \mathbb{R}^D$ are the model parameters to estimate. To be more specific, $\u_i$ and $\v_j$ are the $F$-dimensional latent representations of pollinator $i$ and plant $j$. The parameter $\lambda_{i,j}$ stands for the average number of interactions between pollinator $i$ and plant $j$, and the ground-truth number of interactions $n_{i,j}$ is the realization of a Poisson sampling process with parameter $\lambda_{i,j}$. The observed data $y_{i,j}$ is under-counted through a {Binomial} detection process, with the detection probability $p_{i,j}$. The detection probability $p_{i,j}$ was modeled as a linear function of the entry features $\bm z_{i,j}$.
Under this model, $\bm \Lambda=\bm U\bm V^\T$ where $[\bm \Lambda]_{i,j}=\lambda_{i,j}$ is a low-rank matrix model if $F$ is relatively small.
Hence, estimating $\bm u_i$'s, $\bm v_j$'s, and $\bm \theta$ can be regarded as a variant of Poisson matrix completion problem. 

\subsection{Challenges}
To extend the Poisson-Binomial model to tensor cases and more general settings,
there are a couple of key challenges to be addressed.
First, the linear model in \eqref{eq:linear_feature}, i.e., $p_{i,j}=\bm z_{i,j}^\T\bm \theta$, may not be expressive enough, as the relation between the features and the detection probability is highly likely to be {\it nonlinear}. The linear model assumption was used for computational convenience, but is a performance-limiting factor in the model of \citep{fu2019link} (as will be seen in the {experiments}).
Second, perhaps more importantly, there {were} no theoretical characterizations of the Poisson-Binomial model, despite of its popularity in ecological data analysis \citep{fu2019link,hutchinson2011incorporating,royle2004n,dennis2015computational}. In the context of UC-MC, it is unclear if the underlying Poisson parameters $\lambda_{i,j}$ for all $(i,j)$ could be recovered from a subset of observations $y_{i,j}, ~(i,j)\in \bm \varOmega$. 
The recoverability analysis can not be covered by existing TC theory, as the ``extra'' task of estimating $p_{i,j}$ {makes UC-TC} a much harder learning problem compared to the traditional TC problems.

\section{Problem Formulation}
In this work, we generalize the Poisson-Binomial model to cover tensor data, with {\it nonilinear} attribute incorporation---and more importantly---with recoverability support.
\subsection{Generative Model}
We generalize the model \eqref{eq:poisson_n_mixture} by considering under-counted tensors with a size of $I_1\times\cdots\times I_K$ generated as follows:
\begin{subequations}\label{eq:nonlinear_model_tensor}
	\begin{align} 
	\lambda_{\bm i} & = \sum_{f=1}^F  \prod_{k=1}^K \bm U_k(i_k,f),~\bm U_k \ge \bm 0, \forall k\in [K], \label{eq:lambda_model}\\
	n_{\bm i} &\sim {\sf Poisson}(\lambda_{\bm i}), \label{eq:poisson}\\
	p_{\bm i} &= g\left(\bm z_{\bm i}\right),~0\le p_{\bm i}\le 1,\label{eq:p_model}\\
	y_{\bm i}&\sim {\sf Binomial}(n_{\bm i},p_{\bm i}), \label{eq:binomial}
	\end{align}
\end{subequations}
where ${\bm i}=(i_1,\dots,i_K)$ is a shorthand notation as {defined} before, $\bm U_k(i_k,:){ \in\mathbb{R}^F}$ denotes the $F$-dimensional latent representation of entity $i_k$ of aspect $k$, $y_{\bm i}$ denotes the {\it observed} count indexed by $\bm i$,  $n_{\bm i}$ is the corresponding true count, $\lambda_{\bm i}$ {stands for} the Poisson parameter (which is the expectation of the true count), and  $p_{\bm i}$ {represents} the detection probability as in the model of \eqref{eq:poisson_n_mixture}. 
The vector $\bm z_{\bm i} \in \mathbb{R}^{D}$ collects the features of entry $\bm i$. 
The {\it unknown} nonlinear function $g(\cdot): \mathbb{R}^{D} \rightarrow [0,1]$ is used to model the dependence between the detection probability $p_{\bm i}$ and $\bm z_{\bm i}$. In \eqref{eq:nonlinear_model_tensor}, we have imposed nonnegativity constraints on $\bm U_k$'s in order to make sure that the Poisson parameters are all nonnegative.  

Given $	y_{\bm i}$ for $\bi \in \bm \varOmega \subset [I_1]\times \ldots \times [I_K]$ and $	\bm z_{\bm i}$ for $\bi \in \bm \varXi \subset [I_1]\times \ldots \times [I_K]$, 
our goal is to {estimate $\lambda_{\bi}$ and $p_{\bi}$ for all $\bi $.}

\subsection{Maximum Likelihood Estimation-Based UC-TC}
Assuming that $y_{\bi}$'s are independently sampled from the generative model in \eqref{eq:nonlinear_model_tensor}, the log-likelihood of the observations can be expressed as follows:
\begin{align}
&\log \prod_{\bm i\in \bm \varOmega}{\sf Pr}(y_{\bm i}; \lambda_{\bm i}, p_{\bm i}) \nonumber\\
&= \log\left(\prod_{\bi\in\bm \varOmega} \sum_{n=y_{\bi}}^\infty {\sf Pr}(N_{\bi}=n;\lambda_{\bi}) {\sf Pr}(y_{\bi}|N_{\bi}=n; p_{\bi})\right) \nonumber\\
&= \sum_{\bi\in\bm \varOmega} \log\left( \sum_{n=y_{\bi}}^\infty\left(\frac{\lambda_{\bi}^{n} e^{-\lambda_{\bi}}}{n!} \frac{n! p_{\bi}^{y_{\bi}} (1-p_{\bi})^{n-y_{\bi}}}{y_{\bi}!(n-y_{\bi})!} \right)\right) \nonumber\\
&= \sum_{\bm i\in \bm \varOmega} \left[y_{\bm i}\log \lambda_{\bm i}+y_{\bm i}\log p_{\bm i}-\lambda_{\bm i}p_{\bm i}-\log y_{\bm i}!\right]; \label{eq:log_likelihood}
\end{align}
see the detailed derivation in the supplementary material in Sec. \ref{app:mle}. 
Under the generative model in \eqref{eq:nonlinear_model_tensor}, we have $p_{\bi}=g(\z_{\bi})$. Since $g(\cdot)$ is unknown and nonlinear, we represent it using a neural network denoted as $g_{\bm \theta} :\mathbb{R}^D \rightarrow [0,1]$:
\begin{align} \label{eq:fcnn}
   g_{\bm \theta}(\bm z) =  \sigma( \bm w_L^\T\bm \sigma(\bm W_{L-1} \bm \sigma (\ldots \bm \sigma(\bm W_1\bm z) ))),
\end{align}
where $\bm W_\ell \in \mathbb{R}^{d_{\ell} \times d_{\ell-1}}$ denotes the weight matrix at $\ell$th layer ($d_0=D$), $\bm w_{L} \in \mathbb{R}^{d_{L-1}}$ denotes the last layer's weights,  $\bm \theta$ denotes the parameters of the NN $(\{ \W_\ell \}_{\ell=1}^{L-1},\w_L),$
and $\bm \sigma(\cdot)=[\sigma(\cdot),\ldots,\sigma(\cdot)]^\T$ denotes the activation function {with a proper dimension}. 
Hence, combining \eqref{eq:log_likelihood} with \eqref{eq:lambda_model} and \eqref{eq:p_model},
we formulate the MLE criterion as follows:
\begin{subequations} \label{eq:max_likelihood_optim_tensor}
	\begin{align}
	\underset{\{\bm U_k\}, \bm \theta,  \{p_{\bm i}\}}{\rm minimize}&~L(\bm U,\bm \theta, \tP) := \sum_{\bm i \in \bm \varOmega}\Bigg[\bigg( \sum_{f=1}^F  \prod_{k=1}^K \bm U_k(i_k,f)\bigg)p_{\bm i}\Big.\nonumber \\
  	-y_{\bm i}&\log \bigg( \sum_{f=1}^F  \prod_{k=1}^K \bm U_k(i_k,f)\bigg)  \Big.-y_{\bm i}\log p_{\bm i}\Bigg], \label{eq:objective_fn}\\
	{\rm subject\ to}~~
 & {\bm U_k \in {\cal U}_k, ~\forall k \in [K],\label{eq:nonnegative_emb}}\\\
	& p_{\bi} = g_{\bm \theta}(\z_{\bi}), ~ \forall \bm i \in \bm \varXi,~ {g_{\bm \theta} \in {\cal G}},\label{eq:p_constraint}
	\end{align}
\end{subequations}
{where ${\cal U}_k$ is a constraint set that encodes the prior knowledge of $\bm U_k$ (e.g., nonnegativity),
and ${\cal G}$ represents the function class where $g_{\bm \theta}$ is taken from.}
Given the formulated MLE, there are a couple of key aspects that are of interest. First, if one solves the formulated learning criterion in \eqref{eq:max_likelihood_optim_tensor}, can the {\it optimal} solution recover the latent parameters $\lambda_{\bi}$ and $p_{\bm i}$ over all $\bm i$? 
This question will be answered in Sec.~\ref{sec:recovery}.
Second, the criterion \eqref{eq:max_likelihood_optim_tensor} presents a challenging optimization problem that involves tensor decomposition and neural network learning---both are nontrivial problems. 
A BCD algorithm to effectively tackle this problem will be proposed in Sec.~\ref{sec:algorithm}.

\section{Recoverability Analysis}\label{sec:recovery}
Our goal is to show that the {optimal solution} given by \eqref{eq:max_likelihood_optim_tensor} provably recovers $\lambda_{\bi}$'s and $p_{\bi}$'s under the generative model \eqref{eq:nonlinear_model_tensor} (up to certain reasonable ambiguities).

\subsection{Analysis Setup}
To better present the analysis, we use ``$^\natural$'' to denote the ground-truth parameters in the model \eqref{eq:nonlinear_model_tensor}---e.g., $\lambda_{\bi}^\natural$ and $p_{\bi}^\natural$ denote the ground-truth Poisson parameter and ground-truth detection probability of entry $\bm i$, respectively. 
We have the corresponding tensor notations $[\tlambda^\natural]_{\bi} = \lambda^\natural_{\bi}$ and $[\tP^\natural]_{\bi} = p^\natural_{\bi}$. 
The ``hat'' notation is used to denote the terms constructed from the optimal solution given by \eqref{eq:max_likelihood_optim_tensor}.
For instance,
 $\{\widehat{\bm U}_k\}_{k=1}^K$, $\widehat{g}_{\bm \theta}$ and $\widehat{p}_{\bi}, \forall \bi$ denote the estimates given by the solution obtained from solving \eqref{eq:max_likelihood_optim_tensor}.
We consider the following assumptions:
\begin{assumption} [\textbf{Bounded Parameters}] \label{as:bound}
There exist scalars  $p_{\min}, p_{\max}, \beta_u, \alpha_u > 0$ such that
$
p^\natural_{\bi} = g^\natural(\z_{\bi}),~0<p_{\min} \leq p^\natural_{\bi} \leq p_{\max} ,~\forall \bi$ and $0<\beta_u\leq \U^\natural_k(i_k,f)\leq \alpha_u,~\forall i_k \in [I_k], f \in [F], k \in [K]$.
{In addition, the constraints in \eqref{eq:max_likelihood_optim_tensor}, namely, ${\cal U}_k$ and ${\cal G}$ {satisfy} ${\cal U}_k = \{\bm U \in \mathbb{R}^{I_k \times F}~|~ \beta_u \le \bm U(i_k,f) \le \alpha_u, ~\forall i_k,f\}$ and 
 $ 
{\cal G} = \{g_{\bm \theta}: \mathbb{R}^D \rightarrow [p_{\min},p_{\max}]\}$, respectively.}
\end{assumption}

\begin{assumption}[\textbf{Approximation Error}] \label{as:approx}
 Assume that there exists $\widetilde{g}_{\bm \theta}\in {\cal G}$ such that $|\widetilde{g}_{\bm \theta}(\z) -g^\natural(\bm z)| \le \nu $ for all $\z \in \mathbb{R}^D$, where $0\leq \nu <\infty$. 
 In addition, assume that the class ${\cal G}$ has {a complexity measure $\mathscr{R}_{\cal{G}}$}. 
\end{assumption}
{
\begin{assumption} [\textbf{Similar Attribute Subset}]\label{as:similar_z}
   There {exists} an index set $\bm \varTheta$ whose elements are sampled uniformly at random (without replacement) from $[I_1]\times \ldots \times [I_K]$ and a scalar $\zeta>0$ such that
   $
      \max_{\bi,\bm j\in \bm \varTheta} \| \bm z_{\bi} -\bm z_{\bm j}\|_2 \leq \zeta.
    $
\end{assumption}
}
\begin{assumption}[\textbf{Lipschitz Continuity}]\label{as:countinuous_g}
Assume that the {ground-truth} function $g^\natural$ and any function ${g}_{\bm \theta}\in {\cal G}$ are Lipschitz continuous, i.e., for certain $L_g, L_{\bm \theta}>0$ and for any pair of $\bm z_{\bi}$, $\bm z_{\bm j}$, $|g^\natural(\bm z_{\bi}) - g^\natural({\bm z}_{\bm j}) | \leq L_g \|\bm z_{\bi} -\bm z_{\bm j} \|_2, \label{eq:lip_g_nat}$ and $|{g}_{\bm \theta}(\bm z_{\bi}) - {g}_{\bm \theta}({\bm z}_{\bm j}) | \leq L_{\bm \theta} \|\bm z_{\bi} -\bm z_{\bm j} \|_2$ hold.
\end{assumption}
Assumption~\ref{as:approx} requires that the neural network class ${\cal G}$ contains a function $\widetilde{g}_{\bm \theta}$ that is closer to the ground-truth function $g^\natural$. {In practice}, if deeper/wider neural networks {are employed}, the value {of} $\nu$ {is} smaller. {For the complexity measure $\mathscr{R}_{\cal{G}}$ used in Assumption~\ref{as:approx}, we use the {\it sensitive complexity} parameter introduced in \citep{lin2019generalization}. For deeper and wider neural network function classes, the parameter $\mathscr{R}_{\cal{G}}$ gets larger---also see supplementary material Sec. \ref{app:complexity_RG} for more details.}
Assumptions~\ref{as:similar_z}-\ref{as:countinuous_g} together imply that there {exists} a set of $p_{\bi}$'s {that} are similar. 
For example, in an epidemic dataset (e.g., {the} COVID-19 {dataset} \citep{zhang2020interactive}) {that} records the number of detected infectious disease cases, when the testing capacity and the population of a number of locations are similar with each other, it is reasonable to assume that the corresponding detection probabilities $p_{\bi}$'s are similar. 
The Lipschitz continuity assumption on $g_{\bm \theta}$ given by {Assumption~\ref{as:countinuous_g} requires that the learned ${g}_{\bm \theta}$ is smooth enough, which can be achieved via bounding ({or penalizing}) the norm of the neural network parameter $\bm \theta$ during the learning process. 

\subsection{Main Result}
\color{black}
Our main result is as follows:
\begin{theorem}\label{thm:main}
Suppose that the Assumptions~\ref{as:bound}-\ref{as:countinuous_g} hold true.
Assume that the entries of index sets $\bm \varOmega$ and $\bm \varXi$ are sampled from $[I_1]\times \ldots \times [I_K]$ uniformly at random  
{with replacement}
and satisfies
    $
        \bm \varOmega \subseteq \bm \varXi.
   $ 
   {In addition, let $\bm z_{\bi}, \dots, \bm z_{\bi_S}$ {be} the set of observed features}. {Assume} that each $\bm z_{\bi}$ is drawn from a distribution ${\cal D}$. 
Then, for $\delta > 0$, the following hold with probability of at least $1-5\delta-3e^{-\alpha(e^2-3)}$:
\begin{subequations}\label{eq:mainresults}
\begin{align}
 \frac{\left\|  \tlambda^\natural - \widehat{\xi} \widehat{\tlambda} \right\|^2_{\rm F}}{{\prod _k I_k}}&\leq  {\cal O}\left(\frac{1}{ p^2_{\min}}\varrho_1\right),\\
 \underset{\bm z_{\bi} \sim \mathcal{D}} {\mathbb{E}}\left[(p_{\bi}^\natural-\nicefrac{1}{\widehat{\xi}}\widehat{g}_{\bm \theta}(\bm z_{\bi}))^2\right] &\le {\cal O}\left(\frac{p_{\max}}{\beta p^2_{\min}}\varrho_1+\varrho_2\right),\label{eq:mainresults2}
\end{align}
\end{subequations}
where $\varrho_1 = F\alpha_u^K\left({{\eta} +\zeta^2(L_g+L_{\theta})^2 } +\sqrt{\frac{\log(1/\delta)}{|\bTheta|}}\right)$,
\begin{align*}
     \varrho_2 &= \sqrt{\frac{\log(1/\delta)}{S}} + \frac{\left(\|\bZ\|_{\rm F}\mathscr{R}_{\mathcal{G}}\right)^{1/4}}{\sqrt{S}},\\
 \eta^2 &=    \frac{K\sum_k I_k +  \|\bZ\|_{\rm F}\mathscr{R}_{\mathcal{G}}}{\sqrt{T}}  + c\sqrt{\frac{\log(1/\delta)}{T}} +\alpha\nu,  
\end{align*}
  $c = \alpha\max\{ |\log \beta|,\log\alpha  \}$, $\alpha = F\alpha_u^K p_{\max}$, $\beta = F\beta_u^K p_{\min}$,  $\bm Z=[\bm z_{\bi_1},\dots, \bm z_{\bi_{S}}] \in \mathbb{R}^{D \times S}$, $T=|\bOmega|$, and $\widehat{\xi}$ denote the global scaling ambiguity. 
\end{theorem}
The proof of Theorem~\ref{thm:main} is relegated to the supplementary material in Sec. \ref{app:main_theorem}. The result reveals several interesting insights. First, the results in \eqref{eq:mainresults} show that the proposed MLE criterion can correctly recover the average true counts $\lambda^\natural_{\bi}$'s and the detection probabilities $p^\natural_{\bi}$'s, up to {a certain} global scaling ambiguity between the entries. 
Second, the bounds indicate that the estimation {accuracy is} better when there are more observations (i.e., larger $|\bm \varOmega|$) and more similar features (i.e., smaller $\zeta$ and larger $|\bm \varTheta|$).
Third, the result suggests that an appropriate choice of the neural network class plays a role in ensuring good estimation accuracy. This is because if {one uses} deeper/wider neural networks, the approximation error $\nu$ becomes smaller, but the complexity paramater $\mathscr{R}_{\mathcal{G}}$ {is} larger. Hence, there is a trade-off {to strike} while choosing the network {architecture of} $g_{\bm \theta}$.

\section{Algorithm Design}\label{sec:algorithm}
The proposed MLE in \eqref{eq:max_likelihood_optim_tensor} is a combination of Poisson tensor decomposition and neural network learning---both are hard optimization problems.
To tackle this problem, we design a BCD-based \citep{razaviyayn2013unified,bertsekas1999nonlinear} algorithm. 
We first re-formulate \eqref{eq:max_likelihood_optim_tensor} using a regularized form by considering nonnegativity constraints for $\bm U_k$'s:
\begin{subequations}\label{eq:reformulated_mle}
\begin{align} 
	\underset{\{\bm U_k\}, \bm \theta,  \{p_{\bm i}\}}{\rm minimize}&~\frac{1}{|\bm \varOmega|}L(\bm U,\bm \theta, \tP)+\frac{\mu}{|\bm \varXi|} \sum_{\bm i \in \bm \varXi}\ell(g_{\bm \theta}(\bm z_{\bm i}),p_{\bm i}), \\
	{\rm subject\ to}~~ &
	\bm U_k \ge \bm 0, \forall k \in [K], \\
	& 0 \le p_{\bm i} \le 1, ~ \forall \bm i \in \bm \varOmega \cup \bm \varXi,~ g_{\bm \theta} \in {\cal G},
\end{align}
\end{subequations}
where $\mu > 0$ is a regularization parameter and $\ell(\cdot,\cdot)$ denotes a certain distance/divergence measure, e.g., the least squares function $\ell(x,y) = (x-y)^2$.

{
Note that in the re-formulated problem \eqref{eq:reformulated_mle}, we did not explicitly use the bounds on ${\cal U}_k$ and ${\cal G}$ in Assumption~\ref{as:bound}. 
The reason is twofold: First, the bounds serve for analytical purposes, yet not easy to know exactly in practice. Second, ignoring the bounds is inconsequential in terms of performance (as will be seen in experiments), but substantially simplifies the algorithm design.
Nonetheless, as one will see, the update of $\bm U_k$ naturally results in positive and bounded solutions, under reasonable conditions (cf. the comments after Eq.~\ref{eq:Uupdatecomment}). In addition, as $g_{\bm \theta}$ can be constructed to have a sigmoid output layer, the value of $g_{\bm \theta}(\z_{\bi})$ is naturally bounded away from 0 and 1.
}

Also note that we adopt this regularized optimization design since it gives flexibility to handle different cases such as $\bm \varOmega \subseteq \bm \varXi $ and $\bm \varOmega \supset \bm \varXi $---as we will see in the following section.  Our BCD updates are designed as follows:

{\bf The $\textbf{\textit{U}}_k$-Subproblem.}
When $ {\bm\theta}$, $p_{\bi}$, $\bm U_j,~j \neq k$ are fixed, the subproblem w.r.t. $\bm U_k$ is given by
{
\begin{align} \label{eq:Uk_subproblem1}
&\minimize_{\bm U_k \ge \bm 0}~ \sum_{\bm i \in \bm \varOmega}\left[\bigg(\sum_{f=1}^F {\bm U}_k (i_k,f) \prod_{j\neq k} \bm U_j(i_j,f)  \bigg) p_{\bm i}\right. \nonumber\\
&\left.~~~~-y_{\bm i}\log \bigg(\sum_{f=1}^F \bm U_k (i_k,f) \prod_{j\neq k} \bm U_j(i_j,f)\bigg)\right].
\end{align}
}
From \eqref{eq:Uk_subproblem1}, the update for $\bm U_k$ is as follows:
\begin{align}\label{eq:Uupdatecomment}
   \bm U_k(i_k,f) \leftarrow  \frac{\sum_{\bm i \in \bm \varOmega}y_{\bm i}\alpha_{\bm i}^{(f)}}{\sum_{\bm i \in \bm \varOmega} \prod_{j\neq k} \bm U_j(i_j,f) p_{\bm i}}, ~\forall \bm i, f,
\end{align}
where $\alpha_{\bm i}^{(f)} = \frac{\overline{\bm U}_k (i_k,f) \prod_{j\neq k} \bm U_j(i_j,f)}{\sum_{f=1}^F \overline{\bm U}_k (i_k,f) \prod_{j\neq k} \bm U_j(i_j,f) }$ and $\overline{\bm U}_k$ {denotes $\bm U_k$ from the previous iteration}.
{Note that the above solution is always {strictly} positive if $\U_k>\bm 0$ for all $k$ and $p_{\bi}>0$ for all {$\bi$}.
} 
Our update rule design for $\bm U_k$ follows the ideas in \citep{chi2012tensors,fu2019link}, which is essentially a {\it majorization minimization} step w.r.t. $\U_k$; {see Sec. \ref{app:algo} in the supplementary material.} 

{\bf The $p$-Subproblem.} To update $p_{\bi}$, we fix $\bm U_k$'s and $\bm \theta$ and consider different cases. 

{\it Case 1.} When both $y_{\bi}$ and $\bm z_{\bi}$ are available, i.e., $\bi \in \bm \varOmega \cap \bm \varXi$, $p_{\bi}$ is updated via optimizing the following subproblem:
\begin{align}\label{eq:pcase1}
    \min_{p_{\bi}\in[0,1]}&\frac{1}{|\bm \varOmega|}(\lambda_{\bm i}p_{\bm i}-y_{\bm i}\log(p_{\bm i})) +\frac{\mu}{|\bm \varXi|}  \ell(g_{\bm \theta}(\bm z_{\bm i}),p_{\bm i}),
\end{align}
where $\lambda_{\bm i}=\sum_{f=1}^F  \prod_{k=1}^K \bm U_k(i_k,f)$ is obtained using the current estimates of $\bm U_k$'s.

{\it Case 2.} Suppose that the observation $y_{\bi}$ is not available, but $\bm z_{\bi}$ is available, i.e., ${\bi} \in \bm \varXi - \bm \varOmega \cap \bm \varXi$. Then, we consider the following subproblem to update such $p_{\bi}$'s:
\begin{align}\label{eq:pcase2}
	\min_{p_{\bi}\in[0,1]}\ell(g_{\bm \theta}(\bm z_{\bm i}),p_{\bm i}).
\end{align}

{If we choose {a convex $\ell(\cdot)$} in \eqref{eq:pcase1}-\eqref{eq:pcase2}, the problems can be optimally solved using any standard techniques such as projected gradient decent. Moreover, if we {set $\ell(\cdot)$ to be one of the} popular choices such {Euclidean} distance {and} KL divergence, closed-form solutions for the problems in \eqref{eq:pcase1}-\eqref{eq:pcase2} exist---see Table \ref{tab:pi_updates} in the supplementary material.}

{\it Case 3.}\footnote{Our {recoverability} analysis is built upon the assumption $\bm \varOmega \subseteq \bm \varXi $; see Theorem \ref{thm:main}. However, in practice, there are cases where $y_{\bi}$ is observed, but $\bm z_{\bi}$ is not available (i.e., $\bm \varOmega - \bm \varOmega \cap \bm \varXi \neq \emptyset$). Our algorithm can still operate under such cases.} 
When $y_{\bi}$ is observed, but $\bm z_{\bi}$ is not accessible, i.e., $\bi \in \bm \varOmega - \bm \varOmega \cap \bm \varXi$, we update $p_{\bi}$ via solving the following subproblem:
	\begin{align}\label{eq:pcase3}
    \min_{p_{\bi}\in(0,1]}~\lambda_{\bm i}p_{\bm i}-y_{\bm i}\log(p_{\bm i}),
\end{align}
which gives the update rule for $p_{\bi}$ as $p_{\bi}\leftarrow \left[\nicefrac{y_{\bi}}{\lambda_{\bi}}\right]_{[0,1]}$.

{\bf The {$\bm \theta$}-Subproblem. } \label{sec:NNdesign}
 By fixing the $\bm U_k$'s and all the $p_{\bi}$'s, the subproblem for updating $\bm \theta$ is as follows:
	\begin{align*}
	\min_{\bm \theta}\sum_{\bm i \in \bm \varXi}\ell(g_{\bm \theta}(\bm z_{\bm i}),p_{\bm i}).
	\end{align*}
This is an unconstrained neural network learning problem.  Many {off-the-shelf} neural network learning algorithms that use gradient, e.g., \texttt{Adam} \citep{kingma2015adam} and \texttt{Adagrad} \citep{duchi2011adaptive}, can be used to handle this subproblem.

More design details and the description of the algorithm are provided in supplementary material in Sec. \ref{app:algo}.  
{The proposed algorithm is referred to} as the {\it \underline{Un}der-\underline{C}ounted Data Prediction Via \underline{Le}arner-Aided \underline{T}ensor \underline{C}ompletion} (\texttt{UncleTC}) algorithm.

{A remark is that our algorithm uses a batch (instead of stochastic) update rule for the $\U_k$'s. This can also be replaced by stochastic tensor decomposition algorithms (see, e.g., \citep{fu2020block,pu2022stochastic} and the references in \citep{fu2020computing}). Using stochastic algorithms for the $\U_k$-subproblems may further improve efficiency.}

\section{Experiments}
{In this section, we evaluate our proposed method through a series of synthetic and real data experiments.} {The source code is available at \url{https://github.com/shahanaibrahimosu/undercounted-tensor-completion}.}
{\paragraph{Baselines.}
We employ a number of {low-rank} tensor completion-based baselines, namely \texttt{NTF-CPD-KL}\citep{chi2012tensors}, {\texttt{ HaLRTC} }\citep{liu2012tensor},  \texttt{NTF-CPD-LS} \citep{shashua2005nonnegative}, \texttt{BPTF-CPD} \citep{schein2015bayesian}, and \texttt{NTF-Tucker-LS} \citep{kim2007nonnegative}. Among them, both \texttt{NTF-CPD-KL} and \texttt{BPTF} consider Poisson modeling, but they do not have a Binomial {detection stage in their models} to accommodate the under-counting effect. 
We also consider two recent NN-based tensor completion methods, \texttt{CostCo} \citep{liu2019costco} and \texttt{POND} \citep{tillinghast2020prob}, as baselines. In order to incorporate the side features, the \texttt{CostCo} method is slightly modified from its original implementation by concatenating side features with the indices of the observed entries as the input to their networks. 
In addition, we also include a tensor version of \citep{fu2019link}, where a linear {function} $ g_{\bm \theta}({\bm z_{\bi}}) = \bm \theta_1^\T\bm z_{\bi} + \theta_2$ is used to learn $g$ function. This baseline is referred to as \texttt{UncleTC(Linear)}.
More details on the implementation are provided in the supplementary material in Sec. \ref{app:exp}.}
\subsection{Synthetic-Data Simulations} \label{sec:experiments}

\begin{table}[t]
  \centering
\caption{Average $\text{MAE}_{p}$ and $\text{MAE}_{\lambda}$ over 20 random trials {under} various values {of} $\gamma_{\varOmega}$; $\gamma_{\bm \varXi}=0.3, \gamma_{\Theta}=0.2, {\rm SNR}=40$dB with $\bm \varTheta \cup \bm \varOmega \subseteq \bm \varXi$.  
}  \resizebox{0.99\linewidth}{!}{
    \begin{tabular}{|c|c|c|c|c|}
    \hline
    \textbf{Algorithm} & \textbf{Metric} & \multicolumn{1}{c|}{$\gamma_{\varOmega}=0.05$} & \multicolumn{1}{c|}{$\gamma_{\varOmega}=0.1$} & \multicolumn{1}{c|}{$\gamma_{\varOmega}=0.3$} \\
    \hline
    \hline
    \multirow{2}[4]{*}{\texttt{UncleTC}} & $\text{MAE}_{p}$ & \textbf{0.156 $\pm$ 0.055} & \textbf{0.111 $\pm$ 0.038} & \textbf{0.063 $\pm$ 0.047}  \\
\cline{2-5}          & $\text{MAE}_{\lambda}$   & \textbf{0.104 $\pm$ 0.063} & \textbf{0.063 $\pm$ 0.041} & \textbf{0.037 $\pm$ 0.031} \\
    \hline
    \multirow{2}[4]{*}{\texttt{UncleTC(Linear)}} & $\text{MAE}_{p}$   & 0.231 $\pm$ 0.072 & 0.189 $\pm$ 0.046 & 0.193 $\pm$ 0.180 \\
\cline{2-5}          & $\text{MAE}_{\lambda}$  & 0.135 $\pm$ 0.069 & 0.092 $\pm$ 0.039 & 0.090 $\pm$ 0.089 \\
    \hline
    {\texttt{NTF-CPD-KL}} &  $\text{MAE}_{\lambda}$   & 0.306 $\pm$ 0.119 & 0.258 $\pm$ 0.128 & 0.190 $\pm$ 0.124 \\
    \hline
     {\texttt{BPTF}} &  $\text{MAE}_{\lambda}$   & 0.547 $\pm$ 0.143	& 0.493$ \pm$ 0.161 &	0.345 $\pm$ 0.104 \\
    \hline
    {\texttt{HaLRTC}} &  $\text{MAE}_{\lambda}$   & 0.503 $\pm$ 0.382 &	0.495 $\pm$ 0.391 &	0.436 $\pm$ 0.282 \\
    \hline
    {\texttt{NTF-LS}} &  $\text{MAE}_{\lambda}$   & 0.742 $\pm$ 0.462	&0.684 $\pm$ 0.421	& 0.403 $\pm$ 0.099 \\
    \hline
    {\texttt{NTF-Tucker-LS}} &  $\text{MAE}_{\lambda}$   & 0.639 $\pm$ 0.494	& 0.526 $\pm$ 0.299 &	0.341 $\pm$ 0.072 \\
    \hline
    {\texttt{CostCo}} &  $\text{MAE}_{\lambda}$   & 0.964 $\pm$ 0.235 &	0.853 $\pm$ 0.213 &	0.809 $\pm$ 0.147 \\
    \hline
    {\texttt{POND}} &  $\text{MAE}_{\lambda}$   & 0.841 $\pm$ 0.146 &	0.877 $\pm$ 0.098 &	0.885 $\pm$ 0.087 \\
    \hline
    \hline
    \end{tabular}%
    }
  \label{tab:varying_Omega}%
\end{table}%

\begin{table}[t]
  \centering
\caption{Average $\text{MAE}_{p}$ and $\text{MAE}_{\lambda}$ over 20 random trials {under} various values of $\zeta$;  $\gamma_{\varOmega}=0.2, \gamma_{\bm \varXi}=0.3,  \gamma_{\bm \varTheta}= 0.2$ with $\bm \varTheta \cup \bm \varOmega \subseteq \bm \varXi$. 
}
  \resizebox{0.99\linewidth}{!}{
    \begin{tabular}{|c|c|c|c|c|}
    \hline
    \textbf{Algorithm} & \textbf{Metric} & \multicolumn{1}{c|}{${\rm SNR}=0$dB} & \multicolumn{1}{c|}{${\rm SNR}=10$dB} & \multicolumn{1}{c|}{${\rm SNR}=40$dB} \\
    \hline
    \hline
    \multirow{2}[4]{*}{\texttt{UncleTC}} & $\text{MAE}_{p}$  & \textbf{0.154 $\pm$ 0.007} & \textbf{0.047 $\pm$ 0.018} & \textbf{0.035 $\pm$ 0.001} \\
\cline{2-5}          & $\text{MAE}_{\lambda}$ &  \textbf{0.089 $\pm$ 0.010} & \textbf{0.026 $\pm$ 0.007} & \textbf{0.023 $\pm$ 0.001} \\
    \hline
    \multirow{2}[4]{*}{\texttt{UncleTC(Linear)}} & $\text{MAE}_{p}$  & 0.304 $\pm$ 0.012 & 0.109 $\pm$ 0.041 & 0.129 $\pm$ 0.049 \\
\cline{2-5}          & $\text{MAE}_{\lambda}$ & 0.193 $\pm$ 0.054 & 0.052 $\pm$ 0.016 & 0.065 $\pm$ 0.020 \\
    \hline
   {\texttt{NTF-CPD-KL}} &  $\text{MAE}_{\lambda}$ &  0.299 $\pm$ 0.055 & 0.111 $\pm$ 0.001 & 0.107 $\pm$ 0.001 \\
   \hline
     {\texttt{BPTF}} &  $\text{MAE}_{\lambda}$   & 0.420 $\pm$ 0.051	& 0.414 $\pm$ 0.047	& 0.427 $\pm$ 0.070\\
    \hline
    {\texttt{HaLRTC}} &  $\text{MAE}_{\lambda}$   & 0.423 $\pm$ 0.068 &	0.441 $\pm$ 0.109 &	0.440 $\pm$ 0.126 \\
    \hline
    {\texttt{NTF-LS}} &  $\text{MAE}_{\lambda}$   & 0.479 $\pm$ 0.062 &	0.474 $\pm$ 0.057 &	0.472 $\pm$ 0.046 \\
    \hline
    {\texttt{NTF-Tucker-LS}} &  $\text{MAE}_{\lambda}$   & 0.366 $\pm$ 0.047	& 0.375 $\pm$ 0.050 &	0.378 $\pm$ 0.045 \\
    \hline
    {\texttt{CostCo}} &  $\text{MAE}_{\lambda}$   & 0.987 $\pm$ 0.256 &	0.876 $\pm$ 0.432 &	0.890 $\pm$ 0.357 \\
    \hline
    {\texttt{POND}} &  $\text{MAE}_{\lambda}$   & 0.824 $\pm$ 0.096 &	0.785 $\pm$ 0.131 &	0.837 $\pm$ 0.087 \\
    \hline
    \hline
    \end{tabular}%
    }
  \label{tab:varying_SNR}%
\end{table}%

\begin{table}[t]
  \centering
\caption{Average $\text{MAE}_{p}$ and $\text{MAE}_{\lambda}$ over 20 random trials {under} various values of $\gamma_{\varTheta}$;  $\gamma_{\varOmega}=0.2, \gamma_{\bm \varXi}=0.7, {\rm SNR}=40$dB with $\bm \varTheta \cup \bm \varOmega \subseteq \bm \varXi$.  
}  \resizebox{0.99\linewidth}{!}{
    \begin{tabular}{|c|c|c|c|c|}
    \hline
    \textbf{Algorithm} & \textbf{Metric} & \multicolumn{1}{c|}{$\gamma_{\Theta}=0$} & \multicolumn{1}{c|}{$\gamma_{\Theta}=0.3$} & \multicolumn{1}{c|}{$\gamma_{\Theta}=0.5$} \\
    \hline
    \hline
    \multirow{2}[4]{*}{\texttt{UncleTC}} & $\text{MAE}_{p}$ & \textbf{0.163 $\pm$ 0.007} & \textbf{0.126 $\pm$ 0.044} & \textbf{0.104 $\pm$ 0.076} \\
\cline{2-5}          & $\text{MAE}_{\lambda}$   & \textbf{0.062 $\pm$ 0.003} & \textbf{0.055 $\pm$ 0.028} & \textbf{0.054 $\pm$ 0.148} \\
    \hline
    \multirow{2}[4]{*}{\texttt{UncleTC(Linear)}} & $\text{MAE}_{p}$   & 0.231 $\pm$ 0.005 & 0.252 $\pm$ 0.029 & 0.201 $\pm$ 0.081 \\
\cline{2-5}          & $\text{MAE}_{\lambda}$  & 0.078 $\pm$ 0.005 & 0.104 $\pm$ 0.041 & 0.214 $\pm$ 0.174 \\
    \hline
    {\texttt{NTF-CPD-KL}} &  $\text{MAE}_{\lambda}$   & 0.185 $\pm$ 0.006 & 0.107 $\pm$ 0.002 & 0.156 $\pm$ 0.078 \\
   \hline
     {\texttt{BPTF}} &  $\text{MAE}_{\lambda}$   & 0.381 $\pm$ 0.024 &	0.445 $\pm$ 0.092	& 0.473 $\pm$ 0.174\\
    \hline
    {\texttt{HaLRTC}} &  $\text{MAE}_{\lambda}$   & 0.436 $\pm$ 0.030	 & 0.533 $\pm$ 0.212 &	0.561 $\pm$ 0.352 \\
    \hline
    {\texttt{NTF-LS}} &  $\text{MAE}_{\lambda}$   & 0.434 $\pm$ 0.030 &	0.472 $\pm$ 0.068 &	0.644 $\pm$ 0.412 \\
    \hline
    {\texttt{NTF-Tucker-LS}} &  $\text{MAE}_{\lambda}$   & 0.345 $\pm$ 0.039 &	0.411 $\pm$ 0.080 &	0.564 $\pm$ 0.337 \\
    \hline
    {\texttt{CostCo}} &  $\text{MAE}_{\lambda}$   & 0.876 $\pm$ 0.178 &	0.814 $\pm$ 0.301	& 0.987 $\pm$ 0.239 \\
    \hline
    {\texttt{POND}} &  $\text{MAE}_{\lambda}$   & 0.926 $\pm$ 0.158 &	0.851 $\pm$ 0.165 &	0.859 $\pm$ 0.112
 \\
    \hline
    \hline
    \end{tabular}%
    }
  \label{tab:varying_Theta}%
\end{table}%

{\bf Data Generation.} The data generating process follows \eqref{eq:nonlinear_model_tensor}. We first generate $\bm U_k \in \mathbb{R}^{I_k \times F}, ~\forall k$ by randomly sampling its entries from a uniform distribution between 0 and $\kappa$.  
We fix $K=3$, $F=3$ and $I_k=20,~ \forall k$, unless specified {otherwise}. 
We also fix {$\kappa=15$} so that $\lambda_{\bm i}$'s are not unreasonably small.  The feature vectors $\bm z_{\bm i} \in \mathbb{R}^{D}$ {with $D=10$} are generated by randomly sampling its entries from {the standard normal distribution}. 
We {use the} ground-truth nonlinear function (unknown to our algorithm)  $ g(\bm z) = {\rm sigmoid}(\bm \nu^{\top}(0.5\bm z^3+0.2\bm z))$, 
where the vector $\bm \nu \in \mathbb{R}^{D}$ is generated by randomly sampling its entries from {the} uniform distribution between 0 and 1. 
We observe only a subset of $y_{\bi}$'s such that each $\bi$ belongs to $\bm \varOmega$ with probability $\gamma_\varOmega \in (0,1]$.  Similarly, only a portion of $\bm z_{\bi}$'s are observed such that each $\bi$ belongs to $\bm \varXi$ with probability $\gamma_\varXi \in (0,1]$.
 We also make a subset of $\bm z_{\bi}$'s similar such that
$
    \bm z_{\bi} = \bm \varphi + \bm n_{\bi},~ \bi \in \bm \varTheta,~\bm \varTheta \subseteq \bm \varXi,
$
with each $\bi$ belongs to $\bm \varTheta$ with probability $\gamma_{\varTheta} \in (0,1]$.
Here, $\bm \varphi \in \mathbb{R}^D$ is a random vector whose entries are drawn from the standard normal distribution and $\bm n_{\bi}$ denotes additive noise whose entries are zero mean Gaussian with variance $\sigma^2$. Under this setting, we define the {\it signal-to-noise} (SNR) ratio ${\rm SNR} = 10\log_{10}\left(\nicefrac{\|\bm \varphi\|_2^2}{D \sigma^2}\right)$dB.  As the value of ${\rm SNR}$ increases, the {feature} vectors $\bm z_{\bi} \in \bm \varTheta$ are more similar.

{\bf Metric.} To characterize the performance of recovering $p_{\bi}$'s and $\lambda_{\bi}$'s, we employ \textit{mean absolute error} (${\sf MAE}$) which is defined as ${\sf MAE}_\lambda = \nicefrac{1}{\prod_k I_k}\sum_{\bm i}\left |\nicefrac{\widehat{\lambda}_{\bm i}}{\widehat{\lambda}^{(\text{avg})}} -\nicefrac{\lambda^\natural_{\bm i}}{\lambda^{\natural(\text{avg})} }\right|,$
where $\widehat{\lambda}_{\bm i}$ is the estimated value corresponding to $\lambda^\natural_{\bm i}$, and the subscript ${(\text{avg})}$ denotes the average across all values, {e.g.}, $\lambda^{(\text{avg})} = \nicefrac{1}{\prod_k I_k}\sum_{\bm i } \lambda_{\bi}$. Note that the metric ${\sf MAE}_\lambda$ {involves} entry-wise normalization of the estimates and the ground truth by the average in order to {remove} the scaling ambiguity. {The metric ${\sf MAE}_p$ is defined in the same way, with all the $\lambda$-terms replaced by their $p$ counterparts.}.

{\bf Implementation Settings.}
 To learn the nonlinear function, we use a neural network $ g_{\bm \theta}(\cdot)$  with 3 hidden layers and 20 {\sf ReLU} activation functions {in each hidden layer}. {The regularization parameter $\mu$ is fixed to be 2000.  More implementation details and experiments using different $\mu$'s are provided in the supplementary material in Sec. \ref{app:exp}.

\paragraph{Results.}
Table.~\ref{tab:varying_Omega} {shows the performance under various $\gamma_{\bm \varOmega}$'s of all the methods under test}. 
{Note that for the baselines that do not take detection probability into their models, only the ${\sf MAE}_{\lambda}$ results are presented.}  
{The results {show} that all the {non-Poisson} baselines including the NN-based methods struggle to estimate the true counts $\lambda_{\bi}$. This shows the importance of tailored modeling for integer data.}
{Among the algorithms using Poisson models,
the proposed \texttt{UncleTC} mothod estimates $\lambda_{\bm i}$'s with much higher accuracy relative to $\texttt{NTF-CPD-KL}$, as our method explicitly models the under-counting effect}. One can also note that \texttt{UncleTC} significantly outperforms \texttt{UncleTC(Linear)}, in {terms of both ${\sf MAE}_p$ and ${\sf MAE}_{\lambda}$}. This is due to the ability of \texttt{UncleTC} to better handle the nonlinear relationships between the detection probabilities $p_{\bi}$ and the side information $\bm z_{\bi}$, by using its NN-based learning.
In addition, we note that as $\gamma_{\bm \varOmega}$ is higher, i.e., more observations are available, the performance becomes better, {as expected and indicated in our theoretical claim.}

In Table \ref{tab:varying_SNR} and Table \ref{tab:varying_Theta}, we show the {performance of the algorithms under vairous} ${\rm SNR}$ and $\gamma_{\bm \varTheta}$, respectively.
In Table \ref{tab:varying_SNR}, as {\rm SNR} increases from 0 to 40dB, i.e., $\zeta$ in Assumption \ref{as:similar_z} becomes smaller and the feature vectors $z_{\bi}$'s become more similar, the {MAE}s decrease noticeably. {The impact on $\text{MAE}_p$
is obviously spelled out, as a reduction of around 7 times is observed}. {Table \ref{tab:varying_Theta} shows that as the size of $\bm \varTheta$ gets larger, the estimation accuracy {of the proposed method is positively impacted}. Both tables echo the results presented by Theorem \ref{thm:main}.

\begin{table}[t]
  \centering
  \caption{Statistics of $\nicefrac{\widehat{\lambda}_{\bm i}}{{\lambda}^\natural_{\bm i}}$ and $\nicefrac{\widehat{p}_{\bm i}}{{p}^\natural_{\bm i}}$ for different trials of the proposed \texttt{UncleTC} with settings $\gamma_{\varOmega}=0.3, \gamma_{\varXi}=0.3, \gamma_{\bm \varTheta}=0.2, \zeta=40\text{dB}$, and $g(\bm z) = {\rm sigmoid}(\bm \nu^{\top}(0.1\log(\bm z^2)+0.1 \bm z^2))$}
  \resizebox{0.99\linewidth}{!}{
    \begin{tabular}{|c|c|c|c|}
    \hline
     \#Trial  & mean$\pm$std of $\nicefrac{\widehat{\lambda}_{\bm i}}{{\lambda}^\natural_{\bm i}}$  & mean$\pm$std of $\nicefrac{\widehat{p}_{\bm i}}{{p}^\natural_{\bm i}}$ & mean$\pm$std of $\nicefrac{\widehat{\lambda}_{\bm i}\widehat{p}_{\bm i}}{{\lambda}^\natural_{\bm i}{p}^\natural_{\bm i}}$  \\
    \hline
    \hline
    1     & 1.16 $\pm$ 0.04 & 0.86 $\pm$ 0.08 & 0.99 $\pm$ 0.08 \\
    \hline
    2     & 1.29 $\pm$ 0.04 & 0.78 $\pm$ 0.08 & 1.00 $\pm$ 0.07 \\
    \hline
    3     & 0.98 $\pm$ 0.03 & 1.02 $\pm$ 0.06 & 0.99 $\pm$ 0.07 \\
    \hline
    4     & 1.33 $\pm$ 0.05 & 0.76 $\pm$ 0.10 & 1.00 $\pm$ 0.09 \\
    \hline
    5     & 1.09 $\pm$ 0.03 & 0.93 $\pm$ 0.06 & 1.00 $\pm$ 0.06 \\
    \hline
    6     & 1.16 $\pm$ 0.05 & 0.86 $\pm$ 0.11 & 0.99 $\pm$ 0.12 \\
    \hline
    7     & 1.07 $\pm$ 0.03 & 0.94 $\pm$ 0.07 & 1.00 $\pm$ 0.06 \\
    \hline
    8     & 1.24 $\pm$ 0.03 & 0.81 $\pm$ 0.05 & 0.99 $\pm$ 0.05 \\
    \hline
    9     & 0.83 $\pm$ 0.02 & 1.21 $\pm$ 0.05 & 1.00 $\pm$ 0.04 \\
    \hline
    10    & 0.80 $\pm$ 0.02 & 1.23 $\pm$ 0.06 & 0.99 $\pm$ 0.07 \\
    \hline
    \hline
    \end{tabular}%
    }
  \label{tab:scaling}%
\end{table}%

Table \ref{tab:scaling} validates a key result from Theorem \ref{thm:main}. {That is,} there exists a global scaling {factor and a counter-scaling factor} for $\lambda_{\bi}$ and $p_{\bi}$, {respectively}. In other words, assuming perfect estimation for $\widehat{\lambda}_{\bm i}$ and $\widehat{p}_{\bi}$ by the proposed \texttt{UncleTC}, we have
$
 \widehat{\lambda}_{\bm i} = \xi  {\lambda}^\natural_{\bm i}, ~ \widehat{p}_{\bm i} =  \nicefrac{1}{\xi}{p}^\natural_{\bm i}, \forall \bm i
$ for a certain scalar $\xi > 0$.
 To see if the \texttt{UncleTC} algorithm attains good estimation of $\lambda_{\bi}$ and $p_{\bi}$ up to a global scaling/counter-scaling ambiguity, we report the mean and standard deviation of $\nicefrac{\widehat{\lambda}_{\bm i}}{{\lambda}^\natural_{\bm i}}$ and $\nicefrac{\widehat{p}_{\bm i}}{{p}^\natural_{\bm i}}$ 
 {taken over all the ${\bm i}$'s}
 in each random trial. {One can see that only small standard deviations exist in all the trials, suggesting that the scaling/counter-scaling factors are approximately identical over all the indices.}   Similar observations {are made on} $\nicefrac{\widehat{p}_{\bm i}}{{p}^\natural_{\bm i}}$. One can also note that {the scaling factors of $\widehat{\lambda}_{\bi}$ and $\widehat{p}_{\bi}$} are approximately reciprocal to each other ({cf. the last column of Table~\ref{tab:scaling}}). This observations corroborate the findings in Theorem \ref{thm:main}.

\subsection{Real-Data Experiments}
{\bf COVID-19 Data.}
The dataset is obtained from the COVID-19 impact analysis platform hosted by the University of Maryland, College Park, United States \citep{zhang2020interactive}. 
The dataset includes the number of reported COVID-19 cases of about 2270  US counties during 2020-2021, with 25 different attributes such as social distancing index, population density, testing capacity and so on. 
In order to extract a count-type tensor from the COVID-19 dataset, we represent each county using two coordinates, i.e., the rounded value of latitude and longitude of its geographical center.
This way, we obtain an integer tensor of size $43 \times 80 \times 60$ (with 18.42\% of entries observed), recording the COVID-19 case numbers of 642 counties over 60 days from January 1, 2021 to March 1, 2021.

{\bf Pollinator-Plant Interaction (PPI) Data.} 
The Plant-Pollinator Interaction (PPI) dataset is a publicly available dataset, collected by the researchers at the H.J. Andrews
(HJA) Long-term Ecological Research site in Oregon, USA \citep{Jones2017,Daly1957}. 
We consider a subset of the dataset by selecting 50 plants and 50 pollinators that {have} the highest {numbers} of interactions. This leads to a $50 \times 50 \times 37$ integer tensor with $22.06\%$ observed entries. 
 The dataset also includes about 37 entry attributes corresponding to each observation, such as flower color of the plant, pollinator tongue length, weather conditions, and so on. 

  For both datasets, since the features are in different numerical units and range, we normalize each feature value using min-max normalization \citep{jain2005score}, across all observations. More details on the datasets and tensor construction are given in the supplementary material in Sec. \ref{app:exp}. 
  \begin{table}[t]
  \centering 
\caption{The {\sf rRMSE} of various methods {on the} COVID-19 and PPI {datasets}.}
 \resizebox{0.75\linewidth}{!}{
\begin{tabular}{|l|c|c|}
    \hline
    \textbf{Method} & \textbf{COVID-19} & \textbf{PPI} \\
    \hline
    \hline
    \texttt{UncleTC} & \textbf{3.52 $\pm$ 0.38} & \textbf{9.89 $\pm$ 1.67} \\
    \hline
    \texttt{UncleTC(Linear)} & 3.90 $\pm$ 0.33 & 10.21 $\pm$ 1.45 \\
    \hline
    \texttt{HaLRTC} & 6.14 $\pm$ 0.45 & 10.77 $\pm$ 1.25 \\
    \hline
   \texttt{BPTF} & 6.72 $\pm$ 0.42 & 11.02 $\pm$ 1.39 \\
    \hline
   \texttt{NTF-CPD-KL} & \textbf{3.61 $\pm$ 0.52} & \textbf{10.17 $\pm$ 1.12} \\
    \hline
    \texttt{NTF-CPD-LS} & 7.00 $\pm$ 0.38 & 11.25 $\pm$ 1.27 \\
    \hline
    \texttt{NTF-Tucker-LS} & 6.82 $\pm$ 0.48 & 11.22 $\pm$ 1.27 \\
    \hline
    \texttt{CostCo} & 7.12 $\pm$ 0.34 & 11.01 $\pm$ 1.14 \\
    \hline
    \texttt{POND} & 6.92 $\pm$ 0.51 & 10.95 $\pm$ 1.31 \\
    \hline
    \hline
    \end{tabular}%
    }
  \label{tab:covid_ppi}
\end{table}%
  \begin{figure}[t]
	\centering
	\includegraphics[scale=0.44]{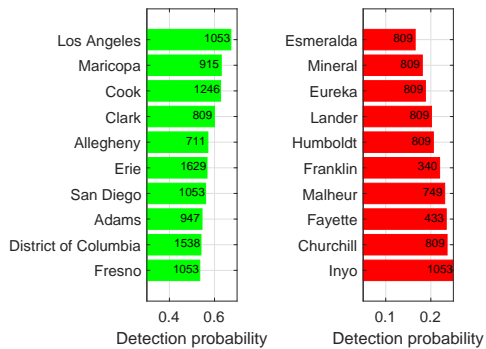}
	\caption{The top-10 counties that have the highest \texttt{UncleTC}-estimated average detection probabilities (left)  and lowest average detection probabilities (right) for COVID-19. Inside the bar, we show the average {numbers} of COVID-19 tests done per 1000 people during the selected period.}
 \label{fig:detp_covid}
\end{figure}

\paragraph{Quantitative Evaluation.}

Since the ground-truth $\lambda_{\bi}^\natural$'s and $p_{\bi}^\natural$'s are unknown for real data, we employ a count data prediction metric, i.e., \textit{relative root mean square} ({\sf rRMSE}) \citep{fu2019link}
for evaluation. Let $\{y_{\bm i}\}_{\bm i \in \bm \varDelta}$ be the set of actual holdout observations and $\{\widehat{y}_{\bm i}\}_{\bm i \in \bm \varDelta}$ be the corresponding predicted values. Then, the {\sf rRMSE} is defined as $
	{\sf rRMSE} = \nicefrac{1}{\overline{y}}\sqrt{\nicefrac{\sum_{\bm i \in \bm \varDelta}|y_{\bm i} - \widehat{y}_{\bm i}|^2}{|\bm \varDelta|}}$,
where $\overline{y}$ is the mean value of the actual observations $\{y_{\bm i}\}_{\bm i \in \bm \varDelta}$. 
Note that, for the proposed method, the observations are predicted via $\widehat{y}_{\bm i} = \widehat{\lambda}_{\bm i}\widehat{p}_{\bm i}$.

Table \ref{tab:covid_ppi} presents the performance of various approaches on the COVID-19 and PPI datasets. One can see that the proposed \texttt{UncleTC} exhibits promising performance on both datasets.  One can also see that the performance of the Poisson model-based method \texttt{NTF-CPD-KL} {has a comparable {\sf rRMSE} performance relative to our method}. This is expected, as the Poisson-Binomial model with parameters $\lambda$ and $p$ can be considered as a Poisson model with the parameter $\lambda p$ (see Lemma \ref{lem:key}). 
{Nontheless, our method still outperforms \texttt{NTF-CPD-KL}, as the incorporation of the entry attributes in our framework may be beneficial.}
{Compared to the non-Poisson methods, our approach stands out with obviously lower {\sf rRMSEs}, which is similar to what was observed in the simulations.} 
It can also be noted that \texttt{UncleTC} outperforms its {linear function-based counterpart} \texttt{UncleTC(Linear)} in both cases, which is also consistent with our simulation results.

{\bf Qualitative Evaluation.}
In Fig. \ref{fig:detp_covid}, we analyze the detection probabilities output by the proposed \texttt{UncleTC} for COVID-19 dataset. Here, we present the top 10 counties having the highest and lowest detection probabilities for COVID-19 cases found by the proposed \texttt{UncleTC} (averaged over all days).   Note that the estimated detection probabilities have a global scaling ambiguity---see Theorem \ref{thm:main}---but their relative ranking order is not affected by this ambiguity. We also report the average number of COVID-19 tests done per 1000 people during the selected period (we removed this feature while training the model to avoid correlated results). One can see that the counties with more tests done are mostly aligned with higher detection probabilities estimated by our algorithm. Also, most counties reported with high detection probabilities are populous, urban counties, which enjoy better infrastructure. These observations suggest that the proposed algorithm outputs plausible results for the detection probabilities---also see the detection probability results for the PPI dataset in the supplementary material in Sec. \ref{app:exp}.

\section{Conclusion}

We proposed an under-counted tensor completion framework, which is motivated by the commonly seen under-counting effects in data acquisition across many domains, e.g., ecology and epidemiology.
Our model uses a Poisson-Binomial data-generating perspective as often advocated in computational ecology, where the true integer data are under-counted by a Binomial detection process.
Unlike existing approaches that treat the under-counting effects using relatively simple models, we used a neural network-based design to account for a wide range of unknown nonlinear relations between the side information (attributes) and the under-counting effects. More importantly, we showed that the proposed under-counted tensor completion criterion can {\it provably} recover the underlying Poisson tensor and its entries' under-counting characteristics, under reasonable conditions---which is the first theoretical result of its kind.
We validated our theorem and algorithm using simulated and real data and observed intuitively plausible results from our real data results.

{{\bf Limitations.} A key limitation lies in the model assumption that the under-counted probabilities are functions of observed side attributes. However, such attributes may not always be available.
Provable UC-TC criteria that do not rely on such assumption are highly desirable, yet the current formulation and analysis cannot cover such settings without substantial changes and re-design.
}

{{\bf Acknowledgement.} This work is supported in part by the National Science Foundation (NSF) under Project NSF IIS-1910118. {The PPI dataset is provided by the HJ Andrews Experimental Forest research program, funded by the National Science Foundation's Long-Term Ecological Research Program (DEB 2025755), US Forest Service Pacific Northwest Research Station, and Oregon State University}.} The authors thank Dr. Julia A. Jones for helpful guidance with the PPI data and associated features.

\bibliographystyle{icml2023}

\onecolumn

\newpage
\appendix
\begin{center}
	\normalsize
	{\bf 
		Supplementary Material of ``
		Under-Counted Tensor Completion with Neural Incorporation of Attributes''    }
\end{center}

\section{Notation} 
The notations $x$, $\x$, $\X$ and $\tX$ represent a scalar, a vector, a matrix, and a tensor, respectively. $x_i$ denotes the $i$th element of the vector $\bm x$. $[\X]_{i,j}$ and $\X(i,j)$ both mean the $(i,j)$th entry of $\X$. 
 Similarly,  $[\tX]_{\bi}$ belongs to the $\bi$th entry of the tensor $\tX \in \mathbb{R}^{I_1 \times \ldots \times I_K}$ where $\bi= (i_1,\ldots,i_K)$.  $\mathbb{Z}_+$ denote the set of all nonnegative integers. $\bm x \ge \bm 0$, $\bm X \ge \bm 0$, and $\tX \ge \bm 0$ imply that all the associated entries are greater than 0.
$\|\x\|_2$, $\|\X\|_{\rm F}$ and $\|\tX\|_{\rm F}$ all mean the Euclidean (Frobenius) norm of the augment. $\|[\tX]_{\bm \varOmega}\|_{\rm F}$ indicates the Euclidean norm of the vector formed by concatenating the entries indexed by all unique $\bi$'s belonging to $\bm \varOmega$, i.e., $\|[\tX]_{\bm \varOmega}\|^2_{\rm F} = \sum_{\bi \in \text{Supp}(\bm \varOmega)}x_{\bi}^2$, where $\text{Supp}(\bm \varOmega)$ denotes the set of all unique entries in $\bm \varOmega$. {$\|\tX\|_{\infty}$ denotes the max norm of $\tX$ such that $\|\tX\|_{\infty} = \underset{\bi}{\max}~[\tX]_{\bi}$.}
$[I]$ means an integer set $\{1,2,\ldots,I\}$. 
  $^\T$ and $^\dag$ denote transpose and pseudo-inverse, respectively. $\bm x\circ\bm y$ denotes the outer product of two vectors $\bm x$ and $\bm y$, i.e., $\bm x\circ\bm y = \bm x \bm y^{\top}$. $\circledast$ denotes the element-wise product (also known as Hadamard product). The function ${\rm sigmoid}(x)$ represents $\frac{1}{1+\exp(-x)} $. The constant $e$ denotes the Euler's number.  $[x]_{[0,1]}$ denotes $ \min\{\max\{x,0\},1\}$.

  \section{MLE Formulation} \label{app:mle}
  Assuming that $y_{\bi}$'s are independently sampled from the generative model in \eqref{eq:nonlinear_model_tensor}, the log-likelihood of the observations can be expressed as follows:
\begin{align}
\log \prod_{\bm i\in \bm \varOmega}{\sf Pr}(Y_{\bi}=y_{\bm i}; \lambda_{\bm i}, p_{\bm i}) 
&= \log\left(\prod_{\bi\in\bm \varOmega} \sum_{n=y_{\bi}}^\infty {\sf Pr}(N_{\bi}=n;\lambda_{\bi}) {\sf Pr}(y_{\bi}|N_{\bi}=n; p_{\bi})\right) \nonumber\\
&= \sum_{\bi\in\bm \varOmega} \log\left( \sum_{n=y_{\bi}}^\infty\left(\frac{\lambda_{\bi}^{n} e^{-\lambda_{\bi}}}{n!} \frac{n! p_{\bi}^{y_{\bi}} (1-p_{\bi})^{n-y_{\bi}}}{y_{\bi}!(n-y_{\bi})!} \right)\right) \label{eq:log_likelihood_1},
\end{align}
where $Y_{\bi}$ denotes the random variable associated with the observation $y_{\bi}$ and $N_{\bi}$ denotes the random variable associated with the true count $n_{\bi}$. To circumvent the infinite sum in the log-likelihood, we invoke the folowing lemma:

\begin{lemma}\citep{dennis2015computational}\label{lem:key}
Assume that the observation $y$ follows the below generative model:
\begin{subequations}\label{eq:poisson_binomial_model}
	\begin{align} 
	n &\sim {\sf Poisson}(\lambda), \label{eq:poisson11}\\
	y&\sim {\sf Binomial}(n,p). \label{eq:binomial1}
	\end{align}
\end{subequations}
	Then, the model in \eqref{eq:poisson_binomial_model} is equivalent to
   $ y\sim  {\sf Poisson}(\lambda p) $.
\end{lemma}
Lemma~\ref{lem:key} states that the Poisson-Binomial model admits an equivalent Poisson model whose Poisson parameter is the multiplication of the original Poisson parameter and the detection probability. The derivation is based on an interesting equality:
\[\sum_{n=y}^\infty \frac{\lambda^{n} e^{-\lambda}}{n!}  \frac{n!p^{y} (1-p)^{n-y}}{y!(n-y)!} =  \frac{(p\lambda)^{y}e^{-\lambda p}}{y!};	\]
{see the derivation in \citep{fu2019link}.}

Applying the above equality in \eqref{eq:log_likelihood_1}, we obtain:
\begin{align}
\log \prod_{\bm i\in \bm \varOmega}{\sf Pr}(Y_{\bi}=y_{\bm i}; \lambda_{\bm i}, p_{\bm i}) &= \sum_{\bm i\in \bm \varOmega} \log \left(\frac{(p_{\bi}\lambda_{\bi})^{y_{\bi}}e^{-\lambda_{\bi} p_{\bi}}}{y_{\bi}!} \right) \nonumber\\
&= \sum_{\bm i\in \bm \varOmega} \left[y_{\bm i}\log \lambda_{\bm i}+y_{\bm i}\log p_{\bm i}-\lambda_{\bm i}p_{\bm i}-\log y_{\bm i}!\right]. 
\end{align}
{Here, the MLE is formulated and derived under the same spirit of the MC case in \citep{fu2019link}.}

  \section{More Details on Algorithm Design} \label{app:algo}
  In this section, we provide more details on the implementation of the proposed algorithm \texttt{UncleTC}. Our formulation is given below:
  \begin{subequations} \label{eq:recast_max_likelihood_optim_tensor11}
	\begin{align}
	\underset{\{\bm U_k\}_{k=1}^K, \bm \theta,  \{p_{\bm i}\}}{\rm minimize}~&\frac{1}{|\bm \varOmega|}\sum_{\bm i \in \bm \varOmega}\Bigg[\bigg( \sum_{f=1}^F  \prod_{k=1}^K \bm U_k(i_k,f)\bigg)p_{\bm i}\Big. 
 	-y_{\bm i}\log \bigg( \sum_{f=1}^F  \prod_{k=1}^K \bm U_k(i_k,f)\bigg)  \Big.-y_{\bm i}\log p_{\bm i}\Bigg] 
+\frac{\mu}{|\bm \varXi|} \sum_{\bm i \in \bm \varXi}\ell(g_{\bm \theta}(\bm z_{\bm i}),p_{\bm i}),\label{eq:objective_fn} \\
	{\rm subject\ to}~~&
	\bm U_k \ge \bm 0, \forall k \in [K], \label{eq:nonnegative_emb}\\
	& 0 \le p_{\bm i} \le 1, ~ \forall \bm i \in \bm \varOmega \cup \bm \varXi,
	\end{align}
\end{subequations}
where $\mu > 0$ is a regularization parameter and $\ell(\cdot,\cdot)$ denotes a certain distance/divergence measure, e.g., the least squares function $\ell(x,y) = (x-y)^2$. 
  \subsection{The $\textbf{\textit{U}}_k$-Subproblem}
The subproblem for each $\bm U_k$ is given below:
  \begin{align} \label{eq:U_subproblem11}
&\underset{\bm U_k \ge \bm 0}{\rm minimize}~ w(\U_k):=\sum_{\bm i \in \bm \varOmega}\left[\bigg(\sum_{f=1}^F \bm U_k (i_k,f) \prod_{j\neq k} \bm U_j(i_j,f)  \bigg) p_{\bm i}-y_{\bm i}\log \bigg(\sum_{f=1}^F \bm U_k (i_k,f) \prod_{j\neq k} \bm U_j(i_j,f)\bigg)\right].
\end{align}

To obtain the update rule for $\bm U_k$ from \eqref{eq:U_subproblem11}, we have the following result:
\begin{lemma} \label{lem:sk_upper_bound}
Let $\overline{\bm U}_k$ denote the current estimate of $\bm U_k$. Then, the objective function in \eqref{eq:U_subproblem11} can be upper-bounded by the following function:
\begin{align}
    &s(\bm U_k; \overline{\bm U}_k) =
    \sum_{\bm i \in \bm \varOmega}\Bigg[\Bigg(\sum_{f=1}^F\bm U_k(i_k,f)  \prod_{j\neq k} \bm U_j(i_j,f)\Bigg) p_{\bm i} - y_{\bm i}\sum_{f=1}^F\alpha_{\bm i}^{(f)}\log \bigg(\frac{\bm U_k (i_k,f) \prod_{j\neq k} \bm U_j(i_j,f)}{\alpha_{\bm i}^{(f)}}\bigg)\Bigg],\label{eq:sfunction}\\
    &\text{where}~~ \alpha_{\bm i}^{(f)} = \frac{\overline{\bm U}_k (i_k,f) \prod_{j\neq k} \bm U_j(i_j,f)}{\sum_{f=1}^F \overline{\bm U}_k (i_k,f) \prod_{j\neq k} \bm U_j(i_j,f) }.\nonumber
\end{align}
\end{lemma}
The construction of the surrogate function $s(\bm U_k; \overline{\bm U}_k)$ follows the idea in \citep{chi2012tensors,fu2019link}; The differences include that the model in \citep{chi2012tensors} did not have $p_{\bi}$ terms and that the work in \citep{fu2019link} did not consider the tensor case. Here, we use the Jensen's inequality to construct a ``tight'' upper bound, i.e.,  $w(\U_k) \leq s(\U_k;\overline{\U}_k)$ {and the equality can be attained at $\U_k=\overline{\bm U}_k$}; see Sec. \ref{app:sk_upper_bound} for the proof of Lemma \ref{lem:sk_upper_bound}. 
To be more precise, the update of $\U_k$ is through solving the following subproblem:
\begin{align}\label{eq:Uk_update}
    \U_k\leftarrow \arg\min_{\U_k\geq \bm 0}~s(\U_k;\overline{\U}_k).
\end{align}
By taking the derivative of $s(\cdot;\overline{\bm U}_k)$ w.r.t. $\bm U_k$ and equating the derivative to zero, we obtain:
\begin{align} \label{eq:update_U}
   \bm U_k(i_k,f) &\leftarrow  \frac{\sum_{\bm i \in \bm \varOmega}y_{\bm i}\alpha_{\bm i}^{(f)}}{\sum_{\bm i \in \bm \varOmega} \prod_{j\neq k} \bm U_j(i_j,f) p_{\bm i}}.
\end{align}
For efficient implementation of the above updates in the presence of missing data, we adopt a strategy proposed in \citep{chi2012tensors} for handling sparse tensors. 

\subsection{The $p$-{Sub}problem} 
Table \ref{tab:pi_updates} summarizes the update rules for {the $p_{\bi}$-subproblem} in Sec. \ref{sec:algorithm} {under} different choices of the {regularization} function $\ell(\cdot)$.
The complete description of the proposed \texttt{UncleTC} algorithm is provided in Algorithm \ref{algo:proposed}. 

\begin{table*}[t]
    \centering
    \caption{Updates for $p_{\bi}$ subproblems for different loss functions}
    \resizebox{0.85\linewidth}{!}{
    \begin{tabular}{l|c|c}
        ~ & $\ell(x,y) = (x-y)^2$ & $\ell(x,y) = x\log \frac{x}{y}-x+y$  \\ 
        \hline
        \hline
        Case 1: $\forall \bi  \in  \bm \varOmega \cap \bm \varXi$ & $p_{\bi} \leftarrow \frac{(2\bar{\mu} g_{\bm \theta}(\bm z_{\bi})-\bar{\lambda}_{\bi}) + \sqrt{(2\bar{\mu} g_{\bm \theta}(\bm z_{\bi})-\bar{\lambda}_{\bi})^2+8\bar{\mu} \bar{y}_{\bi}}}{4\bar{\mu}}$ & $p_{\bi} \leftarrow \frac{\bar{y}_{\bi}+\bar{\mu}g_{\bm \theta}(\bm z_{\bi})}{\bar{\lambda}_{\bi}+\bar{\mu}}$  \\ 
        \hline
        Case 2: $\forall {\bi} \in \bm \varXi - \bm \varOmega \cap \bm \varXi$ & $p_{\bi}  \leftarrow g_{\bm \theta}(\bm z_{\bi})$ &  $p_{\bi}  \leftarrow g_{\bm \theta}(\bm z_{\bi})$ \\ 
        \hline
        Case 3: $\forall \bi \in \bm \varOmega - \bm \varOmega \cap \bm \varXi$ &  $p_{\bi}\leftarrow \left[\nicefrac{y_{\bi}}{\lambda_{\bi}}\right]_{[0,1]}$ &  $p_{\bi}\leftarrow \left[\nicefrac{y_{\bi}}{\lambda_{\bi}}\right]_{[0,1]} $ \\ 
        \hline
        \hline
    \end{tabular}
    \label{tab:pi_updates}
    }\\
    \footnotesize{$\bar{\mu}=\nicefrac{\mu}{|\bm \varXi|}$, $\bar{\lambda}_{\bi}=\nicefrac{\lambda_{\bi}}{|\bm \varOmega|}$, $\bar{y}_{\bi}=\nicefrac{y_{\bi}}{|\bm \varOmega|}$}
\end{table*}

	\begin{algorithm}[t]
	
	\SetKwInOut{Input}{input}
	\SetKwInOut{Output}{output}
	\SetKwRepeat{Repeat}{repeat}{until}
	\Input{Observations $y_{\bm i}, \forall \bm i \in \bm \varOmega$, feature vectors $\bm z_{\bm i}$,~$\forall \bm i \in \bm \varXi$, Initializations  $\bm \theta^{(0)}$, $p_{\bi}^{(0)},~ \forall \bi \in \bm \varXi$, and $\bm U_k^{(0)}$, $\forall k$.}
	

	 $t \leftarrow 0$;
	 
	\Repeat{the stopping criterion is reached}{
	
	 $\lambda^{(t+1)}_{\bm i}\leftarrow\sum_{f=1}^F  \prod_{k=1}^K \bm U^{(t)}_k(i_k,f),~ \forall \bm i$;
	
    $\tilde{p}_{\bm i}^{(t+1)} \leftarrow g_{\bm \theta^{(t)}}(\bm z_{\bm i}),~ \bi \in \bm \varXi$;

	  $p^{(t+1)}_{\bi}\leftarrow{\arg\min}_{p_{\bi}\in[0,1]}~\frac{1}{|\bm \varOmega|}(\lambda^{(t+1)}_{\bm i}p^{(t)}_{\bm i}- y_{\bm i}\log(p^{(t)}_{\bm i}) ) +\frac{\mu}{|\bm \varXi|}  \ell(\tilde{p}_{\bm i}^{(t+1)},p^{(t)}_{\bm i}), ~ \forall \bi \in \bm \varOmega \cap \bm \varXi$;

	 $p^{(t+1)}_{\bi}\leftarrow\arg\min_{p_{\bi}\in[0,1]}\ell(\tilde{p}_{\bm i}^{(t+1)},p^{(t)}_{\bm i}),~ \forall \bi \in \bm \varXi - \bm \varOmega \cap \bm \varXi$;

	 $p^{(t+1)}_{\bi}\leftarrow \left[\nicefrac{y_{\bi}}{\lambda^{(t+1)}_{\bi}}\right]_{[0,1]}, ~\forall \bi \in \bm \varOmega - \bm \varOmega \cap \bm \varXi$;
    
  	$\bm \theta^{(t+1)} \leftarrow \arg\min_{\bm \theta}\sum_{\bm i \in \bm \varXi}\ell(g_{\bm \theta}(\bm z_{\bm i}),p^{(t+1)}_{\bm i})$;  
	
    \For{$k=1$ {\bf to} $K$}{
    
        $r \leftarrow 0$;
        
    	\Repeat{the stopping criterion is reached}{
    	
        $\alpha_{\bi}^{(f)} \leftarrow \frac{{\bm U}^{(r)}_k (i_k,f) \prod_{j\neq k} \bm U_j^{(t)}(i_j,f)}{\sum_{f=1}^F {\bm U}_k^{(r)} (i_k,f) \prod_{j\neq k} \bm U^{(t)}_j(i_j,f) },~\forall \bi,f$;
        
    	 $\bm U_k^{(r+1)}(i_k,f) \leftarrow  \frac{\sum_{\bm i \in \bm \varOmega}y_{\bm i}\alpha_{\bm i}^{(f)}}{\sum_{\bm i \in \bm \varOmega} \prod_{j\neq k} \bm U_j^{(t)}(i_j,f) p^{(t+1)}_{\bm i}},~\forall i_k \in [I_k],f$;

    	  $r \leftarrow r+1$;
    	  
    	 }
    	 $\bm U_k^{(t+1)} \leftarrow \bm U_k^{(r)} $;
    	 
    	$\bm U_k^{(0)} \leftarrow \bm U_k^{(r)} $;
    	
    	 }

	 $t \leftarrow t+1$;
	 }

	 \Output{estimates $\{\widehat{\bm U}_k\}_{k=1}^K$,  $\widehat{\bm \theta}$, and $\widehat{p}_{\bi},~\forall \bi \in \bm \varOmega \cup \bm \varXi$}
	 \caption{ \texttt{UncleTC}}\label{algo:proposed} 


\end{algorithm}


\section{Proof of the Main Result (Theorem \ref{thm:main})} \label{app:main_theorem}

Let us first construct the following constraint sets: 
\begin{subequations}\label{eq:landg}
\begin{align}
    {\cal L}&=\{ \tlambda~|~\tlambda = \sum_{f=1}^F \U_1(:,f)\circ\ldots\circ\U_K(:,f),  \|\U_k(:,f)\|_{2}\leq u,~\U_k\geq \bm 0, \forall k,f \}, \label{eq:l}\\
    {\cal M} &=\left\{\left.\tM=\tlambda \circledast\tP~\right|\beta\leq [\tM]_{\bi}\leq \alpha, \forall \bi, \tlambda\in {\cal L},[\tP]_{\bi}=g_{\bm \theta}(\bm z_{\bi}),~g_{\bm \theta}\in{\cal G}, {\bm z_{\bi}}\in \mathbb{R}^D,~ \forall \bi \in \bm \varXi\right\}.
\end{align}
\end{subequations}

Under Assumption \ref{as:bound}, we can choose the parameters $u$, $\alpha$, and $\beta$ as follows:
{ 
\begin{align} \label{eq:upper_bounds}
    u = \sqrt{F}\alpha_u, ~\alpha = F\alpha_u^K p_{\max},~\beta = F\beta_u^K p_{\min}.
\end{align}
}
Using these notations, the MLE in \eqref{eq:max_likelihood_optim_tensor} can be re-expressed as follows:
\begin{align}\label{eq:Mexpression}
    \widehat{\tM}=\text{arg~min}_{\tM \in {\cal M}}&~\sum_{\bi\in\bm \varOmega} f(m_{\bi};y_{\bi}, \bm z_i),
\end{align}
where 
$f(m_{\bi};y_{\bi},\bm z_{\bi}) = m_{\bi} - y_{\bi} \log m_{\bi}$, where $m_{\bi} = \lambda_{\bi}g_{\bm \theta}(\bm z_{\bi})$.

\paragraph{\large Estimating $\nicefrac{\|\tM^\natural - \widehat{\tM}\|_{\rm F}}{\prod_k I_k}$:}

Our goal is to characterize the estimation accuracies of $\lambda_{\bi}^\natural$ and $p_{\bi}^\natural$. To achieve this goal, we first characterize the term $\|\tM^\natural - \widehat{\tM}\|_{\rm F}$. We have the following result:

\begin{theorem}\label{thm:Merror}
{Suppose that the assumptions of Theorem \ref{thm:main} hold true.}
    Let $T=|\bm \varOmega|$. Assume that $\{y_{\bi}\}_{\bi \in \bm \varOmega}$ are i.i.d. samples under the generative model \eqref{eq:nonlinear_model_tensor} and that $\bm z_{\bi_1},\dots,\bm z_{\bi_S}$ are the set of observed features. Then, under \eqref{eq:Mexpression},  with probability at least $1-2\delta-3e^{-\alpha(e^2-3)}$, we have the following relation, for any $0<\delta < 1$:
    \begin{align} \label{eq:Merror}
 \frac{\|\tM^\natural - \widehat{\tM}\|^2_{\rm F} }{\prod_k I_k} \leq C(\alpha,\beta)\left(2 \mathfrak{R}_{T}({\cal F})  + (5f_{\max}-f_{\min})\sqrt{\frac{2\log(4/\delta)}{T}} + F \alpha_u^K \left(1+\frac{p_{\max}(e^2-2)}{p_{\min}}\right)\nu \right),
    \end{align}
    where $C(\alpha,\beta) = \frac{4\alpha \gamma}{1-e^{-\gamma}}$, $\gamma = \frac{1}{8\beta}(\alpha -\beta)^2$, $f_{\max}   \leq \alpha (1+ (e^2-2)\max\{ |\log \beta|,\log\alpha\}),$ $f_{\min} = \beta -\alpha(e^2-2)\log \alpha$,
    \begin{align}
{\mathfrak{R}}_T &\leq \left( \frac{4}{\sqrt{T}} + \frac{12}{\sqrt{T}} \sqrt{F\sum_k I_k \log\left( 6K L_f\sqrt{T}(F \alpha_u)^{K} \right)  + \left(    {F\alpha^K_uL_f\sqrt{T}\|\bm Z\|_{\rm F}\mathscr{R}_{\mathcal{G}}} \right) }         \right),    \nonumber
\end{align}
 $L_f = 1 + \frac{\alpha(e^2-2)}{\beta}$, $\bm Z=[\bm z_{\bi_1},\dots, \bm z_{\bi_S}] \in \mathbb{R}^{D \times S}$, and {$\mathscr{R}_{\mathcal{G}}$ denotes the complexity measure of the class ${\cal G}$.}
\end{theorem}
The proof is relegated to Appendix \ref{app:thm_M_error}.

\paragraph{\large Estimating $\nicefrac{\|[\tM^\natural - \widehat{\tM}]_{\bm \varTheta}\|_{\rm F}}{|\bm \varTheta|}$:}

Next, we proceed to estimate $\nicefrac{\|[\tM^\natural - \widehat{\tM}]_{\bm \varTheta}\|_{\rm F}}{|\bm \varTheta|}$ using the result in Theorem \ref{thm:Merror}.

Consider the following result:
\begin{lemma}\label{lem:gap}
     {Assume that $\bi\in \bm \varTheta$ is drawn from $[I_1]\times \ldots \times [I_K]$ uniformly at random (without replacement)}. {Consider that $\tM_1,\tM_2\in \{\tM~|~\tM\in \mathbb{R}_+^{I_1 \times \dots \times I_K},\|\tM\|_{\infty}\leq \alpha\}$ holds. Let us define}
    \begin{equation}\label{eq:tau}
    \begin{aligned}
     \tau(\bm \varTheta; \tM_1,\tM_2) =
      &  \left|  \frac{1}{\sqrt{|\bm \varTheta|}} \left\| \left( \tM_1 -\tM_2\right)_{\bm \varTheta}   \right\|_{\rm F} - \frac{1}{\sqrt{I^K}}  \left\| \tM_1 -\tM_2  \right\|_{\rm F}  \right|. \nonumber
    \end{aligned}
    \end{equation}
     Then, with probability greater than $1-\delta$, we have
    \begin{align*}
        \tau(\bm \varTheta; \tM_1,\tM_2) &\leq   \alpha\left(\log\left(\frac{2}{\delta}\right) \left( 1 - \frac{(|\bm \varTheta|-1)}{{\prod_k I_k}}  \right) \frac{1}{2|\bm \varTheta|}    \right)^{1/4}.
    \end{align*}
\end{lemma}
The proof is provided in Appendix \ref{app:lem_gap}.

Combining Lemma \ref{lem:gap} (here, $\alpha = F\alpha_u^K p_{\max}$ using Assumption \ref{as:bound}) with Theorem~\ref{thm:Merror}  and denoting $\eta = \frac{1}{\sqrt{\prod_k I_k}}  \left\| \tM^\natural - \widehat{\tM}  \right\|_{\rm F}$, we have the following with probability at least $1-3\delta-3e^{-\alpha(e^2-3)}$:
\begin{align} \label{eq:MestimerrorTheta}
    \frac{\left\| \left[   \tM^\natural - \widehat{\tM} \right]_{\bm \varTheta}\right\|_{\rm F}}{\sqrt{|\bm \varTheta|}} \leq \eta  + \tau(\bm \varTheta; \tM^\natural,\widehat{\tM}),
\end{align}
where 
$\eta^2$ is given by \eqref{eq:Merror} from Theorem \ref{thm:Merror}.

\paragraph{\large Recovering $\lambda_{\bi}$'s:}
Next, we proceed to characterize the estimation accuracies of $\lambda_{\bi}$'s using the result in \eqref{eq:MestimerrorTheta}. Under Assumptions~\ref{as:similar_z} and \ref{as:countinuous_g}, we have the following representation:
\begin{align*} 
    p^\natural_{\bi} &= g^\natural(\z_{\bi}) = \xi + s_{\bi},~ |s_{\bi}| \le \zeta L_{g} , ~\forall \bi\in \bm \varTheta,\\
    \widehat{p}_{\bi} &= \widehat{g}_{\bm \theta}(\z_{\bi}) = {\xi}' + {s}'_{\bi},~ |{s}'_{\bi}| \le \zeta L_{\bm \theta} , ~\forall \bi\in \bm \varTheta ,
\end{align*}
where $0\le \xi,\xi' \le 1$ are certain scalars.

Hence, we get the following relation, with probability at least $1-3\delta-3e^{-\alpha(e^2-3)}$:
\begin{align*}
     (\eta  +\tau(\bm \varTheta; \tM^\natural,\widehat{\tM}) )\sqrt{|\bm \varTheta|}&\geq \left\| \left[   \tM^\natural - \widehat{\tM} \right]_{\bm \varTheta}\right\|_{\rm F}= \sqrt{\sum_{\bi \in \bm \varTheta} (m^\natural_{\bi}-\widehat{m}_{\bi})^2}\\
     &= \sqrt{\sum_{\bi \in \bm \varTheta} (\lambda^\natural_{\bi}p^\natural_{\bi}-\widehat{\lambda}_{\bi}\widehat{p}_{\bi})^2}= \sqrt{\sum_{\bi \in \bm \varTheta} (\lambda^\natural_{\bi}(\xi+s_{\bi})-\widehat{\lambda}_{\bi}(\xi'+s'_{\bi}))^2}\\
        &= \sqrt{\sum_{\bi \in \bm \varTheta} (\xi\lambda^\natural_{\bi}-\xi'\widehat{\lambda}_{\bi}+s_{\bi}\lambda^\natural_{\bi}-s'_{\bi}\widehat{\lambda}_{\bi})^2}\\
         &\ge \left|\sqrt{\sum_{\bi \in \bm \varTheta} (\xi\lambda^\natural_{\bi}-\xi'\widehat{\lambda}_{\bi})^2}- \sqrt{\sum_{\bi \in \bm \varTheta} (s_{\bi}\lambda^\natural_{\bi}-s'_{\bi}\widehat{\lambda}_{\bi})^2}\right|.
\end{align*}
Since $| s_{\bi}| \leq \zeta L_g$, $| s'_{\bi}| \leq \zeta L_{\bm \theta}$, and $\lambda_{\bi}  \le F\alpha_u^K$, $\forall \tlambda \in {\cal L}$ by \eqref{eq:upper_bounds}, the above relation can be further expressed as
\begin{align}
      (\eta +\tau(\bm \varTheta; \tM^\natural,\widehat{\tM})  )\sqrt{|\bm \varTheta|}+ \zeta(L_g+L_{\theta})F\alpha_u^K\sqrt{|\bm \varTheta|}  
     &\geq \xi\sqrt{\sum_{\bi \in \bm \varTheta} (\lambda^\natural_{\bi}-\nicefrac{\xi'}{\xi}\widehat{\lambda}_{\bi})^2} \nonumber\\
    & = \xi \left\| \left[   \tlambda^\natural - \nicefrac{\xi'}{\xi} \widehat{\tlambda} \right]_{\bm \varTheta}\right\|_{\rm F} \nonumber\\
    \Longrightarrow \frac{\left\| \left[   \tlambda^\natural - \nicefrac{\xi'}{\xi} \widehat{\tlambda} \right]_{\bm \varTheta}\right\|_{\rm F} }{ \sqrt{|\bm \varTheta| } } &\leq \frac{ (\eta +\tau(\bm \varTheta; \tM^\natural,\widehat{\tM})  ) + \zeta(L_g+L_{\theta})F\alpha_u^K}{\xi} \nonumber\\
    &\leq \frac{ (\eta +\tau(\bm \varTheta; \tM^\natural,\widehat{\tM})  ) + \zeta(L_g+L_{\theta})F\alpha_u^K}{p_{\min}}.
    \label{eq:lambda_acc} 
\end{align}
Let $\xi'/\xi = \widehat{\xi}$. 
Our next goal is to characterize $\left\|   \tlambda^\natural - \widehat{\xi} \widehat{\tlambda}\right\|_{\rm F}$ using the result in \eqref{eq:lambda_acc}. 
To proceed, consider the term
\begin{align} \label{eq:tcalLq}
     \tau(\bm \varTheta; \tlambda^\natural, \widehat{\xi} \widehat{\tlambda} ) =\left|\frac{\left\| \left[  \tlambda^\natural - \widehat{\xi} \widehat{\tlambda} \right]_{\bm \varTheta}\right\|_{\rm F} }{\sqrt{|\bm \varTheta|}} -   \frac{\left\|  \tlambda^\natural - \widehat{\xi} \widehat{\tlambda} \right\|_{\rm F} }{\sqrt{{\prod_k I_k}}}\right| .
\end{align}
We again invoke Lemma~\ref{lem:gap} for characterizing the term $\tau(\bm \varTheta; \tlambda^\natural, \widehat{\xi} \widehat{\tlambda} )$, and obtain that with probability of at least $1-\delta$,
\begin{align}\label{eq:Lq}
    \tau(\bm \varTheta; \tlambda^\natural, \widehat{\xi} \widehat{\tlambda} )&\leq   F\alpha_u^{K}\left(\log\left(\frac{2}{\delta}\right) \left( 1 - \frac{(|\bm \varTheta|-1)}{{\prod_k I_k}}  \right) \frac{1}{2|\bm \varTheta|}    \right)^{1/4}.
\end{align}

Combining the result in \eqref{eq:tcalLq} and \eqref{eq:Lq} with \eqref{eq:lambda_acc}, we obtain the following result with probability at least $1-4\delta-3e^{-\alpha(e^2-3)}$:
\[  \frac{\left\|  \tlambda^\natural - \widehat{\xi} \widehat{\tlambda} \right\|_{\rm F} }{\sqrt{{\prod_k I_k}}}\leq  \underbrace{\tau(\bm \varTheta; \tlambda^\natural, \widehat{\xi} \widehat{\tlambda} ) +  \frac{ (\eta +\tau(\bm \varTheta; \tM^\natural, \widehat{\tM} )  ) + \zeta(L_g+L_{\theta})F\alpha_u^K}{p_{\min}}}_{\eta'} .\]

\paragraph{\large Recovering $p_{\bi}$'s:}
Next, we proceed to bound the estimation accuracies for $p_{\bi}$'s. Consider the following chain of inequalities:
\begin{align*}
   \|\tM^\natural - \widehat{\tM}\|_{\rm F}  &= \left\|  \tlambda^\natural \circledast\tP^\natural - \widehat{\xi} \widehat{\tlambda} \circledast \nicefrac{1}{\widehat{\xi}}\widehat{\tP} \right\|_{\rm F}\\
&=\left\|  \tlambda^\natural \circledast \tP^\natural - {\tlambda}^\natural\circledast\nicefrac{1}{\widehat{\xi}}\widehat{\tP} + {\tlambda}^\natural\circledast\nicefrac{1}{\widehat{\xi}}\widehat{\tP}- \widehat{\xi} \widehat{\tlambda} \circledast \nicefrac{1}{\widehat{\xi}}\widehat{\tP} \right\|_{\rm F}\\
&\geq \left|\left\|  \tlambda^\natural \circledast (\tP^\natural - \circledast\nicefrac{1}{\widehat{\xi}}\widehat{\tP}) \right\|_{\rm F} - \|(\tlambda^\natural - \widehat{\xi} \widehat{\tlambda})\circledast\nicefrac{1}{\widehat{\xi}}\widehat{\tP}\|_{\rm F}\right|\\
&\geq \lambda_{\min}\left\|  \tP^\natural - \nicefrac{1}{\widehat{\xi}}\widehat{\tP} \right\|_{\rm F} - \|(\tlambda^\natural - \widehat{\xi} \widehat{\tlambda})\circledast\nicefrac{1}{\widehat{\xi}}\widehat{\tP}\|_{\rm F}\\
&\geq \frac{\beta}{p_{\max}}\left\|  \tP^\natural - \nicefrac{1}{\widehat{\xi}}\widehat{\tP} \right\|_{\rm F} - \|(\tlambda^\natural - \widehat{\xi} \widehat{\tlambda})\circledast\nicefrac{1}{\widehat{\xi}}\widehat{\tP}\|_{\rm F},
\end{align*}
where $\lambda_{\min}=\min_{\bi}\lambda_{\bi} = \frac{\min_{\bi}m_{\bi}}{ \max_{\bi}p_{\bi}} = \frac{\beta}{p_{\max}}$.
The above inequalities imply that with probability at least $1-4\delta-3e^{-\alpha(e^2-3)}$
\begin{align}
\frac{\left\|   \tP^\natural - \nicefrac{1}{\widehat{\xi}}\widehat{\tP} \right\|_{\rm F}}{\sqrt{\prod_k I_k}} &\le     \frac{p_{\max}}{\beta}\left(\frac{\left\|  \tlambda^\natural - \widehat{\xi} \widehat{\tlambda} \right\|_{\rm F} }{p_{\min}\sqrt{\prod_k I_k}}+  \frac{\|\tM^\natural - \widehat{\tM}\|_{\rm F} }{\sqrt{\prod_k I_k}}\right), \nonumber\\
& = \frac{p_{\max}}{\beta}\left(\frac{\eta' }{p_{\min}}+  \eta\right),\label{eq:Perror}
\end{align}
where we have used the result that $\|(\tlambda^\natural - \widehat{\xi} \widehat{\tlambda})\circledast\nicefrac{1}{\widehat{\xi}}\widehat{\tP}\|_{\rm F} \le \frac{1}{p_{\min}}\left\|  \tlambda - \widehat{\xi} \widehat{\tlambda} \right\|_{\rm F}$,
since $[\nicefrac{1}{\widehat{\xi}\widehat{\tP}}]_{\bi} = \nicefrac{1}{\widehat{\xi}} \widehat{p}_{\bi} = \nicefrac{\xi}{\xi'}\widehat{p}_{\bi} \le \nicefrac{1}{p_{\min}} $.

Next, we proceed to characterize the generalization error of the learned nonlinear function $\widehat{g}_{\bm \theta}$. Towards this, we first bound the term $\frac{\left\|   [\tP^\natural - \nicefrac{1}{\widehat{\xi}}\widehat{\tP}]_{\bXi} \right\|_{\rm F}}{\sqrt{S}}$ using the result in \eqref{eq:Perror}. Let us consider
\begin{align*}
\tau(\bXi; \tP^\natural,\nicefrac{1}{\widehat{\xi}} \widehat{\tP}) = \left|\frac{\left\| \left[  \tP^\natural - \nicefrac{1}{\widehat{\xi}} \widehat{\tP} \right]_{\bXi}\right\|_{\rm F} }{\sqrt{S}} -   \frac{\left\|  \tP^\natural - \nicefrac{1}{\widehat{\xi}} \widehat{\tP} \right\|_{\rm F} }{\sqrt{{\prod_k I_k}}}\right| .
\end{align*}

By applying Lemma \ref{lem:gap} combined with \eqref{eq:Perror}, we get that with a probability greater than  $1-5\delta-3e^{-\alpha(e^2-3)}$
\begin{align} \label{eq:Pbound_Xi}
\frac{\left\| \left[  \tP^\natural - \nicefrac{1}{\widehat{\xi}} \widehat{\tP} \right]_{\bXi}\right\|_{\rm F} }{\sqrt{S}}  &\le \tau(\bXi; \tP^\natural,\nicefrac{1}{\widehat{\xi}} \widehat{\tP}) + \frac{p_{\max}}{\beta}\left(\frac{\eta' }{p_{\min}}+  \eta\right),
  \end{align}
  where 
\begin{align*}
    \tau(\bXi; \tP^\natural,\nicefrac{1}{\widehat{\xi}} \widehat{\tP})&\leq   p_{\max}\left(\log\left(\frac{2}{\delta}\right) \left( 1 - \frac{(S-1)}{{\prod_k I_k}}  \right) \frac{1}{2S}    \right)^{1/4}.
\end{align*}

Using the result in \eqref{eq:Pbound_Xi}, we proceed to bound the generalization performance of the learned nonlinear function $\widehat{g}_{\bm \theta}$. Towards this goal, consider the following result:
	\begin{lemma} \label{lem:generalization_g}
	    Under the Assumptions given by Theorem \ref{thm:main}, the following holds with probability greater than $1-\delta$:
	    \begin{align}
	      \mathbb{E}_{\bm z \sim \mathcal{D}} \left[(g^\natural(\bm z)-\nicefrac{1}{\widehat{\xi}}\widehat{g}_{\bm \theta}(\bm z))^2\right] &\le \frac{\left\| \left[  \tP^\natural - \nicefrac{1}{\widehat{\xi}} \widehat{\tP} \right]_{\bXi}\right\|^2_{\rm F} }{{S}} + \frac{32 \left(2 \|\bm Z\|_{\rm F}\mathscr{R}_{\mathcal{G}}\right)^{\frac{1}{4}}}{S^{5/8}}+4 \sqrt{\frac{2\log(4/\delta)}{S}},\label{eq:gen_f1}  
	    \end{align}
     where $\bm Z = [\bm z_{\bi_1},\dots, \bm z_{\bi_S}]$ and $\mathscr{R}_{\mathcal{G}}$ denotes the complexity measure as given by Assumption \ref{as:approx}.
	\end{lemma}
	The proof is given in Appendix \ref{app:generalization_g}.
 
Combining Lemma \ref{lem:generalization_g} with \eqref{eq:Pbound_Xi}, we get the final bound for $\mathbb{E}_{\bm z \sim \mathcal{D}} \left[(g^\natural(\bm z)-\nicefrac{1}{\widehat{\xi}}\widehat{g}_{\bm \theta}(\bm z))^2\right]$ with probability greater than $1-6\delta-3e^{-\alpha(e^2-3)}$.

\color{black}

\section{Proof of Theorem \ref{thm:Merror}}\label{app:thm_M_error}
 For a set $\bm \varOmega=\{\bi_1,\ldots,\bi_T\}$, we assume that $\bm \varOmega\sim { \Pi}$ where ${\Pi}$ is uniform, i.e., each $y_{\bi}$ is observed independently at random with probability $1/(\prod_k I_k)$ \footnote{Here, the set $\bm \varOmega$ can be a multiset, meaning it can contain multiple copies of an index since the the assumed sampling scheme is with replacement. }.
Hence, we define the expected risk as follows:
\begin{align}
    D_{\Pi}(\tM;\tY):=\mathbb{E}_{\bm \varOmega \sim \Pi} \left[f(m_{\bi};y_{\bi}, \bm z_{\bi}) \right] = \sum_{\bi} \frac{1}{\prod_k I_k}f(m_{\bi};y_{\bi}, \bm z_{\bi}).
\end{align}

Accordingly, we define the empirical risk as follows:
\begin{align}
    D_{\bm \varOmega}(\tM;\tY) = \frac{1}{T} \sum_{\bi \in \bm \varOmega} f(m_{\bi};y_{\bi},\bm z_{\bi}).
\end{align}
Eq.\eqref{eq:Mexpression} implies that 
\begin{align} \label{eq:optim}
    D_{\bm \varOmega}(\widehat{\tM};\tY) \le D_{\bm \varOmega}(\widetilde{\tM};\tY)
\end{align}
where $[\widetilde{\tM}]_{\bi}=\lambda_{\bi}^\natural \widetilde{g}_{\bm \theta}(\bm z_{\bi})$ and $\widetilde{g}_{\bm \theta}$ is given by Assumption \ref{as:approx}.

To proceed, consider the following chain of inequalities:
	\begin{align}
	\mathbb{E}_{y}[D_{\Pi}(\widehat{\tM}; \tY)-D_{\Pi}({\tM}^\natural; \tY)] &= \mathbb{E}_y[D_{\Pi}(\widehat{\tM}; \tY)]-D_{{\bm \varOmega}}(\widehat{\tM}; \tY)+D_{{\bm \varOmega}}(\tM^\natural; \tY)-\mathbb{E}_y[D_{\Pi}(\tM^\natural; \tY)]\nonumber\\
	&\quad +D_{{\bm \varOmega}}(\widehat{\tM}; \tY)-D_{{\bm \varOmega}}(\widetilde{\tM}; \tY)+D_{{\bm \varOmega}}(\widetilde{\tM}; \tY)-D_{\bm \varOmega}(\tM^\natural; \tY) \nonumber\\
	&\le \mathbb{E}_y[D_{\Pi}(\widehat{\tM}; \tY)]-D_{{\bm \varOmega}}(\widehat{\tM}; \tY)+D_{{\bm \varOmega}}(\tM^\natural; \tY)-\mathbb{E}_y[D_{\Pi}(\tM^\natural; \tY)]\nonumber\\
 &\quad + |D_{{\bm \varOmega}}(\widetilde{\tM}; \tY)-D_{\bm \varOmega}(\tM^\natural; \tY)|\nonumber \\
	&\le \underset{\tM \in \mathcal{M}}{\text{sup}}\left | D_{{\bm \varOmega}}({\tM}; \tY)-\mathbb{E}_y[D_{\Pi}({\tM}; \tY)]\right|+  | D_{{\bm \varOmega}}(\tM^\natural; \tY)-\mathbb{E}_y[D_{\Pi}(\tM^\natural; \tY)]| \nonumber \\
 &\quad + |D_{{\bm \varOmega}}(\widetilde{\tM}; \tY)-D_{\bm \varOmega}(\tM^\natural; \tY)|\label{eq:expD},
	\end{align}
where expectation is taken w.r.t. $y_{\bi}$'s and the first inequality is by \eqref{eq:optim}.

 Let us consider the L.H.S. of \eqref{eq:expD}:
\begin{align}
   \mathbb{E}_y[D_{\Pi}(\widehat{\tM}; \tY)-D_{\Pi}({\tM}^\natural; \tY)]
    &= \frac{1}{\prod_k I_k}\sum_{\bi}\mathbb{E}_y\left[(\widehat{m}_{\bi} - y_{\bi} \log \widehat{m}_{\bi}) -  (m^\natural_{\bi} - y_{\bi} \log m^\natural_{\bi})\right]\nonumber\\
    &\quad= \frac{1}{\prod_k I_k}\sum_{\bi}\left[(\widehat{m}_{\bi} - m^\natural_{\bi} \log \widehat{m}_{\bi}) -  (m^\natural_{\bi} - m^\natural_{\bi} \log m^\natural_{\bi})\right]\nonumber\\
        &\quad= \frac{1}{\prod_k I_k}\sum_{\bi}\left[m^\natural_{\bi} \log \frac{m^\natural_{\bi}}{\widehat{m}_{\bi}} -(m^\natural_{\bi}-\widehat{m}_{\bi})\right]\nonumber\\
    &\quad := {\sf KL}(\tM^\natural||\widehat{\tM}), \label{eq:kldiv}
\end{align}
where the second equality utilizes Lemma \ref{lem:key} to get $\mathbb{E}[y_{\bi}] = \lambda^\natural_{\bi}p^\natural_{\bi}=m^\natural_{\bi}.$

\paragraph{Upper-bounding the first term on the R.H.S of \eqref{eq:expD}.} 
Next, we characterize the first term on the R.H.S. of \eqref{eq:expD}. Towards this, we invoke the following theorem (Theorem 26.5 in \citep{shalev2014understanding}) :
	
	\begin{Theorem} \citep[Theorem 26.5]{shalev2014understanding} \label{thm:gen}
		Assume that for all $y$ and for all $x$, we have $|f(x;y)| \le f_{\max}$. Then for any $\tM \in \mathcal{M}$, the following holds with probability greater than $1-\delta$:
		\begin{align} \label{eq:DPiminusDS}
	\left| D_{{\bm \varOmega}}({\tM}; \tY)-\mathbb{E}_y[D_{\Pi}({\tM}; \tY)]\right| \le 2 \mathfrak{R}_{T}({\cal F}) + 4f_{\max} \sqrt{\frac{2\log(4/\delta)}{T}},
		\end{align}
		where  ${\cal F}$ denotes the set
		\begin{align*}
		{\cal F} \triangleq \left\{[f(m_{\bi_1};y_{\bi_1},\bm z_{\bi_1}),\ldots,f(m_{\bi_T};y_{\bi_T},\bm z_{\bi_T})]^\T~|~ \tM\in \mathcal{M}, \bi_t \in \bOmega\right\} 
		\end{align*}
		and 
		$\mathfrak{R}_{T}({\cal F})$ denotes the empirical Rademacher complexity of the set ${\cal F}$.
	\end{Theorem}

Applying Theorem~\ref{thm:gen}, with probability greater than $1-\delta$, we have
\begin{align}\label{eq:boundfirst}
    \underset{\tM\in \mathcal{M}}{\text{sup}}\left | D_{{\bm \varOmega}}({\tM}; \tY)-\mathbb{E}_y[D_{\Pi}({\tM}; \tY)]\right|\leq 2 \mathfrak{R}_{T}({\cal F})  + 4 f_{\max}\sqrt{\frac{2\log(4/\delta)}{T}}.
\end{align}
Let us  characterize $f_{\max}$ in \eqref{eq:boundfirst}. 
\begin{lemma} \label{lem:fmax}
Let $f(m_{\bi};y_{\bi},\bm z_{\bi})= m_{\bi}-y_{\bi}\log m_{\bi}$ where $m_{\bi}=\lambda_{\bi}p_{\bi}=\lambda_{\bi}g_{\bm \theta}(\bm z_{\bi})$. Then, under Assumption \ref{as:bound}, the maximum and minimum value taken by the function $f(m_{\bi};y_{\bi},\bm z_{\bi})$, denoted as $f_{\max}$ and $f_{\min}$ are given by
\begin{align*}
f_{\max}&= \alpha + c\max\{ |\log \beta|,\log\alpha  \}\\
f_{\min}  &=\beta- c\log\alpha,
\end{align*}
with probability greater than $1-e^{-\alpha(e^2-3)}$ where $c = \alpha(e^2-2)$.
\end{lemma}

The proof of the lemma is given in Appendix \ref{app:lem_fmax}.

 Next, we aim to characterize the Rademacher complexity $\mathfrak{R}_{T}({\cal F})$. Towards this, we have the following result:
\begin{lemma}\label{lem:rad_F}
{Suppose that the assumptions of Theorem \ref{thm:main} hold true.} Let $\bm \varOmega = \{\bi_1,\dots,\bi_T\}$ and $T=|\bm \varOmega|$. Assume that the observations $y_{\bi_t}$'s are i.i.d. samples under the generative model  \eqref{eq:nonlinear_model_tensor} and that $\bm z_{\bi_1},\dots,\bm z_{\bi_S}$ are the set of observed features. Consider the set
\[ {\cal F} = \{\bm f \in \mathbb{R}^T|~\bm f =[f(m_{\bi_1};y_{\bi_1}, \bm z_{\bi_1}),\ldots,f(m_{\bi_T};y_{\bi_T}, \bm z_{\bi_T})]^\T |~ m_{\bi_t} = [\tM]_{\bi_{t}}, \bi_t \in \bOmega,~ \tM \in {\cal M}\}. \]
With probability greater than $1-e^{-\alpha(e^2-3)}$, the empirical Rademacher complexity of $\cal{F}$ is bounded by
\begin{align}
&{\mathfrak{R}}_T({\cal F}) \leq  \left( \frac{4}{\sqrt{T}} + \frac{12}{\sqrt{T}} \sqrt{F\sum_k I_k \log\left( 6K L_f\sqrt{T}(F \alpha_u)^{K} \right)  + \left(    {F\alpha^K_uL_f\sqrt{T}\|\bm Z\|_{\rm F}\mathscr{R}_{\mathcal{G}}} \right) }         \right),     \nonumber
\end{align}
where $L_f = 1 + \frac{\alpha(e^2-2)}{\beta}$, $\bm Z=[\bm z_{\bi_1},\dots, \bm z_{\bi_S}] \in \mathbb{R}^{D \times S}$ and
    and {$\mathscr{R}_{\mathcal{G}}$ denotes the complexity measure of the class ${\cal G}$.}
\end{lemma}
The proof is provided in Appendix \ref{app:lem_rad_F}. 
By combining Lemma \ref{lem:fmax} and Lemma \ref{lem:rad_F}, we completely characterize \eqref{eq:boundfirst}.

	\paragraph{Upper-bounding the second term on the R.H.S of \eqref{eq:expD}.}
	Next, we upper bound the second term on the R.H.S of \eqref{eq:expD}. Let us consider the Hoeffding's inequality
\begin{lemma} \label{lem:Hoeffding}
Let $F_1,\dots,F_T$ be independent bounded random variables with $F_t \in [f_{\min},f_{\max}]$ for all $t$ where $-\infty < f_{\min} \le f_{\max} < \infty$. Then for all $t \ge 0$,
\begin{align*}
    {\sf Pr}\left( \frac{1}{T}\sum_{t=1}^T (F_t-\mathbb{E}[F_t]) \ge q  \right) \le \exp\left(-\frac{2Tq^2}{(f_{\max}-f_{\min})^2} \right).
\end{align*}
\end{lemma}
To use Lemma \ref{lem:Hoeffding}, let us define the random variable $F_t$ as follows:
\begin{align*}
    F_t &\triangleq f(m_{\bi_t};y_{\bi_t}, \bm z_{\bi_t}).
\end{align*}
 Then, invoking Lemma \ref{lem:Hoeffding}, one can obtain
\begin{align}
    {\sf Pr}\left( \left | D_{{\bm \varOmega}}(\tM^\natural; \tY)-\mathbb{E}_y[D_{\Pi}(\tM^\natural; \tY)]\right| \ge q  \right) \le \exp\left(-\frac{2Tq^2}{(f_{\max}-f_{\min})^2} \right). \label{eq:hoeffding}
\end{align}
where $f_{\max}$ and $f_{\min}$ are given by Lemma \ref{lem:fmax}.
Hence, by substituting $q = (f_{\max}-f_{\min})\sqrt{\frac{\log\left(\frac{1}{\delta}\right)}{2T}}$, where $\delta \in (0,1)$ in \eqref{eq:hoeffding}, we get that with probability greater than $1-\delta$
\begin{align}
    \left | D_{{\bm \varOmega}}(\tM^\natural; \tY)-\mathbb{E}_y[D_{\Pi}(\tM^\natural; \tY)]\right| \le (f_{\max}-f_{\min})\sqrt{\frac{\log\left(\frac{1}{\delta}\right)}{2T}}.\label{eq:DSiminusDP}
\end{align}

\paragraph{Upper-bounding the third term on the R.H.S of \eqref{eq:expD}.}

Consider the following chain of equations:
\begin{align}
   \left|D_{{\bm \varOmega}}(\widetilde{\tM}; \tY)-D_{\bm \varOmega}(\tM^\natural; \tY)\right| &= \left|\frac{1}{T} \sum_{\bi \in \bm \varOmega} f(\widetilde{m}_{\bi};y_{\bi},\bm z_{\bi})-\frac{1}{T} \sum_{\bi \in \bm \varOmega} f(m^\natural_{\bi};y_{\bi},\bm z_{\bi})\right| \nonumber\\
   &= \frac{1}{T}\left| \sum_{\bi \in \bm \varOmega} (\widetilde{m}_{\bi}-y_{\bi}\log\widetilde{m}_{\bi})-({m}^\natural_{\bi}-y_{\bi}\log{m}^\natural_{\bi})\right| \nonumber\\
   &= \frac{1}{T}\left| \sum_{\bi \in \bm \varOmega} (\lambda^\natural_{\bi}(\widetilde{g}_{\bm \theta}(\bm z_{\bi})-{g}^\natural_{\bm \theta}(\bm z_{\bi}))-y_{\bi}\left(\log\lambda^\natural_{\bi}\widetilde{g}_{\bm \theta}(\bm z_{\bi})-\log\lambda^\natural_{\bi}{g}^\natural_{\bm \theta}(\bm z_{\bi})\right)\right| \nonumber\\
   &=\frac{1}{T}\left| \sum_{\bi \in \bm \varOmega} (\lambda^\natural_{\bi}(\widetilde{g}_{\bm \theta}(\bm z_{\bi})-{g}^\natural_{\bm \theta}(\bm z_{\bi}))-y_{\bi}\left(\log\widetilde{g}_{\bm \theta}(\bm z_{\bi})-\log{g}^\natural_{\bm \theta}(\bm z_{\bi})\right)\right|\nonumber\\
   &\le \max_{\bi}\lambda^\natural_{\bi} |\widetilde{g}_{\bm \theta}(\bm z_{\bi})-{g}^\natural_{\bm \theta}(\bm z_{\bi})|+ \max_{\bi}y_{\bi}\frac{|\widetilde{g}_{\bm \theta}(\bm z_{\bi})-{g}^\natural_{\bm \theta}(\bm z_{\bi})|}{p_{\min}},\nonumber\\
   &\le (c_{\lambda}+c/p_{\min})\nu,\nonumber\\
   &= F \alpha_u^K (1+p_{\max}(e^2-2)/p_{\min})\nu,\label{eq:boundthird}
\end{align}
where the first inequality employs the triangle inequality and the Lipschitz continuity of the log function.  The first inequality also employs the Assumptions \ref{as:bound} that the lowerbound of both $\widetilde{g}_{\bm \theta}(\bm z_{\bi})$ and ${g}^\natural_{\bm \theta}(\bm z_{\bi})$ are $p_{\min}$.
The second inequality is obtained by applying  $c_{\lambda}= F\alpha_u^K$ from Assumption \ref{as:bound}, $c = \alpha +\alpha(e^2-3) $ from Lemma \ref{lem:poissontail} with probability greater than $1-e^{-\alpha(e^2-3)}$ and $\alpha = F\alpha_u^Kp_{\max}$.

\paragraph{Putting Together.}

Combining \eqref{eq:expD}, \eqref{eq:kldiv} with the upperbounds \eqref{eq:boundfirst}, \eqref{eq:DSiminusDP}, and \eqref{eq:boundthird}, we get with probability greater than $1-2\delta-3e^{-\alpha(e^2-3)}$.
 \begin{align}\label{eq:dkl_inequality1}
     {\sf KL}(\tM^\natural||\widehat{\tM}) \le 2 \mathfrak{R}_{T}({\cal F})  + (5f_{\max}-f_{\min})\sqrt{\frac{2\log(4/\delta)}{T}} + F \alpha_u^K \left(1+\frac{p_{\max}(e^2-2)}{p_{\min}}\right)\nu,
 \end{align}
 where $f_{\max} =\alpha + c\max\{ |\log \beta|,\log\alpha\}  $ and $f_{\min} =\beta- c\log\alpha$ given by Lemma \ref{lem:fmax} .

By \citep[Lemma 8]{cao2015poisson}, we have
\begin{align} \label{eq:dkl_inequality2}
     {\sf KL}(\tM^\natural||\widehat{\tM})  \geq  \frac{\|\tM^\natural - \widehat{\tM}\|^2_{\rm F} }{C(\alpha,\beta)\prod_k I_k},
\end{align}
where $C(\alpha,\beta) = \frac{4\alpha \gamma}{1-e^{-\gamma}}$ and $\gamma = \frac{1}{8\beta}(\alpha -\beta)^2$.
Hence, combining \eqref{eq:dkl_inequality1} and \eqref{eq:dkl_inequality2}, we have the following with probability greater than $1-2\delta-3e^{-\alpha(e^2-3)}$:
\begin{align} \label{eq:M_error1}
    \frac{\|\tM^\natural - \widehat{\tM}\|^2_{\rm F} }{\prod_k I_k} \leq C(\alpha,\beta)\left(2 \mathfrak{R}_{T}({\cal F})  + (5f_{\max}-f_{\min})\sqrt{\frac{2\log(4/\delta)}{T}} + F \alpha_u^K \left(1+\frac{p_{\max}(e^2-2)}{p_{\min}}\right)\nu \right).
\end{align}

\section{Proof of Lemma \ref{lem:rad_F}} \label{app:lem_rad_F}
Let us consider the following vector $\bm f$:
\[  \bm f =[f(m_{\bi_1};y_{\bi_1}, \bm z_{\bi_1}),\ldots,f(m_{\bi_T};y_{\bi_T}, \bm z_{\bi_T})]^\T \in {\cal F}, \]\
where $i_t \in \bm \varOmega, ~\forall t$ and 
\begin{align} \label{eq:setF}
    {\cal F} = \{\bm f \in \mathbb{R}^T|~\bm f =[f(m_{\bi_1};y_{\bi_1}, \bm z_{\bi_1}),\ldots,f(m_{\bi_T};y_{\bi_T}, \bm z_{\bi_T})]^\T |~ m_{\bi_t} = [\tM]_{\bi_{t}}, ~ \tM \in {\cal M}\}.
\end{align}

The Rademacher complexity of set ${\cal F}$, denoted as ${\mathfrak{R}}_{T}({\mathcal {F}})$, is defined as follows:
\begin{definition}\label{def:rad} \citep{shalev2014understanding}
The empirical Rademacher complexity of a set of vectors ${\cal F} \subset \mathbb{R}^T$ is defined as follows:
\begin{equation}
{{\mathfrak{R}}_{T}({\mathcal {F}})={\frac {1}{T}}\mathbb {E} \left[\sup _{\bm f\in {\mathcal {F}}}\sum _{i=1}^{T}\epsilon _{i}f_i\right]},
\end{equation}
where expectation is w.r.t. $\epsilon_i$'s which are i.i.d. Rademacher random variables taking values from $\{-1,1\}$. 
\end{definition}
By Definition~\ref{def:rad}, the empirical Rademacher complexity of the set $\mathcal{F}$ defined in \eqref{eq:setF} is given by
\begin{equation}
{\mathfrak{R}}_{T}({\mathcal {F}})={\frac {1}{T}}\mathbb {E} _{\sigma }\left[\sup _{\bm f\in {\mathcal {F}}}\sum _{i=1}^{T}\sigma _{i}f(m_{\bi};y_{\bi}, \bm z_{\bi})\right].
\end{equation}

To characterize the Rademacher complexity ${\mathfrak{R}}_{T}({\mathcal {F}})$, we proceed to characterize the covering number of ${\cal F}$ which is defined as follows:
\begin{definition} \label{def:covering}\citep{vershynin2012introduction} 
The $\varepsilon$-net of a set ${\cal F}$ represented by $\overline{\cal F}_{\varepsilon}$ is a finite set such that for any $\f\in{\cal F}$, there is a $\overline{\f}\in \overline{\cal F}_{\varepsilon}$ satisfying
\[  \|\f - \overline{\f}\|_2\leq \varepsilon.  \]
The covering number of ${\cal F}$ is ${\sf N}(\cal F,{\varepsilon})=|\overline{\cal F}_{\varepsilon}|$.
\end{definition}
Consider the following:
\begin{align}
    \|\bm f - \overline{\bm f}\|_2^2 &= \sum_{\bi \in \bm \varOmega} (f(m_{\bi},y_{\bi}, \bm z_{\bi}) -  {f}(\overline{m}_{\bi};y_{\bi}, \bm z_{\bi}))^2 \nonumber\\
    & \leq \sum_{\bi \in \bm \varOmega} L_f^2 (m_{\bi} -\overline{m}_{\bi} )^2 \nonumber \\
    &\leq L_f^2 T \|[\tM -\overline{\tM}]_{\bm \varOmega}\|^2_{\rm F} \nonumber \\
    &\le L_f^2 T \|\left[\tM -\overline{\tM}\right]_{\bm \varXi}\|^2_{\rm F}.\label{eq:epsilon_net_f}
\end{align}
where $\overline{\bm f} \in \overline{\mathcal{F}}_{\varepsilon}$ belongs to an $\varepsilon$-net for $ {\mathcal{F}}$, $\overline{\tM}\in \overline{\cal M}_{\varepsilon}$ belongs to an $\varepsilon$-net of ${\cal M}$,  $\overline{m}_{\bi} = [\overline{\tM}]_{\bi}$, and  $T = |\bm \varOmega|$. The last inequality is by applying the assumption that $\bm \varOmega \subseteq \bm \varXi$. 
To characterize the term $L_f$ in \eqref{eq:epsilon_net_f}, we have the following result:
\begin{fact}\label{fact:continuous}
Assume that there exist two real numbers $\beta>0$ and $\alpha<\infty$ such that $\beta\leq m_{\bi}\leq \alpha$ for all $\bi \in \bm \varOmega$. Then, the following hold for all $\bi \in \bm \varOmega$ with probability greater than $1- e^{-\alpha(e^2-3)}$:
\begin{align*}
    &|f_{\bi}'(m_{\bi};y_{\bi}, \bm z_{\bi})| \leq 1 + \frac{\alpha(e^2-2)}{\beta}.
\end{align*}
\end{fact}
The proof is provided in Appendix \ref{app:fact_Lfbound}. 

Hence, applying Fact~\ref{fact:continuous}, in \eqref{eq:epsilon_net_f}, we have the following relation with probability greater than $1- e^{-\alpha(e^2-3)}$,
\begin{align} \label{eq:ell_f}
    L_f = 1+(\alpha(e^2-2))/\beta.
\end{align}
Next, we consider the term $ \|[\tM -\overline{\tM}]_{\bm \varXi}\|^2_{\rm F}$ in \eqref{eq:epsilon_net_f}:
\begin{align}
   \|[\tM -\overline{\tM}]_{\bm \varXi}\|_{\rm F} = &\left\|  \left[\tlambda\circledast \tP - \overline{\tlambda} \circledast \overline{\tP}\right]_{\bm \varXi}  \right\|_{\rm F}\nonumber \\
    =& \left\| \left[ \tlambda\circledast \tP + \overline{\tlambda} \circledast {\tP} - \overline{\tlambda} \circledast {\tP}   - \overline{\tlambda} \circledast \overline{\tP}\right ]_{\bm \varXi}  \right\|_{\rm F} \nonumber \\
    \le&\left\|  [(\tlambda-\overline{\tlambda} ) \circledast {\tP}]_{\bm \varXi} \right\|_{\rm F} +\left\|  [\overline{\tlambda}\circledast ({\tP}   -  \overline{\tP} )]_{\bm \varXi}  \right\|_{\rm F} \nonumber \\
    \leq & \max_{{\bi \in \bm \varXi}} p_{\bi }\|  [\tlambda-\overline{\tlambda}]_{\bm \varXi} \|_{\rm F}  + \max_{{\bi \in \bm \varXi}} \overline{\lambda}_{\bi} \left\|  [ {\tP}   -  \overline{\tP}]_{\bm \varXi} \right\|_{\rm F} \nonumber \\
    \leq & \max_{{\bi}} p_{\bi }\|  \tlambda-\overline{\tlambda} \|_{\rm F}  + \max_{{\bi}} \overline{\lambda}_{\bi} \|g_{\bm \theta}([\bm Z]_{\bm \varXi})-\overline{g}_{\bm \theta}([\bm Z]_{\bm \varXi})\|_{\rm F}\nonumber \\
    \leq & \|  \tlambda-\overline{\tlambda} \|_{\rm F}  +  c_{\lambda}\|g_{\bm \theta}([\bm Z]_{\bm \varXi})-\overline{g}_{\bm \theta}([\bm Z]_{\bm \varXi})\|_{\rm F}, \label{eq:epsilon_net_M}
\end{align}
where $g_{\bm \theta}([\bm Z]_{\bm \varXi}) = [g_{\bm \theta}(\bm z_{\bi_1}),\dots,g_{\bm \theta}(\bm z_{\bi_S})]^{\top}, ~g_{\bm \theta} \in {\cal G}$, $\overline{g}_{\bm \theta}([\bm Z]_{\bm \varXi}) = [\overline{g}_{\bm \theta}(\bm z_{\bi_1}),\dots,\overline{g}_{\bm \theta}(\bm z_{\bi_S})]^{\top}, ~\overline{g}_{\bm \theta} \in \overline{\cal G}_{\varepsilon}$ belongs to the $\varepsilon$-net of ${\cal G}$, $\bm z_{\bi_1},\dots,\bm z_{\bi_S}$ are the set of observed features, and  $\overline{\tlambda}\in \overline{\cal L}_{\varepsilon}$ belongs to an $\varepsilon$-net of ${\cal L}$. The first inequality uses the triangle inequality and the last inequality uses the facts that $\max_{\bi} p_{\bi }\le 1$ and $\max_{\bi} \overline{\lambda}_{\bi} \le c_{\lambda}$ where we have $c_{\lambda}=F\alpha_u^K$ under Assumption \ref{as:bound}.

Eq. \eqref{eq:epsilon_net_f} and \eqref{eq:epsilon_net_M} imply that that if there exists an $\nicefrac{\varepsilon}{2c_{\lambda}L_f\sqrt{T}}$-net covering for ${\cal G} \circ \bm \varXi$, where 
\begin{align} \label{eq:G_Xi}
    {\cal G} \circ \bm \varXi = \{g_{\bm \theta}([\bm Z]_{\bm \varXi}) = [g(\bm z_{\bi_1}),\dots,g(\bm z_{\bi_S})]^{\top}, ~g_{\bm \theta} \in {\cal G}\},
\end{align}
and an $\nicefrac{\varepsilon}{2L_f\sqrt{T}}$-net covering for ${\cal L}$, then we can construct an $\varepsilon$-net to cover ${\cal F}$. Since $\tlambda$ which is a low-rank tensor, we invoke the following result to get the covering number of for ${\cal L}$:

{
\begin{lemma}\citep{fan2020low}\label{lem:tensorcover}
    Let ${\cal L} =\{  \tX\in\mathbb{R}^{I_1\times \ldots \times I_K}~|~\tX =\sum_{f=1}^F \U_1(:,f)\circ \ldots \circ \U_K(:,f),~\|\U_k(:,f)\|_2\leq u \}$. Then, the covering number of ${\cal L}$ with respect to the Frobenius norm satisfies
    \begin{align}\label{eq:tensorcover}
        {\sf N}({\cal L},{\varepsilon}) \leq \left(\frac{3K}{\varepsilon} (F u^2)^{K/2}  \right)^{F\sum_k I_k}.
    \end{align}
\end{lemma}
}
The proof of Lemma \ref{lem:tensorcover} is also detailed in Sec. \ref{app:tensorcover}.

Invoking Lemma \ref{lem:tensorcover}, we have:
    \begin{align}\label{eq:tensorcover_lambda}
        {\sf N}({\cal L},\nicefrac{\varepsilon}{2 L_f\sqrt{T}}) \leq \left(\frac{6K L_f\sqrt{T}(F \alpha_u)^{K}}{\varepsilon}   \right)^{F\sum_k I_k},
    \end{align}
where we have applied $\|\bm U_k(:,f)\|_2\leq \sqrt{F}\alpha_u$.
The covering number of the set ${\cal G} \circ \bm \varXi$ is given by Lemma 14 of \citep{lin2019generalization} (cf. Lemma \ref{lem:nncover} in Sec. \ref{app:complexity_RG}) as below:
    \begin{align} \label{eq:covering_number_G}
           {\sf N}({\cal G} \circ \bm \varXi, {\varepsilon})\leq \exp\left( \frac{\|\bm Z\|_{\rm F}\mathscr{R}_{\mathcal{G}}}{\varepsilon}\right)^{1/2},
    \end{align}
where $\bm Z = [\bm z_{\bi_1},\dots, \bm z_{\bi_S}] \in \mathbb{R}^{D \times S}$ and $\mathscr{R}_{\mathcal{G}}$ denotes the complexity measure of the class ${\cal G}$.
Applying \eqref{eq:covering_number_G}, we get
\begin{align}\label{eq:tensorcover_g}
           {\sf N}({\cal G} \circ \bm \varXi,\nicefrac{\varepsilon}{2c_{\lambda} L_f\sqrt{T}})\leq \exp\left( \frac{2c_{\lambda}L_f\sqrt{T}\|\bm Z\|_{\rm F}\mathscr{R}_{\mathcal{G}}}{\varepsilon}\right)^{1/2}.
\end{align}

From \eqref{eq:tensorcover_lambda} and \eqref{eq:tensorcover_g}, combined with \eqref{eq:epsilon_net_f} and \eqref{eq:epsilon_net_M}, one can obtain that
\begin{align}
 {\sf N}({\cal F},\varepsilon)&\leq 
   {\sf N}({\cal L},\nicefrac{\varepsilon}{2 L_f\sqrt{T}})   \times    {\sf N}({\cal G},\nicefrac{\varepsilon}{2c_{\lambda} L_f\sqrt{T}}) \nonumber\\
 & \leq \left(\frac{6K L_f\sqrt{T}(F \alpha_u)^{K}}{\varepsilon}   \right)^{F\sum_k I_k} \times  \exp\left( \frac{2c_{\lambda}L_f\sqrt{T}\|\bm Z\|_{\rm F}\mathscr{R}_{\mathcal{G}}}{\varepsilon}\right)^{1/2}. \label{eq:coveringM}
\end{align}
To characterize the Rademacher complexity ${\mathfrak{R}}_{T}({\cal F})$ using the covering number ${\sf N}({\cal F},\varepsilon)$, we invoke the following lemma:
\begin{lemma}{\citep[Lemma~A.5]{bartlett2017spectrally}}\label{lemma:dudley_integral}
The empirical Rademacher complexity of the set ${\cal F} \subset \mathbb{R}^T$ is upper bounded as follows:
\begin{align}\label{eq:dudley_integral}
 {\mathfrak{R}}_{T}({\cal F}) \leq \inf_{a > 0} \left( \frac{4 a}{\sqrt{T}} + \frac{12}{T} \int_{a}^{\sqrt{T}} \sqrt{\log{{\sf N}({\cal F}, \mu)}} d\mu\right).
\end{align}
\end{lemma}

Applying Lemma~\ref{lemma:dudley_integral}, we have the following set of relations:
\begin{align}
 {\mathfrak{R}}_{T}({\cal F}) &\leq \inf_{a > 0} \left( \frac{4 a}{\sqrt{T}} + \frac{12}{T} \int_{a}^{\sqrt{T}} \sqrt{\log{{\sf N}({\cal F}, \mu)}} d\mu\right)\nonumber\\
 &\leq \inf_{a > 0} \left(\frac{4a}{\sqrt{T}} + \frac{12}{\sqrt{T}} \sqrt{ \log  {\sf N}({\cal F},a)    }\right)\nonumber\\
  &\leq \inf_{a > 0} \left(\frac{4a}{\sqrt{T}} + \frac{12}{\sqrt{T}} \sqrt{F\sum_k I_k \log\left( \frac{6K L_f\sqrt{T}(F \alpha_u)^{K} }{a}\right)  + \left(    \frac{c_{\lambda}L_f\sqrt{T}\|\bm Z\|_{\rm F}\mathscr{R}_{\mathcal{G}}}{a} \right) } \right)\nonumber\\
& \leq \left( \frac{4}{\sqrt{T}} + \frac{12}{\sqrt{T}} \sqrt{F\sum_k I_k \log\left( 6K L_f\sqrt{T}(F \alpha_u)^{K} \right)  + \left(    {F\alpha^K_uL_f\sqrt{T}\|\bm Z\|_{\rm F}\mathscr{R}_{\mathcal{G}}} \right) }         \right),\label{eq:dudley_integral1}
\end{align}
where the second inequality is obtained since $\sqrt{\log{{\sf N}({\cal F}, \mu)}}$ increases monotonically with the decrease of $\mu$ and hence
\begin{align*}
    \int_{a}^{\sqrt{T}} \sqrt{\log{{\sf N}({\cal F}, \mu)}} d\mu ~\leq~ \sqrt{T} \sqrt{\log{{\sf N}({\cal F}, a)}},
\end{align*}
the third inequality is by applying \eqref{eq:coveringM}, the last inequality is by setting $a=1$, applying $c_{\lambda}=F\alpha^K_u$ and $L_f$ is given by \eqref{eq:ell_f} with probability greater than $1- e^{-\alpha(e^2-3)}$.
\color{black}

\section{Proof of Lemma \ref{lem:fmax}} \label{app:lem_fmax}
We start by noting that $\beta\leq m_{\bi}\leq \alpha$ (see \eqref{eq:landg} and \eqref{eq:upper_bounds}). By Lemma~\ref{lem:poissontail}, we have
\begin{align}\label{eq:bound_y}
  {\sf Pr}\left(y_{\bi} \leq c \right)\geq 1-e^{-\alpha(e^2-3)},~~c= \alpha (e^2-2).
\end{align}
Also, if $\beta<1$, we have $f_{\bi}=m_{\bi} -y_{\bi}\log m_{\bi} \leq \alpha + c |\log \beta|$. Suppose $\beta\geq 1$, we have $f_{\bi}= m_{\bi} -y\log m_{\bi}\leq m_{\bi} + y_{\bi}\log m_{\bi}\leq \alpha + c\log\alpha$. 
Similarly, we get that $f_{\bi} \ge \beta-c\log \alpha$.
Hence, we get
\begin{align*} 
f_{\max}   &\leq \alpha + c\max\{ |\log \beta|,\log\alpha  \}\\
    f_{\min} &\leq \beta - c \log \alpha,
\end{align*}
 which hold with probability greater than $1-e^{-\alpha(e^2-3)}$.

\section{Proof of Fact \ref{fact:continuous}} \label{app:fact_Lfbound}
We have \begin{align*}
    f_{\bi}(m_{\bi}) = m_{\bi} - y_{\bi} \log m_{\bi}.
\end{align*}
Hence, we get
\begin{align} \label{eq:first_derivative}
    f'_{\bi}(m_{\bi}) &= 1 - \frac{y_{\bi}}{m_{\bi}}.
\end{align}
To proceed, we invoke the following lemma:
\begin{lemma}\citep{cao2015poisson}, Lemma 9 (Tail Bound of Poisson) \label{lem:poissontail}
For $y\sim {\sf Poisson}(m)$ with $m\leq \alpha$, we have
\begin{equation}
    {\sf Pr}(y-m \geq t) \leq e^{-t}
\end{equation}
for all $t\geq t_0$ where $t_0 = \alpha(e^2-3)$, where $e$ is the Euler's number. 
\end{lemma}
Utilizing Lemma~\ref{lem:poissontail}, it can be seen that
\begin{align*}
    &{\sf Pr}(y_{\bi} - \alpha \geq \alpha(e^2-3) ) \leq e^{-\alpha(e^2-3)}\\
    \Longrightarrow \quad & {\sf Pr}(y_{\bi}  \leq \alpha(e^2-3)+\alpha ) \geq 1-e^{-\alpha(e^2-3)}.
\end{align*}
 Combining this result with \eqref{eq:first_derivative}, we get that with probability greater than $1-e^{-\alpha(e^2-3)}$,
\begin{align*}
    |f'_{\bi}(m_{\bi})| \le 1+\frac{y_{\bm i}}{m_{\bm i}} \le 1+\frac{\alpha(e^2-2)}{\beta},
\end{align*}
where the last inequality uses the assumption that $m_{\bm i} \ge \beta$.

\section{Proof of Lemma \ref{lem:tensorcover}} \label{app:tensorcover}
Let $\overline{\cal L}_{\varepsilon}$ denote the $\varepsilon$-net of $\cal L$.
Consider $\overline{\tX} \in {\cal L}_{\varepsilon}$. In addition, let us define the following set:
\begin{align*}
    {\cal R}_k= \{\bm U \in \mathbb{R}^{I_k \times F}~|~ \|\bm U(:,f)\|_2 \le u\}.
\end{align*}
Let $\overline{\bm U}_k$ belongs to $\varepsilon$-net of ${\cal R}_k$ such that
$\|\bm U_k-\overline{\bm U}_k\|_{\rm F} \le \varepsilon$. Then, the cardinality of the $\varepsilon$-net of ${\cal R}_k$ is given by \citep{wainwright2019high}
\begin{align} \label{eq:U_netcover}
    {\sf N}({\cal R}_k, \varepsilon) &\le \left(\frac{3\sqrt{Fu^2}}{\varepsilon}\right)^{I_kF}.
\end{align}

In order to derive the covering number of the set ${\cal L}$, we consider the following chain of relations:
\begin{subequations} \label{eq:norm_diff_tX}
\begin{align}
    \|\tX-\overline{\tX}\|_{\rm F} &= \left\lVert\sum_{f=1}^F \U_1(:,f)\circ \ldots \circ \U_K(:,f)\|_{\rm F} - \sum_{f=1}^F \overline{\U}_1(:,f)\circ \ldots \circ \overline{\U}_K(:,f)\right\rVert_{\rm F}\\
    &\le \left\lVert\sum_{f=1}^F \left(\U_1(:,f)-\overline{\U}_1(:,f)\right)\circ \U_2(:,f)\ldots \circ \U_K(:,f)\right\rVert_{\rm F} \nonumber\\
    &\quad + \left\lVert\sum_{f=1}^F \overline{\U}_1(:,f)\circ \left(\sum_{f=1}^F {\U}_2(:,f)\ldots \circ {\U}_K(:,f)-\overline{\U}_2(:,f)\ldots \circ \overline{\U}_K(:,f)\right)\right\rVert_{\rm F}\\
    &\le (\sqrt{Fu^2})^{K-1} \left\lVert\sum_{f=1}^F \left(\U_1(:,f)-\overline{\U}_1(:,f)\right)\right\rVert_{\rm F} \nonumber \\
    &\quad + \sqrt{Fu^2} \underbrace{\left\lVert\sum_{f=1}^F \U_2(:,f)\circ \ldots \circ \U_K(:,f)\|_{\rm F} - \sum_{f=1}^F \overline{\U}_2(:,f)\circ \ldots \circ \overline{\U}_K(:,f)\right\rVert_{\rm F}}_{Q_{\bm U_1}}\\
    &\le (\sqrt{Fu^2})^{K-1}\|\bm U_1- \overline{\U}_1\|_{\rm F} + \sqrt{Fu^2}Q_{\bm U_1}.
\end{align}
\end{subequations}

In a similar way as followed in \eqref{eq:norm_diff_tX}, we can derive upper-bounds for every $Q_{\bm U_k}, k \in [K]$. Then, we can finally establish the below relationship:
\begin{align} \label{eq:tX_covering}
   \|\tX-\overline{\tX}\|_{\rm F}  &\le  (\sqrt{Fu^2})^{K-1}\sum_{k=1}^K \|\bm U_k-\overline{\bm U}_k\|_{\rm F}.
\end{align}

The result in \eqref{eq:tX_covering} implies that if there exists an $\nicefrac{\varepsilon}{K(\sqrt{Fu^2})^{K-1}}$-net to cover each ${\cal R}_k$, then we can construct an $\varepsilon$-net to cover ${\cal L}$. Then, the cardinality of the $\varepsilon$-net of ${\cal L}$ is given by
\begin{align*}
    {\sf N}(\cal L,\varepsilon) & \le \prod_{k=1}^K {\sf N}\left({\cal R}_k, \nicefrac{\varepsilon}{K(\sqrt{Fu^2})^{K-1}}\right)\\
    &\le \left(\frac{3K}{\varepsilon} (F u^2)^{K/2}  \right)^{F\sum_k I_k},
\end{align*}
where the inequality is by applying \eqref{eq:U_netcover}.
Hence the proof.

\section{Proof of Lemma \ref{lem:gap}} \label{app:lem_gap}
 Let us define 
 \begin{subequations}
\begin{align}
    \widehat{u}&=  \frac{1}{{|\bm \varTheta|}} \left\| \left[ \tM_1 - {\tM_2}\right]_{\bm \varTheta}   \right\|^2_{\rm F},\label{eq:uhat} \\
    u&=  \frac{1}{{\prod_k I_k}} \left\|  {\tM}_1 -\tM_2 \right\|^2_{\rm F}.\label{eq:u}
\end{align}
\end{subequations}
 Next, we consider the following lemma which is the Serfling’s sampling-without-replacement
extension of the Hoeffding’s inequality \citep{serfling1974prob}:

\begin{lemma} \label{lem:serfling}
	Let $X_1, X_2, \dots, X_M$ be a set of samples taken without replacement from $\{x_1, x_2, \dots, x_N\}$ of mean $\mu$. Denote $a= \min_i x_i$ and $b = \max_i x_i$. Then
	\begin{align*}
		& {\sf Pr} \left[ \left| \frac{1}{M} \sum_{i=1}^{M} X_i - \mu \right| \geq t \right] \\
		& \leq 2 \exp\left( - \frac{2Mt^2}{(1-(M-1)/N)(b-a)^2}\right) .
	\end{align*}
	\label{lem:hoeffding}
\end{lemma}
One can see that $u$ in \eqref{eq:u} denotes the mean of $\prod_k {I_k}$ terms of $(\widehat{m}_{\bi}-m_{\bi}^\natural)^2$, $\widehat{u}$ denotes the mean of $|\bm \varTheta|$ samples randomly drawn from $\{(\widehat{m}_{\bi}-m^\natural)^2\}$ without replacement. 
Hence, using the assumption that $\bi\in\bm \varTheta$ is drawn uniformly at random from $[I_1] \times \ldots \times [I_K]$ and invoking Lemma \ref{lem:hoeffding}, we have
\begin{align}
    {\sf Pr}\left[ \left| \widehat{u}-u\right|\geq t \right] 
    \leq &2\exp\left( -\frac{2|\bm \varTheta|t^2}{(1-(|\bm \varTheta|-1)/\prod_k I_k)\alpha^4} \right) \nonumber.
\end{align}
where we also applied $(\widehat{m}_{\bi}-m^\natural_{\bi})^2 \le \alpha^2$.

Let $\delta = 2\exp\left( -\frac{2|\bm \varTheta|t^2}{(1-(|\bm \varTheta|-1)/\prod_k I_k)\alpha^4} \right)$, we have the following result
with probability $1-\delta$:
\begin{align}
     \left| \widehat{u}-u\right| 
    \leq &\alpha^2\sqrt{\log\left(\frac{2}{\delta}\right) \left( 1 - \frac{(|\bm \varTheta|-1)}{\prod_k I_k}  \right) \frac{1}{2|\bm \varTheta|}      }.
\end{align}
Using $|\sqrt{a} -\sqrt{b}|\leq \sqrt{|a-b|}$ for nonnegative $a$ and $b$,
we have
\[   \left| \sqrt{\widehat{u} } -  \sqrt{u } \right|\leq \sqrt{\left| \widehat{u}-u\right|},   \]
which implies that
\begin{align}\label{eq:2ndterm}
  \left| \sqrt{\widehat{u} } -  \sqrt{u } \right|
 & \leq \alpha\left(\log\left(\frac{2}{\delta}\right) \left( 1 - \frac{(|\bm \varTheta|-1)}{\prod_k I_k}  \right) \frac{1}{2|\bm \varTheta|}    \right)^{1/4},
\end{align}
holds with probability greater than $1-\delta$.

\section{Proof of Lemma \ref{lem:generalization_g}} \label{app:generalization_g}

	First, let us define the following notations w.r.t the loss function $\ell(p_1,p_2) = (p_1-p_2)^2,~p_1,p_2 \in [0,1]$ and the observed features $\bm Z= [\bm z_{\bi_1},\dots, \bm z_{\bi_S}]$ where each $\bm z_{\bi_s}$ is sampled i.i.d. from the distribution $\mathcal{D}$:
	\begin{subequations}\label{eq:LXf_def}
		\begin{align} 
		L_{\bm Z}(\widehat{g}_{\bm \theta}) &\triangleq \frac{1}{S}\sum_{s=1}^S\ell(g^\natural(\bm z_{\bi_s}),\nicefrac{1}{\widehat{\xi}}\widehat{g}_{\bm \theta}(\bm z_{\bi_s}))=\frac{1}{S}\sum_{s=1}^S (g^\natural(\bm z_{\bi_s})-\nicefrac{1}{\widehat{\xi}}\widehat{g}_{\bm \theta}(\bm z_{\bi_s}))^2=\frac{\left\| \left[  \tP^\natural - \nicefrac{1}{\widehat{\xi}} \widehat{\tP} \right]_{\bXi}\right\|^2_{\rm F} }{{S}} \\
		L_{\mathcal{D}}(\widehat{g}_{\bm \theta}) &\triangleq \mathbb{E}_{\bm z \sim \mathcal{D}} \left[\ell(g^\natural(\bm z),\nicefrac{1}{\widehat{\xi}}\widehat{g}_{\bm \theta}(\bm z))\right]=\mathbb{E}_{\bm z \sim \mathcal{D}} \left[(g^\natural(\bm z)-\nicefrac{1}{\widehat{\xi}}\widehat{g}_{\bm \theta}(\bm z))^2\right].
		\end{align}
	\end{subequations}

	We invoke Theorem 26.5 in \citep{shalev2014understanding} and get that with probability greater than $1-\delta$: 
	\begin{align}
	L_{\mathcal{D}}(\widehat{g}_{\bm \theta}) &\le   L_{\bm Z}(\widehat{g}_{\bm \theta})+ 2\mathfrak{R}_{S}(\ell \circ \mathcal{G} \circ \bm \varXi)+4\bar{c} \sqrt{\frac{2\log(4/\delta)}{S}} \nonumber\\
	\implies \mathbb{E}_{\bm z \sim \mathcal{D}} \left[(g^\natural(\bm z)-\nicefrac{1}{\widehat{\xi}}\widehat{g}_{\bm \theta}(\bm z))^2\right] &\le \frac{\left\| \left[  \tP^\natural - \nicefrac{1}{\widehat{\xi}} \widehat{\tP} \right]_{\bXi}\right\|^2_{\rm F} }{{S}} + 2\mathfrak{R}_{S}(\mathcal{G} \circ \bm \varXi)+4 \sqrt{\frac{2\log(4/\delta)}{S}} \label{eq:gen_f}
	\end{align}
	where the last inequality utilizes the definitions in \eqref{eq:LXf_def}, the contraction lemma (Lemma 26.9 from \citep{shalev2014understanding}) and also applied $\bar{c}=1$ since $|\ell(\bm f,\bm y)|\le 1$ in our case. The term $\mathfrak{R}_{S}(\mathcal{G} \circ \bm \varXi)$ denotes the empirical Rademacher complexity of the function class $\mathcal{G} \circ \bm \varXi$ (see its definition in \eqref{eq:G_Xi}) which is upperbounded via the sensitive complexity parameter $\mathscr{R}_{\mathcal{G}}$ as follows \citep{lin2019generalization}:
	\begin{align*}
	\mathfrak{R}_{S}(\mathcal{G} \circ \bm \varXi) \le 16 S^{-5/8}\left(2 \|\bm Z\|_{\rm F}\mathscr{R}_{\mathcal{G}}\right)^{\frac{1}{4}}.
	\end{align*}

This completes the proof. 
\color{black}

\section{Proof of Lemma \ref{lem:sk_upper_bound}} \label{app:sk_upper_bound}

Consider the following fact due to the cocavity of the $\log$ function and Jensen's inequality \citep{CVX2004}:
\begin{fact} \label{fact:jensens_eqn}
Assume that $\alpha_f$'s are certain scalars such that $\alpha_f > 0, \sum_f \alpha_f =1 $. Then, for any $u_f > 0$,
\begin{align*}
    \log \sum_f \alpha_f u_f \ge \sum_f \alpha_f\log u_f.
\end{align*}
\end{fact}
To proceed, we define the following scalars for our case:
\begin{align*}
    \alpha_{\bm i}^{(f)} &:= \frac{\overline{\bm U}_k (i_k,f) \prod_{j\neq k} \bm U_j(i_j,f)}{\sum_{f=1}^F \overline{\bm U}_k (i_k,f) \prod_{j\neq k} \bm U_j(i_j,f) }, ~\forall f
\end{align*}
One can see that $\sum_{f=1}^F\alpha_{\bm i}^{(f)}  =1$, $\alpha_{\bm i}^{(f)} > 0,~ \forall f$. Hence, we can use Fact \ref{fact:jensens_eqn} to have the following inequality:
\begin{align*}
\log \bigg(\sum_{f=1}^F \bm U_k (i_k,f) \prod_{j\neq k} \bm U_j(i_j,f)\bigg) & \ge   \sum_{f=1}^F \alpha_{\bm i}^{(f)}\log \bigg(\frac{\bm U_k (i_k,f) \prod_{j\neq k} \bm U_j(i_j,f)}{ \alpha_{\bm i}^{(f)}}\bigg).
\end{align*}
Applying the above relation, we immediately have
\begin{align} \label{eq:upper_bound1}
    w(\bm U_k) &\le s(\bm U_k;\overline{\bm U}_k), ~~\forall \bm U_k.
\end{align}
By substituting $\bm U_k=\overline{\bm U}_k$ in $s(\bm U_k;\overline{\bm U}_k)$, we have
\begin{align}
     s(\overline{\bm U}_k; \overline{\bm U}_k) &=
    \sum_{\bm i \in \bm \varOmega}\Bigg[\sum_{f=1}^F\overline{\bm U}_k(i_k,f)  \prod_{j\neq k} \bm U_j(i_j,f) p_{\bm i}- y_{\bm i}\sum_{f=1}^F\alpha_{\bm i}^{(f)}\log \bigg(\sum_{f=1}^F \overline{\bm U}_k (i_k,f) \prod_{j\neq k} \bm U_j(i_j,f) \bigg)\Bigg]\nonumber\\
& \quad = \sum_{\bm i \in \bm \varOmega}\left[\bigg(\sum_{f=1}^F \overline{\bm U}_k(i_k,f)  \prod_{j\neq k} \bm U_j(i_j,f)\bigg) p_{\bm i} -y_{\bm i}\log \bigg(\sum_{f=1}^F \overline{\bm U}_k (i_k,f) \prod_{j\neq k} \bm U_j(i_j,f)\bigg)\right]\nonumber\\
&= w(\overline{\bm U}_k), \label{eq:upper_bound2}
\end{align}
where the second equality is due to $\sum_f\alpha_{\bm i}^{(f)}  =1$.

By taking the derivative of $s(\bm U_k;\overline{\bm U}_k)$ w.r.t. $\bm U_k(i_k,f)$, we have
\begin{align}
     & \nabla_{\bm U_k(i_k,f)} s({\bm U}_k;\overline{\bm U}_k) =
    \sum_{\bm i_{-k} \in \bm \varOmega_{-k}}\Bigg[ \prod_{j\neq k} \bm U_j(i_j,f) p_{\bm i}\Bigg.\nonumber\\
    &\Bigg. \quad\quad- y_{\bm i}\alpha_{\bm i}^{(f)} \bigg(\frac{\alpha_{\bm i}^{(f)}}{\bm U_k (i_k,f) \prod_{j\neq k} \bm U_j(i_j,f)}\bigg)\bigg(\frac{ \prod_{j\neq k} \bm U_j(i_j,f)}{\alpha_{\bm i}^{(f)}}\bigg)\Bigg]\nonumber\\
    &\quad \quad \quad = \sum_{\bm i_{-k} \in \bm \varOmega_{-k}}\Bigg[ \prod_{j\neq k} \bm U_j(i_j,f) p_{\bm i}- y_{\bm i} \bigg(\frac{\alpha_{\bm i}^{(f)}}{\bm U_k (i_k,f) }\bigg)\Bigg]\nonumber\\
    &\quad \quad \quad = \sum_{\bm i_{-k} \in \bm \varOmega_{-k}}\Bigg[ \prod_{j\neq k} \bm U_j(i_j,f) p_{\bm i}\Bigg.\nonumber\\
    &\Bigg. \quad\quad- y_{\bm i} \bigg(\frac{\overline{\bm U}_k (i_k,f) \prod_{j\neq k} \bm U_j(i_j,f)}{\bm U_k (i_k,f)(\sum_{f=1}^F \overline{\bm U}_k (i_k,f) \prod_{j\neq k} \bm U_j(i_j,f)) }\bigg)\Bigg], \label{eq:derivative_s}
\end{align}
where $\bm i_{-k} = (i_1,\dots,i_{k-1},i_{k+1},\dots, i_{K})$, $\bm \varOmega_{-k}$ collects all $\bm i_{-k}$'s such that the corresponding $y_{\bm i}$'s are observed and the last equality \eqref{eq:derivative_s} is obtained by substituting the definition of $\alpha_{\bm i}^{(f)}$.
Further, by taking the derivative of $w(\bm U_k)$ w.r.t. $\bm U_k(i_k,f)$, we get
\begin{align}
 \nabla_{\bm U_k(i_k,f)} w({\bm U}_k) = \sum_{\bm i_{-k} \in \bm \varOmega_{-k}}\Bigg[ \prod_{j\neq k} \bm U_j(i_j,f) p_{\bm i}- y_{\bm i} \bigg(\frac{ \prod_{j\neq k} \bm U_j(i_j,f)}{(\sum_{f=1}^F \overline{\bm U}_k (i_k,f) \prod_{j\neq k} \bm U_j(i_j,f) }\bigg)\Bigg].\label{eq:derivative_w}
 \end{align}
 From \eqref{eq:derivative_s} and \eqref{eq:derivative_w}, one can see that
 \begin{align}
     \nabla_{\bm U_k(i_k,f)} w(\overline{\bm U}_k) &= \nabla_{\bm U_k(i_k,f)} s(\overline{\bm U}_k;\overline{\bm U}_k).\label{eq:upper_bound3}
 \end{align}
 {
\section{Characterization of the Complexity Measure $\mathscr{R}_{\cal{G}}$} \label{app:complexity_RG}

The work in \citep{lin2019generalization} introduces a complexity measure, called the sensitive complexity and denoted as $\mathscr{R}_{\cal{G}}$, to assess the expressive power of various families of neural network classes. For example, regrading the fully connected neural network function classes, the following result is presented:
\begin{lemma}{\citep[Lemma 14]{lin2019generalization}}\label{lem:nncover}
   Consider the following function class: 
    $${\cal G} = \{ g_{\bm \theta}: \mathbb{R}^D \rightarrow [0,1]~|~ g_{\bm \theta}(\bm z) =  \sigma( \bm w_L^\T\bm \sigma(\bm W_{L-1} \bm \sigma (\ldots \bm \sigma(\bm W_1\bm z+ \bm b_1) +\ldots ) +  b_{L})), ~\forall \bm z \in \mathbb{R}^d \},$$
    where $\bm \sigma(\cdot)=[\sigma(\cdot),\ldots,\sigma(\cdot)]^\T$ denote the activation function, $L$ denotes the number of layers, $\bm W_\ell \in \mathbb{R}^{d_\ell \times d_{\ell-1}}$, $d_0=D$, $\bm b_\ell \in \mathbb{R}^{d_\ell}$, and $\bm \theta$ represents the parameters such that $\bm \theta =(\{ \W_\ell,\b_\ell \}_{\ell=1}^{L-1},\w_L,b_L)$. 
    Then, the covering number of ${\cal G} \circ \bm \varXi$  with respect to the Frobenius norm satisfies
    \begin{align}
           {\sf N}({\cal G} \circ \bm \varXi, {\varepsilon})\leq \exp\left( \frac{\|\bm Z\|_{\rm F}\mathscr{R}_{\mathcal{G}}}{\varepsilon}\right)^{1/2},
    \end{align}
    where 
\begin{align*} 
    {\cal G} \circ \bm \varXi = \{g_{\bm \theta}([\bm Z]_{\bm \varXi}) = [g(\bm z_{\bi_1}),\dots,g(\bm z_{\bi_S})]^{\top}, ~g_{\bm \theta} \in {\cal G}\},
\end{align*}
 $\bm Z = [\bm z_{\bi_1},\dots, \bm z_{\bi_S}] \in \mathbb{R}^{D \times S}$ represents the training data  and
    \[\mathscr{R}_{\mathcal{G}} = \left(2\prod_{\ell=1}^L \rho_\ell s_\ell \right) \left(\sum_{\ell=1}^L \frac{d_\ell^2d_{\ell-1}^2a_\ell}{s_\ell} \right)L^2, \]
    where $\|\bm W_\ell\|_{\rm F}\leq a_\ell$, $\|\bm W_\ell\|_2\leq s_\ell$, the activation function $\sigma_\ell$ is $\rho_\ell$-Lipschitz and $\sigma_\ell(0)=0$. 
\end{lemma}
}

 \begin{figure*}[t]
	\centering
	\subfigure[{\scriptsize $\bm \varOmega \subset \bm \varXi$, $\gamma_{\varOmega}=0.1$, $\gamma_{\varXi}=0.3$, $\gamma_{\bm \varTheta}=0.2$}]
	{\includegraphics[width=0.32\linewidth]{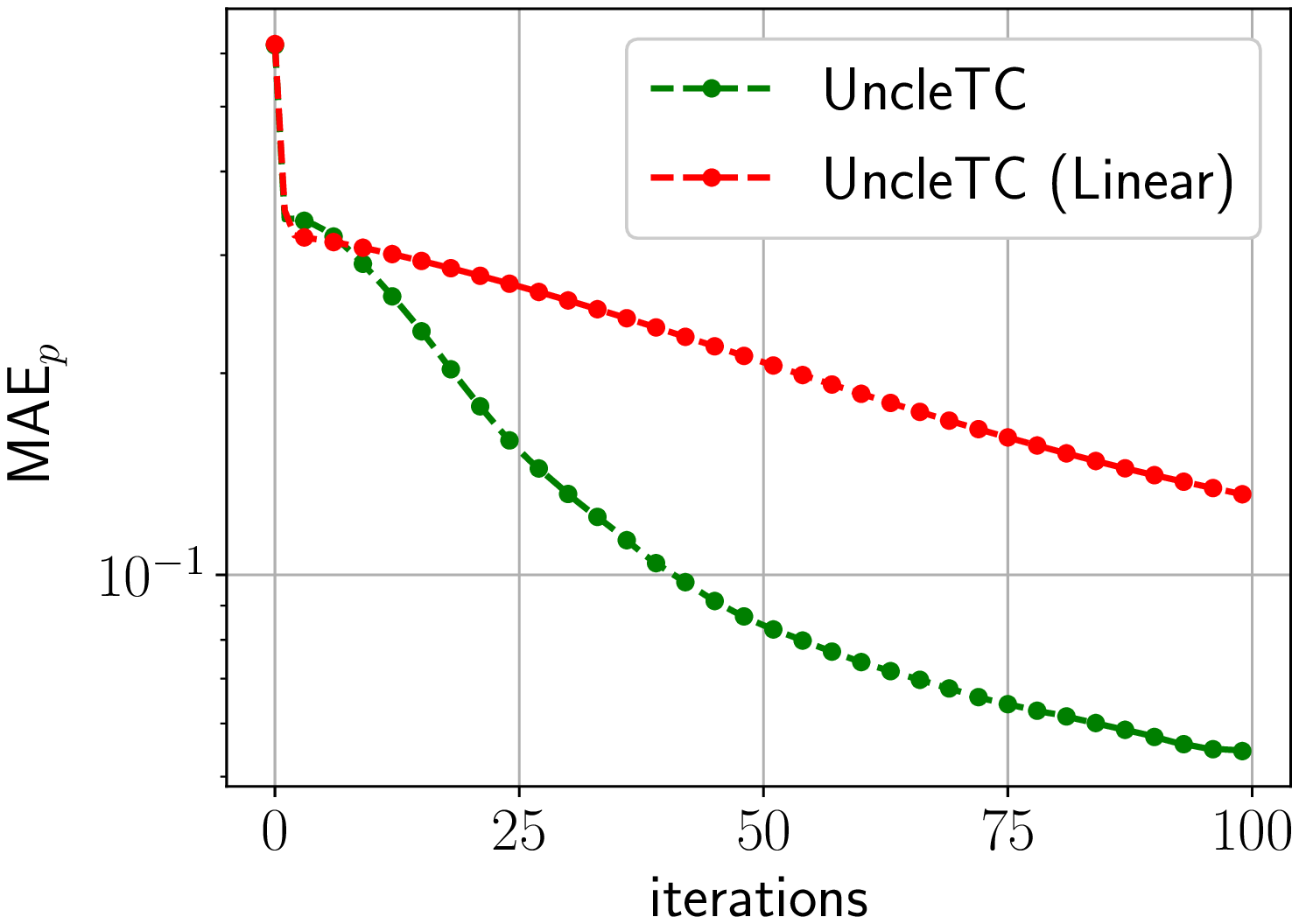}
	\label{fig:mre_p_xi_greater_omega}}
	\subfigure[{\scriptsize $\bm \varOmega = \bm \varXi$, $\gamma_{\varOmega}=0.3$, $\gamma_{\varXi}=0.3$, $\gamma_{\bm \varTheta}=0.2$}]
	{\includegraphics[width=0.32\linewidth]{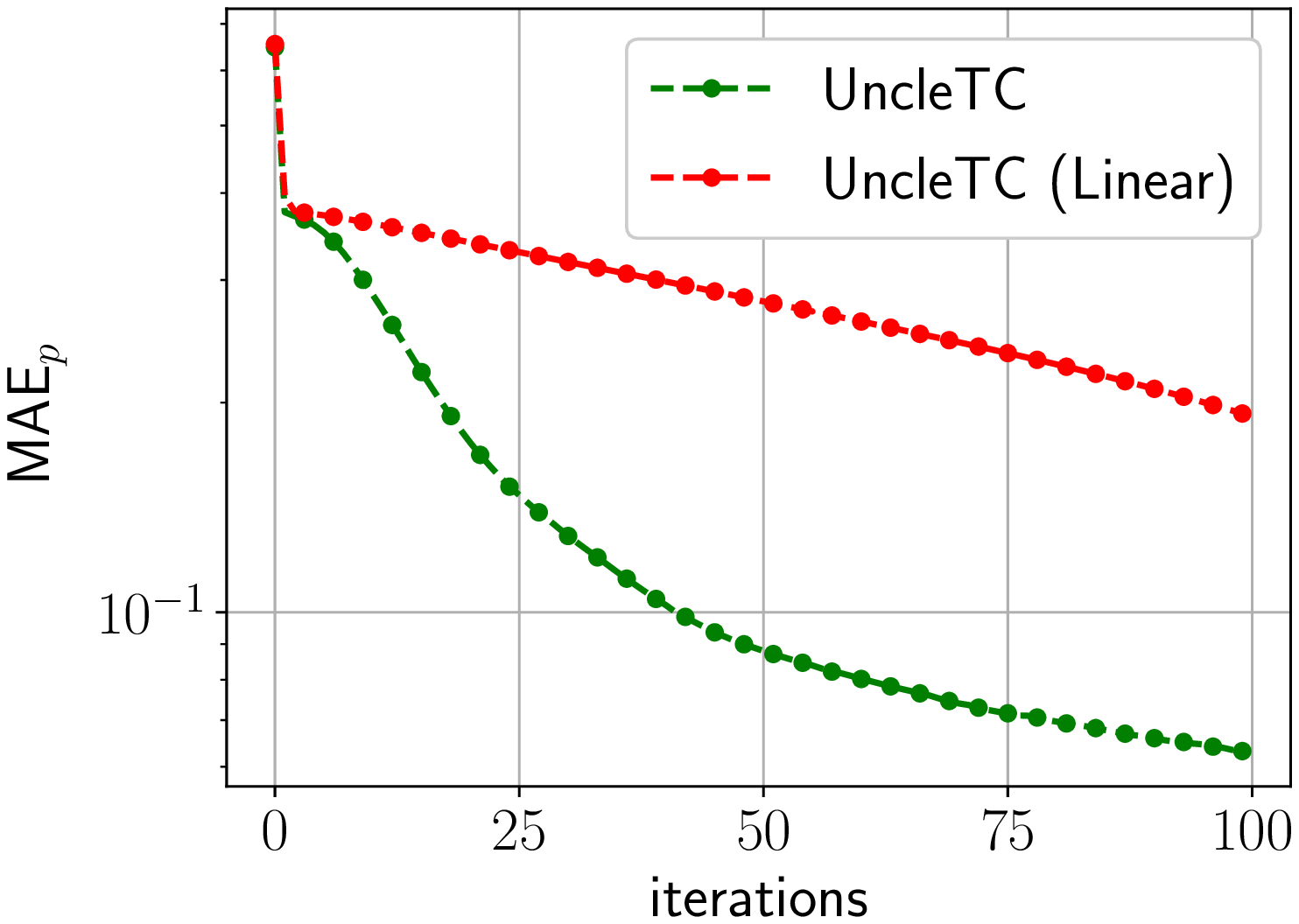}
	\label{fig:mre_p_xi_greater_omega}}
	\subfigure[{\scriptsize $\bm \varOmega \supset \bm \varXi$, $\gamma_{\varOmega}=0.4$, $\gamma_{\varXi}=0.3$, $\gamma_{\bm \varTheta}=0.2$}]
	{\includegraphics[width=0.32\linewidth]{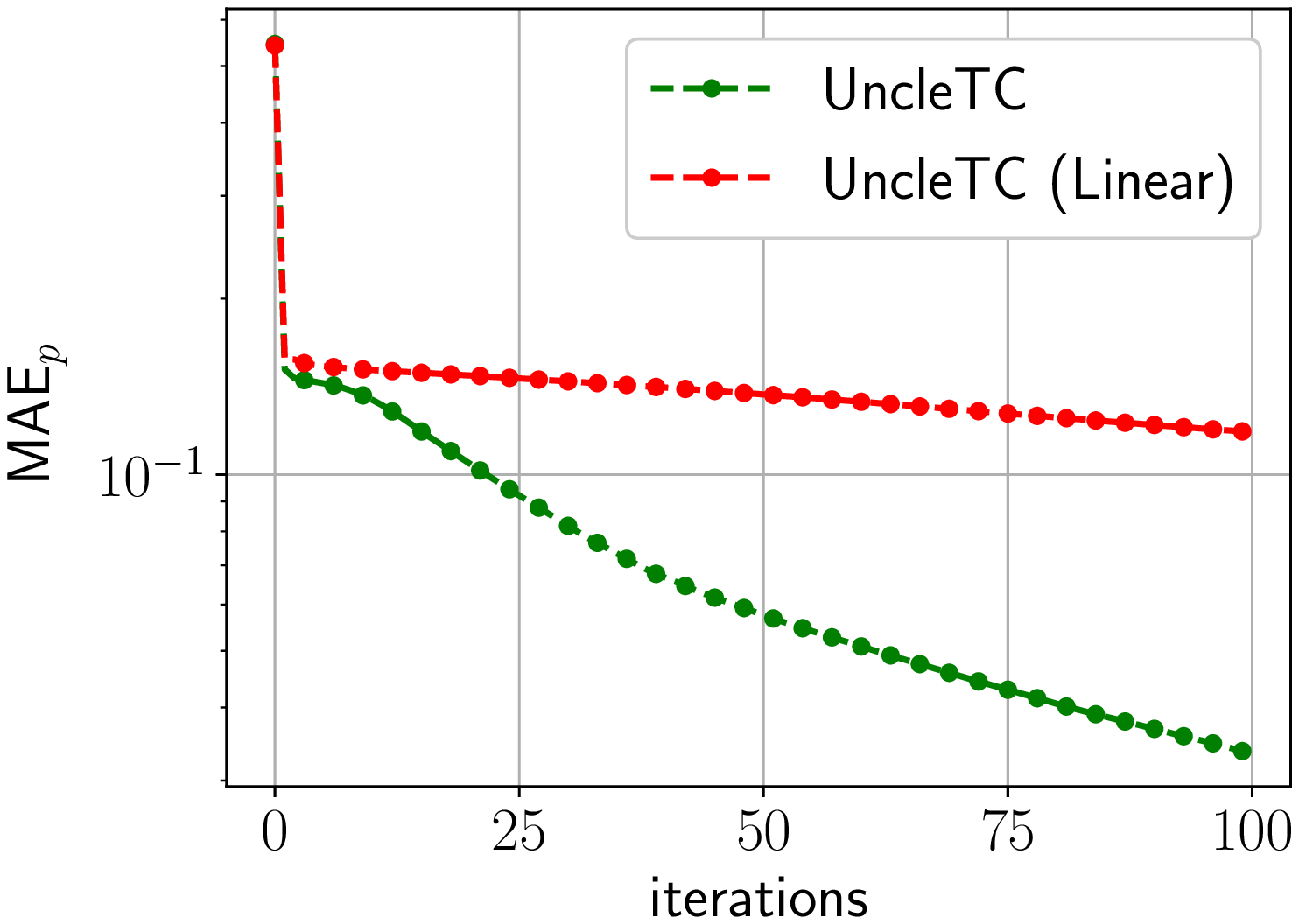}
	\label{fig:mre_p_xi_greater_omega}}
\vfill
	\centering
	\subfigure[{\scriptsize $\bm \varOmega \subset \bm \varXi$, $\gamma_{\varOmega}=0.1$, $\gamma_{\varXi}=0.3$, $\gamma_{\bm \varTheta}=0.2$}]
	{\includegraphics[width=0.32\linewidth]{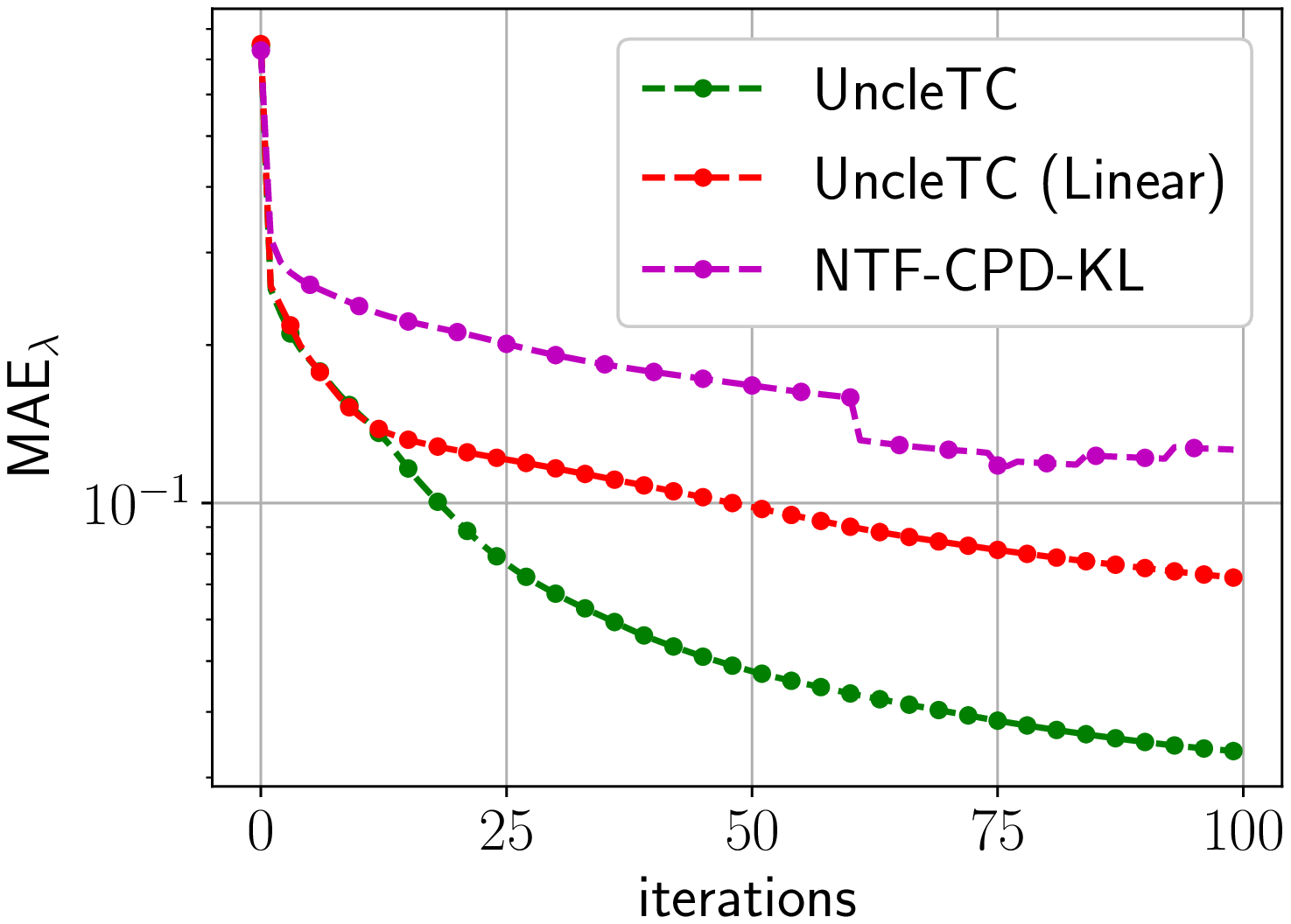}
	\label{fig:mre_p_xi_greater_omega}}
	\subfigure[{\scriptsize $\bm \varOmega = \bm \varXi$, $\gamma_{\varOmega}=0.3$, $\gamma_{\varXi}=0.3$, $\gamma_{\bm \varTheta}=0.2$}]
	{\includegraphics[width=0.32\linewidth]{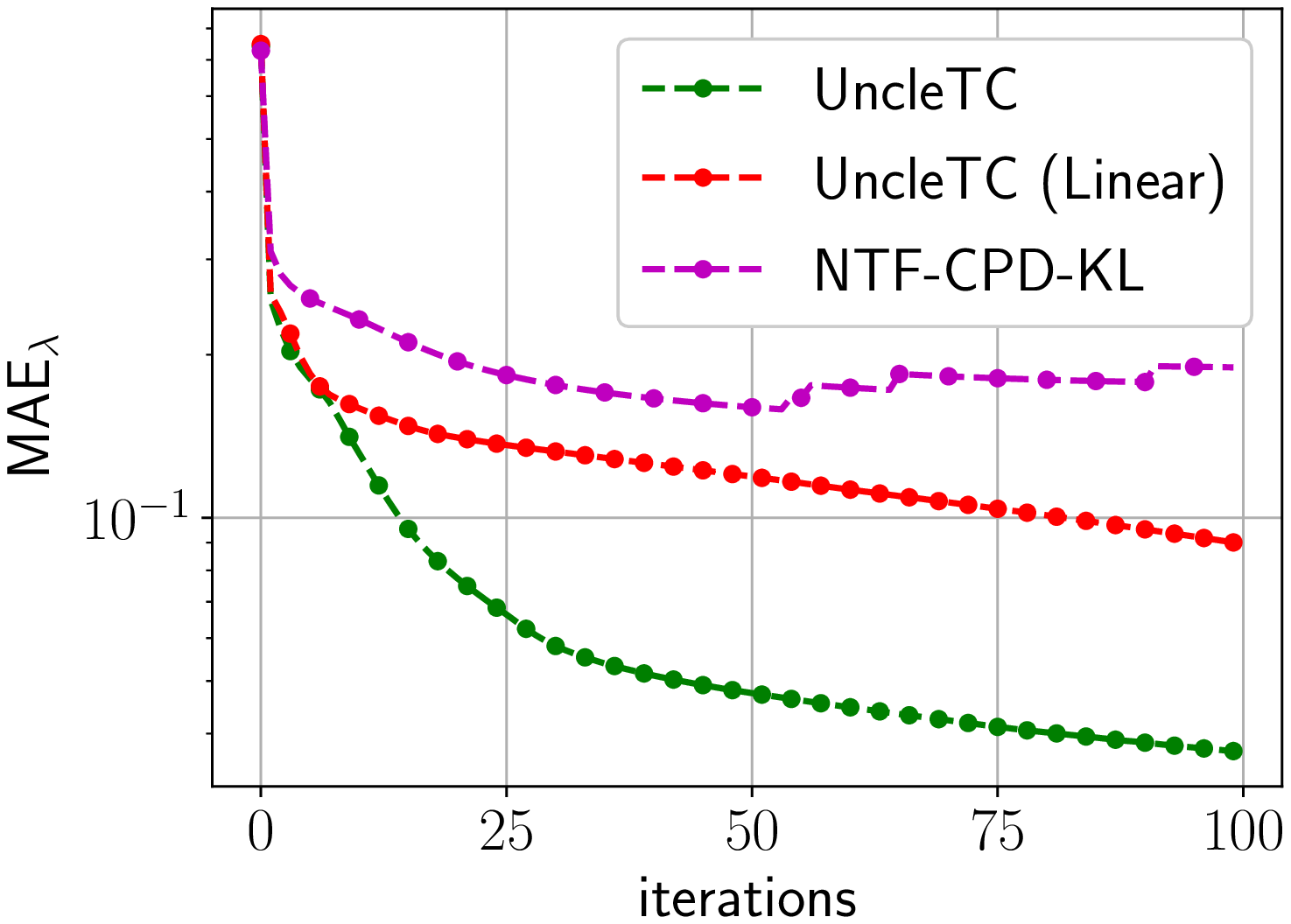}
	\label{fig:mre_p_xi_greater_omega}}
	\subfigure[{\scriptsize $\bm \varOmega \supset \bm \varXi$, $\gamma_{\varOmega}=0.4$, $\gamma_{\varXi}=0.3$, $\gamma_{\bm \varTheta}=0.2$}]
	{\includegraphics[width=0.32\linewidth]{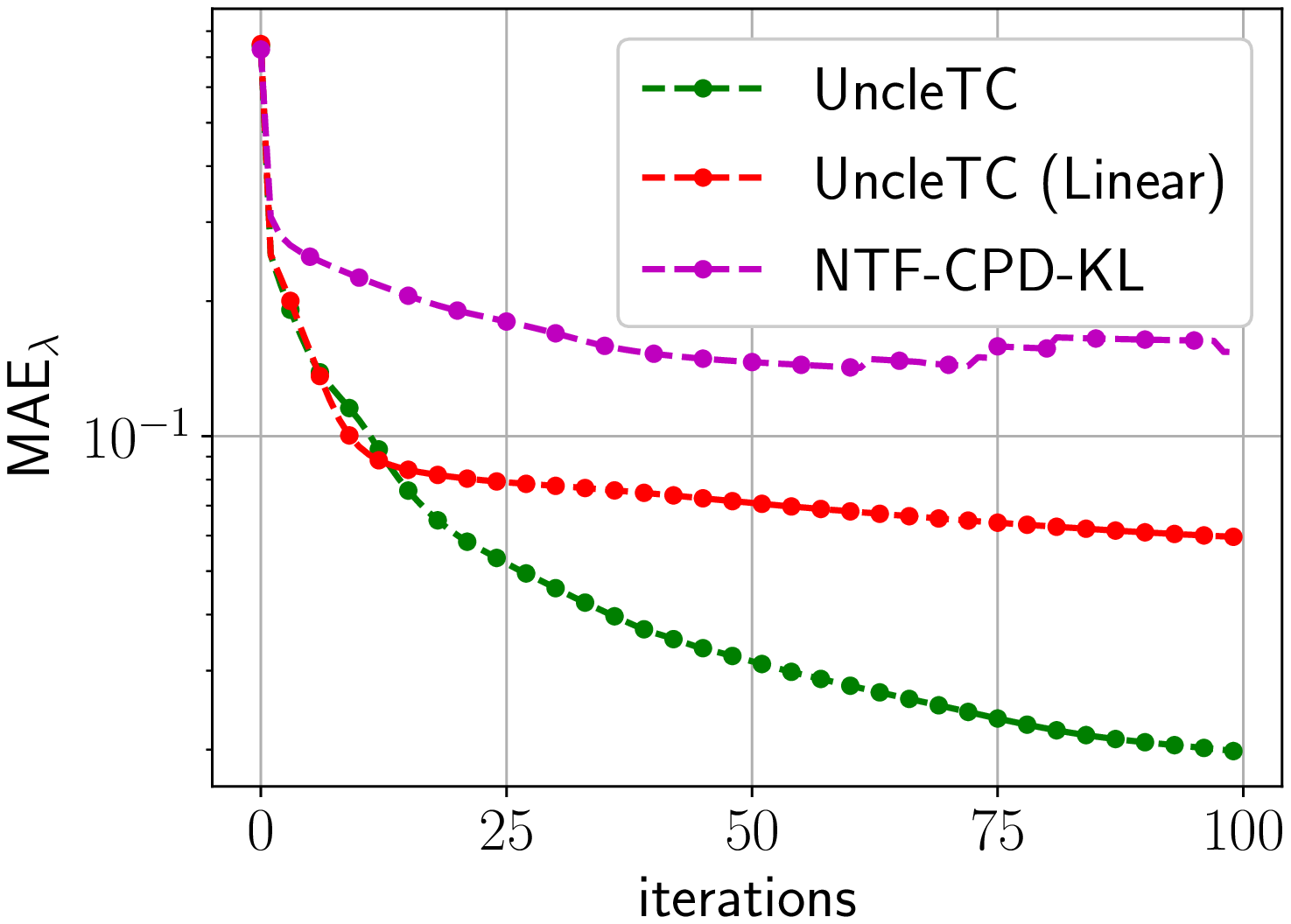}
	\label{fig:mre_p_xi_greater_omega}}
	\caption{Average $\text{MAE}_{p}$ (top) and $\text{MAE}_{\lambda}$ (bottom) over 20 random trials plotted against iterations for different settings $\bm \varOmega \subset \bm \varXi$, $\bm \varOmega = \bm \varXi$, and $\bm \varOmega \supset \bm \varXi$,. $K=3, I_k=20, \forall k,  F=3, D=10, {\rm SNR} = 40\text{dB},  g(\bm z) = {\rm sigmoid}(\bm \nu^{\top}(3\bm z^3+0.2\bm z))$,
 where the vector $\bm \nu \in \mathbb{R}^{D}$ is generated by randomly sampling its entries a uniform distribution between 0 and 1.   }
	\label{fig:plot_mae_lambda_p}
\end{figure*}

\section{More Details on Experiments} \label{app:exp}
\subsection{Synthetic Data Experiments - Implementation Settings.}
 To learn $ g(\cdot)$, we use a neural network $ g_{\bm \theta}(\cdot)$  with 3 hidden layers and 20 {\sf ReLU} activation functions {in each hidden layer}.  
 The optimizer for handling the subproblems of $\bm \theta$ is ${\sf Adam}$ \citep{kingma2015adam} with an initial learning rate of $10^{-3}$. The optimizer uses a batch size of 1024. 
 The subproblems are stopped when the relative change in the corresponding objective functions is smaller than $10^{-6}$. Also, the algorithm is stopped if the relative change in the overall objective function is less than $10^{-6}$ or if 100 BCD iterations are completed. We fix the regularization term in the loss function to be $\ell(x,y) = (x-y)^2$. 
 The regularization parameter $\mu$ is chosen to be a high value, i.e., $\mu =2000$ across all experiments.
 
 \subsection{Additional Synthetic Data Experiments}
 
 In Fig. \ref{fig:plot_mae_lambda_p}, we compare the performance of the proposed approach and two baselines by plotting ${\sf MAE}_{p}$ and ${\sf MAE}_{\lambda}$ over iterations, under different conditions such as $\bm \varOmega \subset \bm \varXi$, $\bm \varOmega \supset \bm \varXi$, and $\bm \varOmega = \bm \varXi$.  The baseline $\texttt{NTF-CPD-KL}$ does not consider the detection probability in their model. {Hence}, we only plot its ${\sf MAE}_{\lambda}$. It can be observed that \texttt{UncleTC} outperforms the baselines {after around 10 iterations under all the settings under test}.

Table \ref{tab:varying_F} {shows the results on a different synthetic dataset}. 
Here, we consider another nonlinear function for the data generation, i.e., $$g(\bm z) = {\rm sigmoid}\left(\bm \nu^{\top}(0.1\log(\bm z^2)+0.1 \bm z^2)\right),$$ where the vector $\bm \nu \in \mathbb{R}^{D}$ is generated by randomly sampling its entries from {the} uniform distribution between 0 and 1. We fix $\gamma_{\varOmega}=0.2$, $\gamma_{\varXi}=0.3$, $\gamma_{\bm \varTheta}=0.2$, ${\rm SNR}=40\text{dB}$, and vary the rank $F$ {of the ground-truth tensor $\tlambda^\natural$}.  One can note that under these settings, the proposed algorithm {still} consistently gives better performance {relative to the baselines}. 

{Table \ref{tab:varying_Fhat} shows the results where wrong tensor ranks $\widehat{F}$'s were used in our algorithms.}
In practice, the underlying data generation process is unknown and hence, the true rank $F$ is hard to know or estimate.
{Hence, it is of interest to understand the algorithm's robustness to using a wrong $\widehat{F}$}.
{One can see that the MAE values are worse when $\widehat{F}<F$ but become better when $\widehat{F}\geq F$.}
This makes sense since underestimating the rank means that less information of the true model is captured by the algorithms. 

{Table~\ref{tab;varying_mu} shows the performance under various $\mu$'s, i.e., the regularization parameter of $\ell$ in \texttt{UncleTC}.}
The settings follow those of Table \ref{tab:varying_F} with rank $F=3$.
The results indicate that using reasonably large $\mu$ ($\mu \ge 1000)$ may be preferable, as we hope that the regularization term can enforce equality.  In practice, one may also use a validation set to choose $\mu$.

\begin{table}[t]
  \centering
  \caption{Average $\text{MAE}_{p}$ and $\text{MAE}_{\lambda}$ over 20 random trials {under} different {values of} $F$. $K=3, I_k=20, \forall k, D=10, \gamma_{\varOmega}=0.2,  \gamma_{\varXi}=0.3, \gamma_{\bm \varTheta}=0.2$, ${\rm SNR}=40\text{dB}, g(\bm z) = {\rm sigmoid}\left(\bm \nu^{\top}(0.1\log(\bm z^2)+0.1 \bm z^2)\right)$. 
  }
  \resizebox{0.65\linewidth}{!}{
    \begin{tabular}{|c|c|c|c|c|}
    \hline
    \textbf{Algorithm} & \textbf{Metric} & \multicolumn{1}{c|}{$F=5$} & \multicolumn{1}{c|}{$F=7$} & \multicolumn{1}{c|}{$F=10$} \\
    \hline
    \hline
    \multirow{2}[4]{*}{\texttt{UncleTC}} & $\text{MAE}_{p}$   & 0.059 $\pm$ 0.001 & 0.049 $\pm$ 0.029 & 0.030 $\pm$ 0.002 \\
\cline{2-5}          & $\text{MAE}_{\lambda}$   &  0.058 $\pm$ 0.007 & 0.052 $\pm$ 0.021 & 0.074 $\pm$ 0.001 \\
    \hline
    \multirow{2}[4]{*}{\texttt{UncleTC (Linear)}} & $\text{MAE}_{p}$   & 0.148 $\pm$ 0.018 & 0.187 $\pm$ 0.026 & 0.077 $\pm$ 0.005 \\
\cline{2-5}          & $\text{MAE}_{\lambda}$  &  0.102 $\pm$ 0.024 & 0.202 $\pm$ 0.024 & 0.131 $\pm$ 0.014 \\
    \hline
   {\texttt{NTF-CPD-KL}} &  $\text{MAE}_{\lambda}$   &  0.166 $\pm$ 0.050 & 0.214 $\pm$ 0.085 & 0.120 $\pm$ 0.001 \\
    \hline
    \hline
    \end{tabular}%
    }
  \label{tab:varying_F}%
\end{table}%

\begin{table*}[t]
  \centering
\caption{Average $\text{MAE}_{p}$ and $\text{MAE}_{\lambda}$ over {20} random trials  for different  $\widehat{F}$ with true rank $F=7$. $K=3, I_k=20, \forall k, D=10, \gamma_{\varOmega}=0.2,  \gamma_{\varXi}=0.3, \gamma_{\bm \varTheta}=0.2$, ${\rm SNR}=40\text{dB}, g(\bm z) = {\rm sigmoid}\left(\bm \nu^{\top}(0.1\log(\bm z^2)+0.1 \bm z^2)\right)$. 
}
  \resizebox{0.95\linewidth}{!}{
    \begin{tabular}{|c|c|c|c|c|c|c|}
    \hline
    \textbf{Algorithm} & \textbf{Metric} & \multicolumn{1}{c|}{$\widehat{F}=3$} & \multicolumn{1}{c|}{$\widehat{F}=5$} & \multicolumn{1}{c|}{$\widehat{F}=7$} & \multicolumn{1}{c|}{$\widehat{F}=10$}& \multicolumn{1}{c|}{$\widehat{F}=12$}\\
    \hline
    \hline
    \multirow{2}[4]{*}{\texttt{UncleTC}} & $\text{MAE}_{p}$   & 0.052 $\pm$ 0.015 & 0.042 $\pm$ 0.014 & 0.046 $\pm$ 0.024  & 0.051 $\pm$ 0.018 & 0.046 $\pm$ 0.019 \\
\cline{2-7}          & $\text{MAE}_{\lambda}$   &0.125 $\pm$ 0.008 & 0.084 $\pm$ 0.011 & 0.060 $\pm$ 0.018 & 0.075 $\pm$ 0.019 & 0.076 $\pm$ 0.019 \\
    \hline
    \multirow{2}[4]{*}{\texttt{UncleTC (Linear)}} & $\text{MAE}_{p}$   &0.110 $\pm$ 0.041 & 0.108 $\pm$ 0.057 & 0.130 $\pm$ 0.045 & 0.112 $\pm$ 0.041 & 0.116 $\pm$ 0.046 \\
\cline{2-7}          & $\text{MAE}_{\lambda}$  &0.141 $\pm$ 0.017 & 0.120 $\pm$ 0.039 & 0.140 $\pm$ 0.046 & 0.155 $\pm$ 0.064 & 0.197 $\pm$ 0.143 \\
    \hline
    {\texttt{NTF-CPD-KL}} & $\text{MAE}_{p}$     & 0.151 $\pm$ 0.029 & 0.134 $\pm$ 0.046 & 0.159 $\pm$ 0.071 & 0.157 $\pm$ 0.072 & 0.151 $\pm$ 0.069 \\
    \hline
    \hline
    \end{tabular}%
    }
  \label{tab:varying_Fhat}%
\end{table*}%
\begin{table}[t]
  \centering
  \caption{Average $\text{MAE}_{p}$ and $\text{MAE}_{\lambda}$ for \texttt{UncleTC} over 20 random trials {under} different $\mu$ values.  We fix $K=3, F=3, D=10, \gamma_{\varOmega}=0.2,  \gamma_{\varXi}=0.3, \gamma_{\bm \varTheta}=0.2, {\rm SNR}=40\text{dB}, g(\bm z) = {\rm sigmoid}\left(\bm \nu^{\top}(0.1\log(\bm z^2)+0.1 \bm z^2)\right)$}
  \resizebox{0.77\linewidth}{!}{
    \begin{tabular}{|c|c|c|c|c|c|}
    \hline
    \textbf{Settings} & \textbf{Metric} & \textbf{$\mu=100$} & \textbf{$\mu=500$} & \textbf{$\mu=1000$} & \textbf{$\mu=2000$} \\
    \hline
    \hline
    \multirow{2}[4]{*}{$I_k=20, \gamma_{\varOmega}=0.3,  \gamma_{\varXi}=0.3$} & $\text{MAE}_{p}$ & 0.184 $\pm$ 0.025 & 0.074 $\pm$ 0.024 & 0.052 $\pm$ 0.011 & 0.042 $\pm$ 0.004 \\
\cline{2-6}          & $\text{MAE}_{\lambda}$ & 0.171 $\pm$0.031  & 0.067 $\pm$ 0.029 & 0.042 $\pm$ 0.014 & 0.028 $\pm$ 0.004 \\
    \hline
    \multirow{2}[4]{*}{$I_k=25, \gamma_{\varOmega}=0.3,  \gamma_{\varXi}=0.3$} & $\text{MAE}_{p}$ & 0.198 $\pm$ 0.054 & 0.083 $\pm$ 0.030 & 0.056 $\pm$ 0.015 & 0.054 $\pm$ 0.013 \\
\cline{2-6}          & $\text{MAE}_{\lambda}$ & 0.184 $\pm$ 0.049 & 0.069 $\pm$ 0.029 & 0.039 $\pm$ 0.013 & 0.033 $\pm$ 0.010 \\
    \hline
    \multirow{2}[4]{*}{$I_k=20, \gamma_{\varOmega}=0.1,  \gamma_{\varXi}=0.3$} & $\text{MAE}_{p}$ & 0.238 $\pm$ 0.069 & 0.151 $\pm$ 0.085 & 0.108 $\pm$ 0.074 & 0.080 $\pm$ 0.044 \\
\cline{2-6}          & $\text{MAE}_{\lambda}$ & 0.256 $\pm$ 0.087 & 0.166 $\pm$ 0.097 & 0.113 $\pm$ 0.084 & 0.078 $\pm$ 0.047 \\
    \hline
    \multirow{2}[4]{*}{$I_k=25, \gamma_{\varOmega}=0.1,  \gamma_{\varXi}=0.3$} & $\text{MAE}_{p}$ & 0.260 $\pm$0.038  & 0.156 $\pm$ 0.073 & 0.141 $\pm$ 0.079 & 0.121 $\pm$ 0.079 \\
\cline{2-6}          & $\text{MAE}_{\lambda}$ & 0.305 $\pm$ 0.045 & 0.203 $\pm$ 0.111 & 0.194 $\pm$ 0.126 & 0.201 $\pm$ 0.134 \\
    \hline
    \hline
    \end{tabular}%
    }
  \label{tab;varying_mu}%
\end{table}%

\subsection{Real Data Experiments - Dataset Details.}
{\bf COVID-19 Data.} The dataset \citep{zhang2020interactive} includes the number of reported COVID-19 cases of about 2270  US counties during 462 days in 2020-2021. The dataset has 25 attributes, including
social distancing index, percentage of people staying home, trips per person, percentage of out-of-county trips, percentage of out-of-state trips, transit mode share, miles per person, work trips per person, non-work trips per person, COVID exposure per 1000 people, percentage of  people older than 60, median income, percentage of African Americans, percentage of Hispanic Americans, population density, number of hot spots per 1000 people, total population, percentage of people working from home, imported COVID cases, hospital beds per 1000 people, testing capacity gap, tests done per 1000 people, unemployment rate, cumulative inflation rate, and unemployment claims per 1000 people.

In order to extract a count-type tensor from the COVID-19 dataset, we adopt the following procedure: Each county is represented using two coordinates, i.e., the rounded value of latitude and longitude of its geographic center (we collected this information from a publicly available website\footnote{\url{https://public.opendatasoft.com/explore/dataset/us-county-boundaries}}). The rounded value of latitude points ranges between 20 and 69, out of which 43 values correspond to the latitudes of at least one county. On the other hand, the discretized longitude points ranges between -165 and -69, out of which 80 discrete points correspond to the longitudes of at least one county.  The third coordinate of the tensor is the dates on which the COVID-19 cases were reported. Hence, each entry of the tensor represents the number of reported COVID-19 cases associated with a latitude point and a longitude point on a particular day. We use the data corresponding to 60 days from January 1, 2021 to March 1, 2021.  We form a $43 \times 80 \times 60$ tensor, which has 19.73\% of observed entries, out of which 15.43\% are nonzeros entries and 4.30\% are zeros.

\begin{figure}[t]
	\centering
	\includegraphics[scale=0.55]{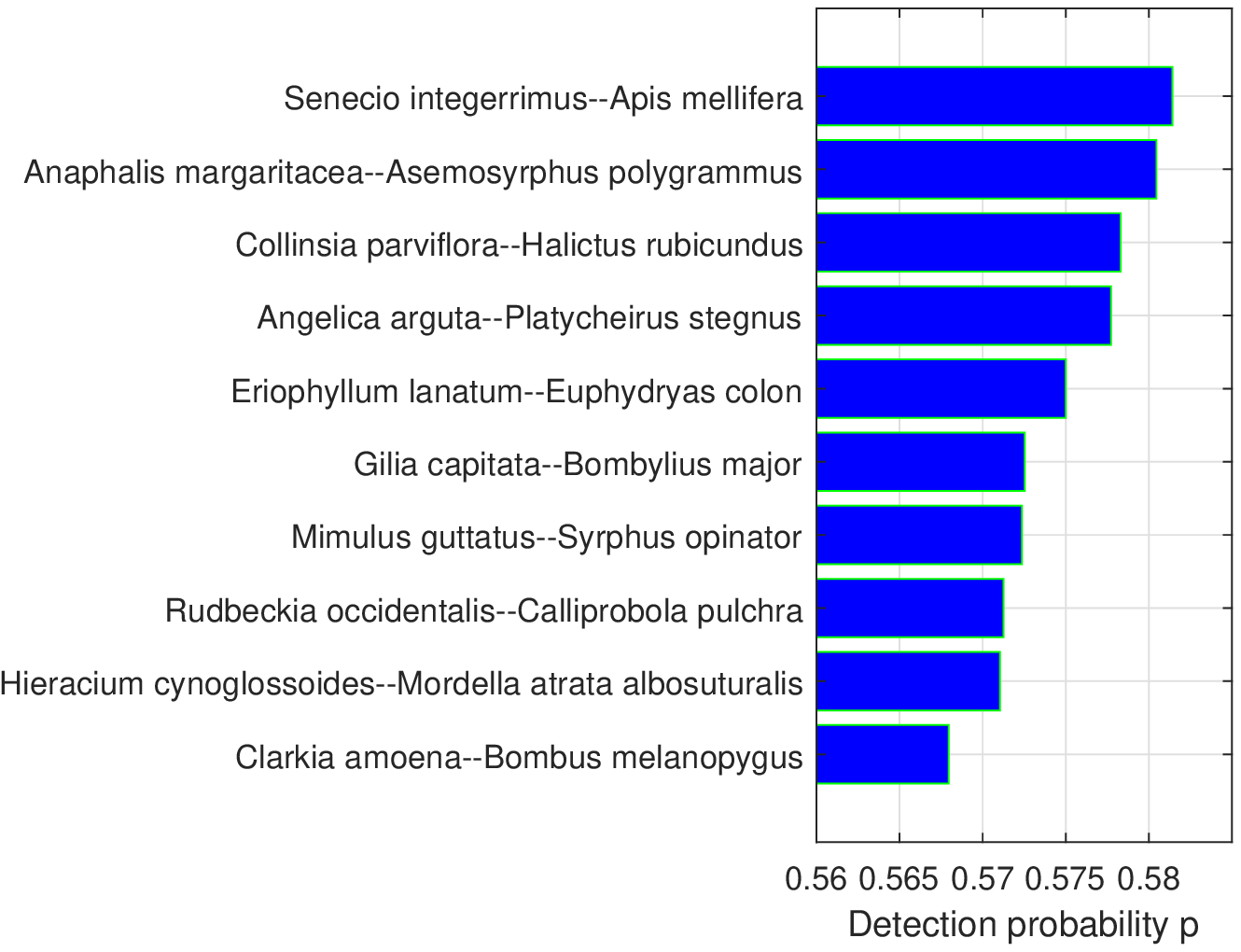}
 \includegraphics[scale=0.55]{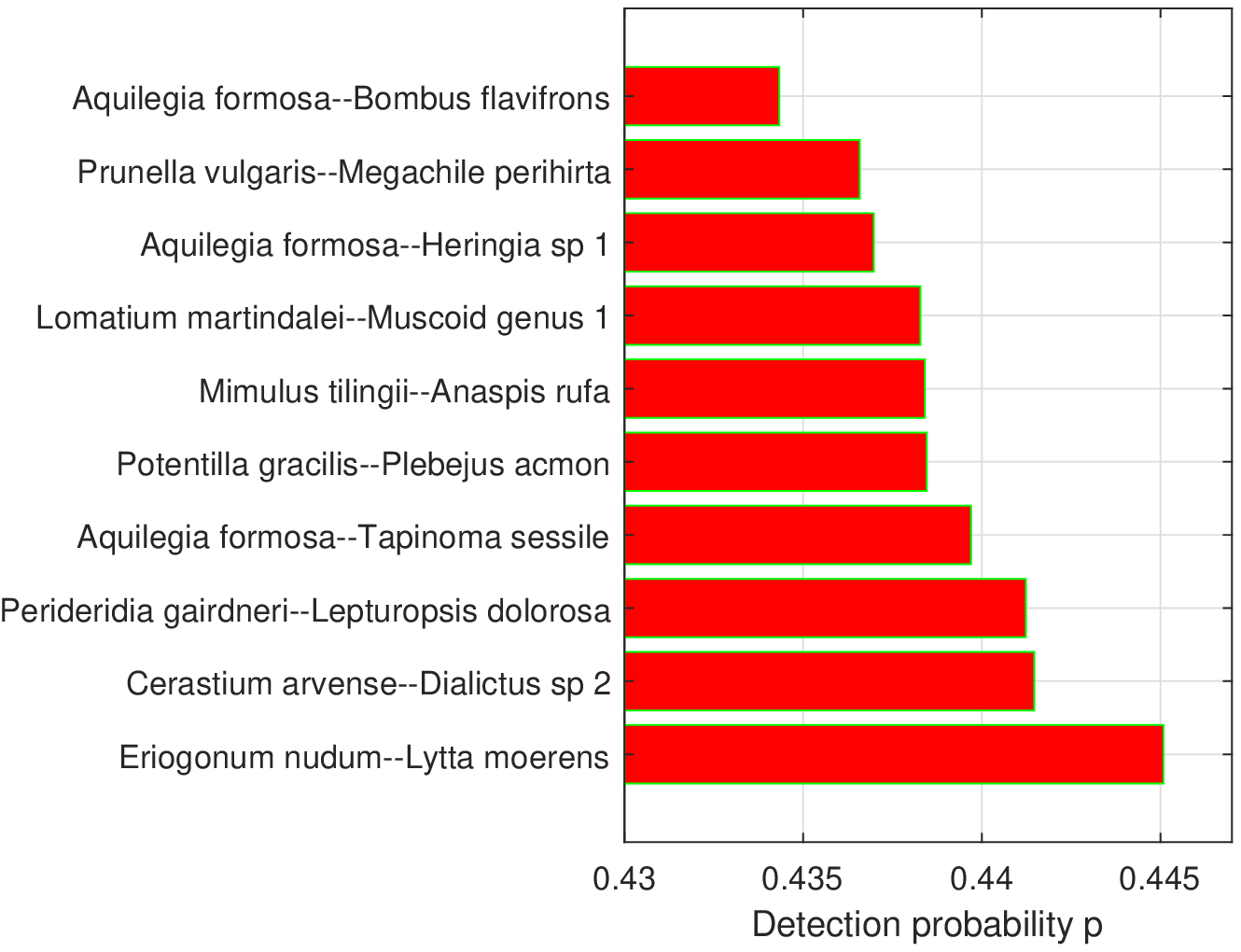}
	\caption{The top 10 plant-pollinator interactions  having highest average detection probabilities (left)  and lowest average average detection probabilities (right) as identified by the proposed \texttt{UncleTC}.}
 \label{fig:detp_ppi_avg}
\end{figure}

\begin{figure}[t]
	\centering
	\includegraphics[scale=0.55]{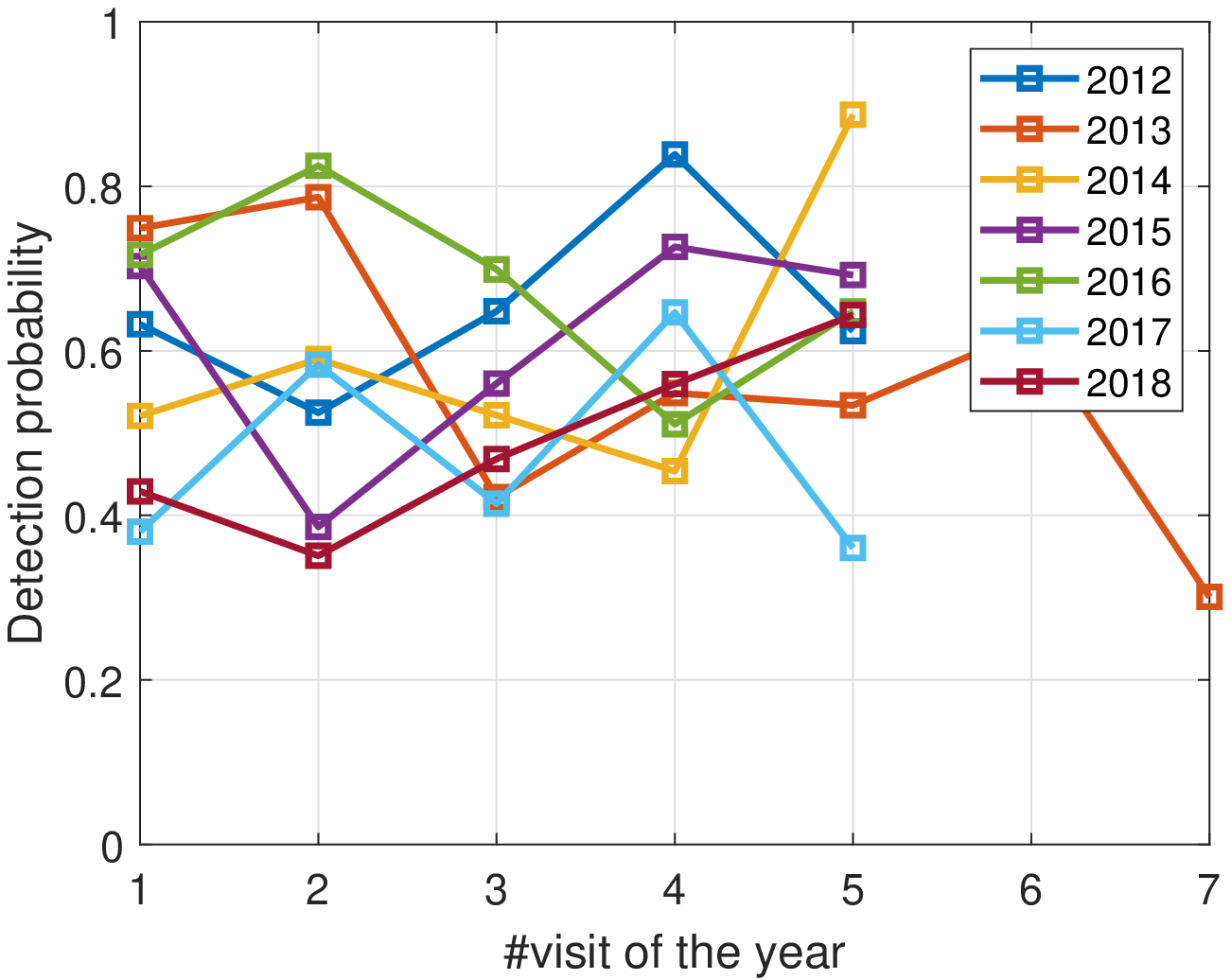}
 \includegraphics[scale=0.55]{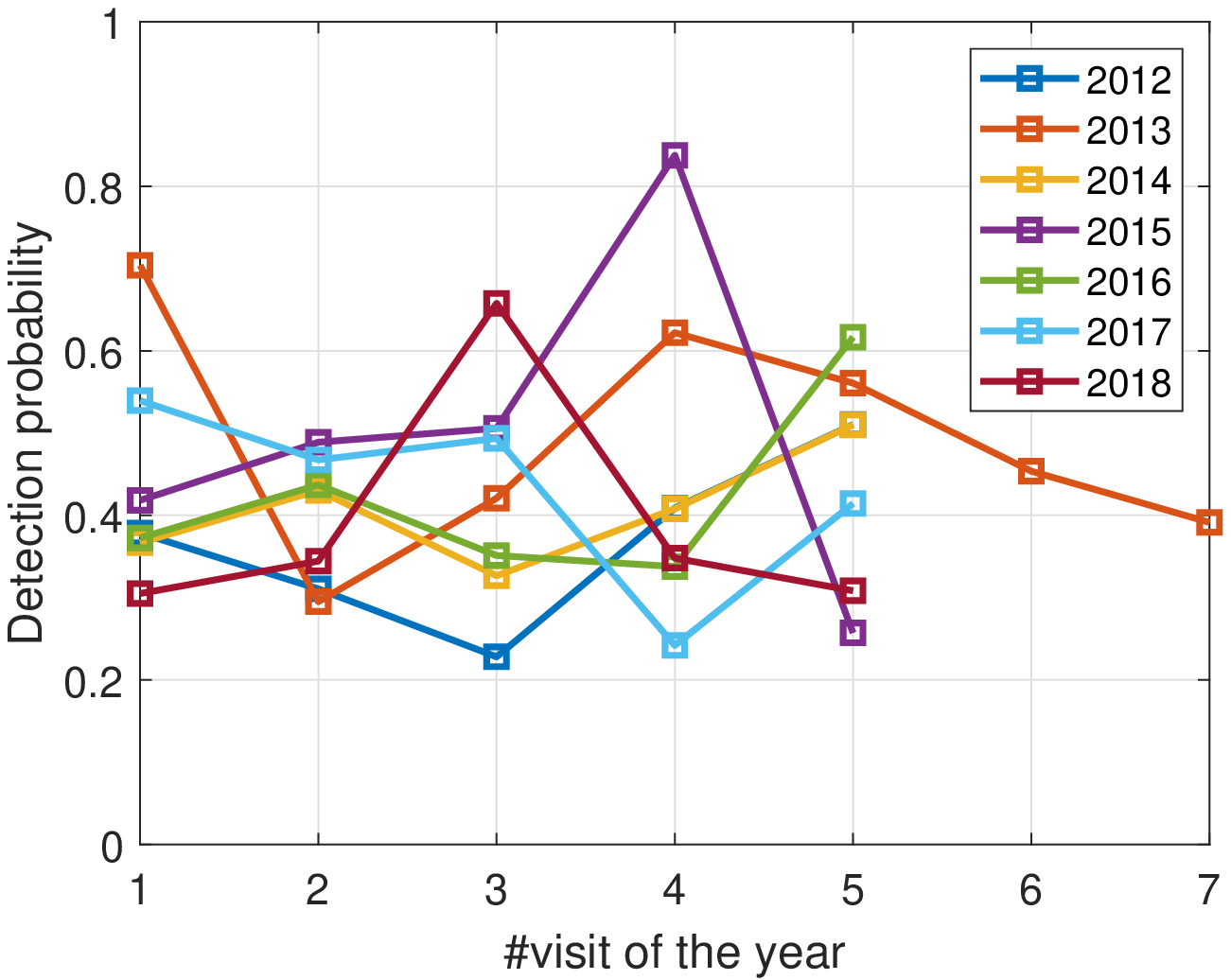}
	\caption{(left) The detection probabilities of the interaction pair \textit{Senecio integerrimus--Apis mellifera}, which {is reported by our algorithm to have} the highest average detection probability. (right) The detection probabilities of the pair \textit{Aquilegia formosa--Bombus flavifrons}, which is reported by our algorithm to have the lowest average detection probability.}
 \label{fig:prob_pattern1}
\end{figure}

\begin{figure}[t]
	\centering
	\includegraphics[scale=0.55]{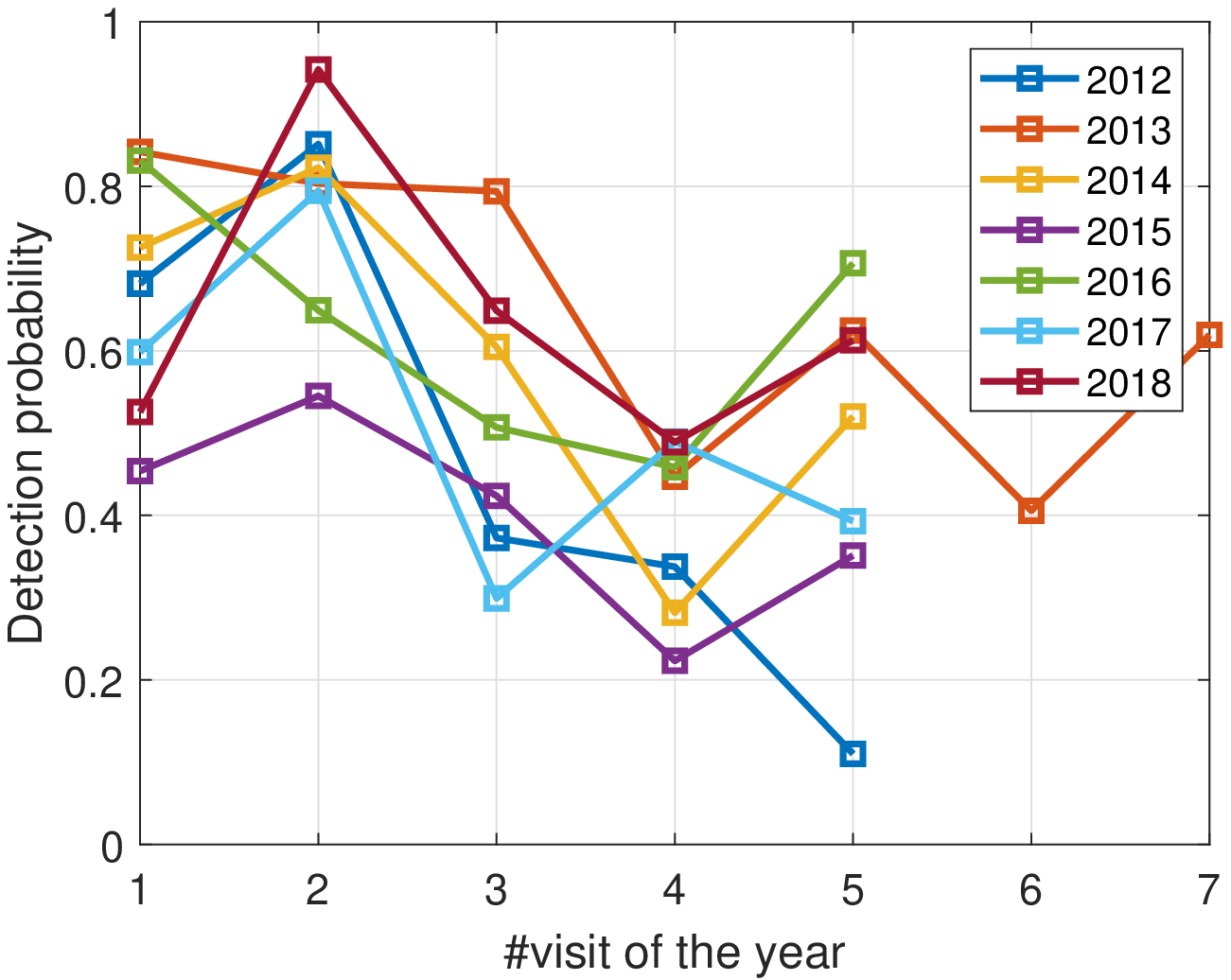}
 \includegraphics[scale=0.55]{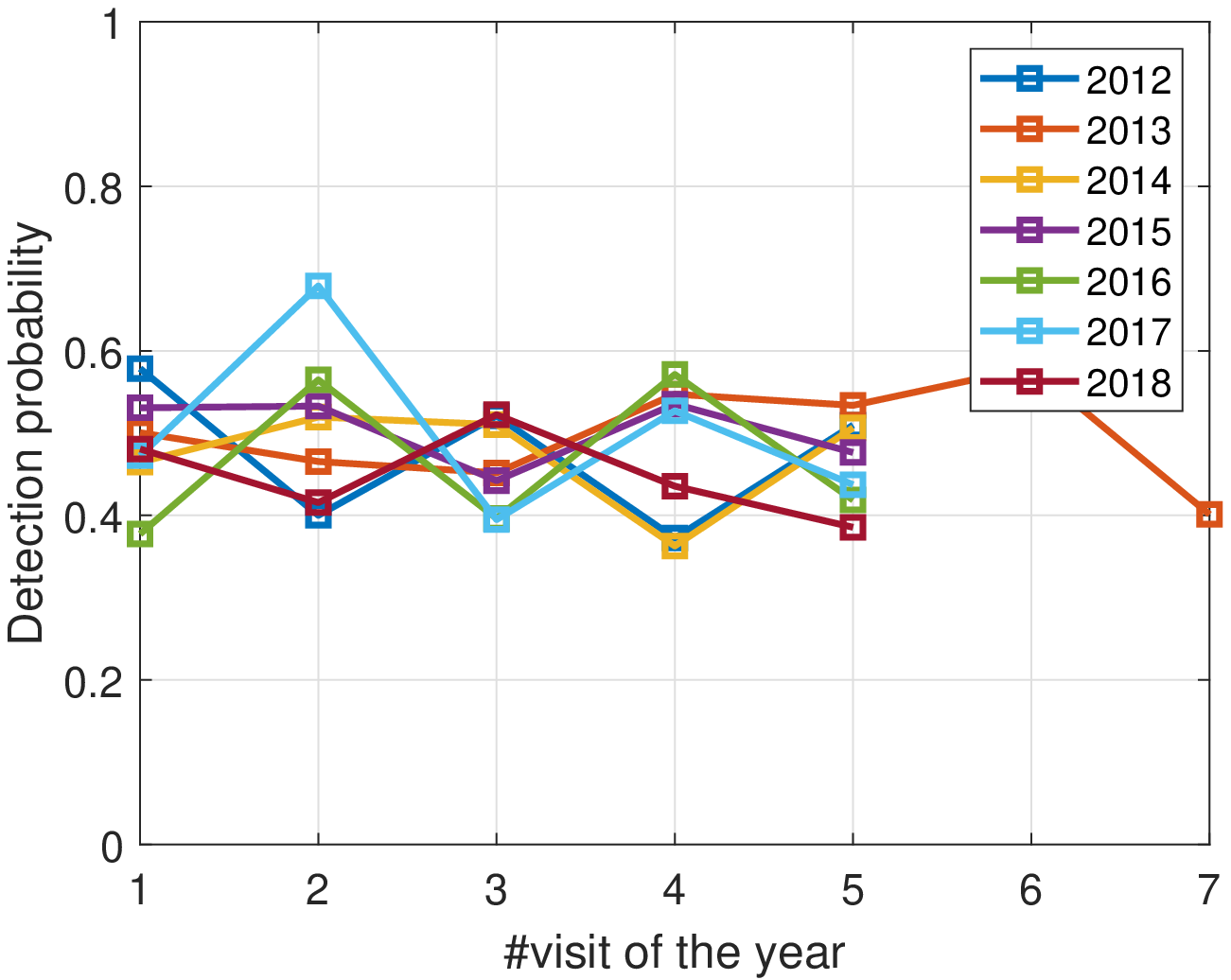}
	\caption{(left) The detection probability of the interaction \textit{Gilia capitata--Apis mellifera} {over different years}, which {is} reported {by our algorithm} {to have} the highest standard deviation {of the} detection probability across 37 visits. (right) The detection probability pattern for the interaction \textit{Eriophyllum lanatum--Bombus mixtus}, which {is} reported {by our method to} have the lowest standard deviation of the detection probability across 37 visits.  }
 \label{fig:prob_pattern2}
\end{figure}

{\bf Pollinator-Plant Interaction (PPI) Data.} 
The Plant-Pollinator Interaction (PPI) dataset is a publicly available dataset, collected by the researchers at the H.J. Andrews
(HJA) Long-term Ecological Research site in Oregon, USA \citep{Jones2017,Daly1957}. The dataset comprises  {plant-pollinator interactions} observed at 
about 12 meadows over {a period of seven years.}
 Each meadow was visited 5-7 times per year by student researchers. 
 The dataset contains interactions between 69 plant species and 215 pollinator species observed during 37 visits. 
 Only 1.07\% of the entries are recorded.  The counts in this dataset are widely believed to be under-counted \citep{fu2019link,seo2018predicting}. We consider a subset of the dataset by selecting 50 plants and 50 pollinators that exhibit the highest number of interactions. This leads to a $50 \times 50 \times 37$ integer tensor with $22.06\%$ observed entries, out of which $4.01\%$ are nonzero entries and $18.05\%$ are zero entries. The zero entries indicate that the plant and pollinator were both observed by the researcher during the visit, but they were not observed to be interacting with one another.

The dataset has 37 attributes:

$\bullet$ 12 plant species static traits: the estimate of reward per flower, the basic ecological life form of a plant, floral structure, peduncle feebleness, the level of exclusion, exclusion by flower color, exclusion by pendant position, time of diel availability, size of pollen grain, the ability of the inflorescence to support the arrival of the flower visitor, the possible affinity of plant species to edaphic characteristics, and dispersal mechanism.

$\bullet$ 11 pollinator species static traits: the length of a species body size, the width of a species body size, the depth of a species body size, individual biomass, size, energy requirement, a species behavior characterized by active time, the level of exclusion, an indicator showing whether the platform is required to approach flowers, tongue length, and tube size.

$\bullet$ 1 plant temporal feature: flower abundance

$\bullet$ 4 temporal features: month, day, wind, and cloud 

$\bullet$ 7 temperature features: mean, max, min, max time, min time, cumulative degree days with 5 degrees,  and cumulative degree days with 10 degrees.

$\bullet$ 2 precipitation features: daily precipitation, and antecedent with $k=0.9$

  \subsection{Real Data Experiments - Implementation Settings.}
 For the proposed \texttt{UncleTC}, we fix {\sf ReLU} as the neural network activation function and {\sf Adam} as the the optimizer. Other parameters such as rank $F$, the regularization parameter $\mu$, batch size, learning rates, and neural network size are selected via the grid search hyper-parameter tuning strategy on the validation set where the {\sf rRMSE} is used to measure the validation score. 
 For the proposed \texttt{UncleTC}, we choose the rank $F$ of the tensor from $[5, 10, 15, 20]$, the regularization parameter $\mu$ from $[500,1000, 2000, 3000]$, the number of hidden units from $[10,20,30]$, the number of hidden layers from $[1,3,5]$, the learning rate from  [0.001, 0.005, 0.01] and the batch size from $[256, 512, 1024]$. For all the low-rank tensor decomposition-based baselines, we select the rank from $[5, 10, 15, 20]$. In addition, the sparsity parameter of the \texttt{BPTF-CPD} \citep{schein2015bayesian} is selected from  $[0.02, 0.05, 0.1, 0.2]$ and the regularization parameter of the \texttt{HaLRTC} \citep{liu2012tensor} is selected from $[10^{-7}, 10^{-6}, 10^{-5}, 10^{-4}]$, based on the recommended range of values from the respective {papers}. The number of iterations to stop training all the iterative algorithms is also chosen {using the validation sets}.

 In the experiments, we perform $5$-fold cross-validation to characterize the prediction performance with $k$ chosen to be 5, where 3 splits serve for training, and the other two for validation and testing, respectively. The roles of the sets are switched in a cyclical way in different trials.

\subsection{Real Data Experiments - Additional Results.}
Fig. \ref{fig:detp_ppi_avg} displays the top 10 interactions with the highest and lowest average detection probabilities, as discovered by our proposed \texttt{UncleTC} method. Generally, interactions involving larger, more common species are more likely to have higher detection probabilities, while smaller, rarer species tend to have lower detection probabilities. The species with the highest average detection probabilities have a larger number of observed interactions compared to those with the lowest average detection probabilities. However, it is worth noting that not all species follow this trend. Other factors such as 
species rareness, 
environmental conditions, or seasonal variability may contribute to these differences \citep{dorado2011rareness, chacoff2012evaluating}. 

In Fig. \ref{fig:prob_pattern1}, we present the periodic variation of detection probabilities for specific species interactions over the years. Both figures illustrate temporal variation across visits and years for two different interactions—one with the highest average detection probability and the other with the lowest. Despite the variations, the overall range remains relatively consistent, with probabilities ranging from approximately 0.4 to 0.8 in the left figure and from 0.2 to 0.6 in the right figure, ultimately leading to the highest and lowest average detection probabilities.

Considering the exceptions where detection probability does not align with species activity levels, we further examined interactions that show the most and least significant variations in detection probabilities over time. Fig. \ref{fig:prob_pattern2} plots the detection probabilities over visits for interactions with the highest (left) and lowest (right) standard deviation of detection probabilities across visits and annually. The left figure confirms that the detection probability of the interaction between two well-known common species (\textit{Gilia capitata} and \textit{Apis mellifera}) is significantly influenced by the time of year, resulting in a relatively lower average detection probability. The right figure provides another example of an interaction between another plant and pollinator species (\textit{Eriophyllum lanatum} and \textit{Bombus mixtus}), with relatively lower detection probabilities over time. 
Such variation may be a product of the phenology of the species (i.e., when flowers are blooming and when insects are flying).


\begin{thebibliography}{62}
\providecommand{\natexlab}[1]{#1}
\providecommand{\url}[1]{\texttt{#1}}
\expandafter\ifx\csname urlstyle\endcsname\relax
  \providecommand{\doi}[1]{doi: #1}\else
  \providecommand{\doi}{doi: \begingroup \urlstyle{rm}\Url}\fi

\bibitem[Acar et~al.(2009)Acar, Dunlavy, and Kolda]{acar2009link}
Acar, E., Dunlavy, D.~M., and Kolda, T.~G.
\newblock Link prediction on evolving data using matrix and tensor
  factorizations.
\newblock In \emph{IEEE International Conference on Data Mining Workshop}, pp.\
   262--269, 2009.

\bibitem[Balazevic et~al.(2019)Balazevic, Allen, and
  Hospedales]{ivana2019tucker}
Balazevic, I., Allen, C., and Hospedales, T.
\newblock {T}uck{ER}: Tensor factorization for knowledge graph completion.
\newblock In \emph{Proceedings of Empirical Methods in Natural Language
  Processing and International Joint Conference on Natural Language
  Processing}, pp.\  5185--5194, 2019.

\bibitem[Bartlett et~al.(2017)Bartlett, Foster, and
  Telgarsky]{bartlett2017spectrally}
Bartlett, P.~L., Foster, D.~J., and Telgarsky, M.~J.
\newblock Spectrally-normalized margin bounds for neural networks.
\newblock In \emph{Advances in Neural Information Processing Systems},
  volume~30, 2017.

\bibitem[Bertsekas(1999)]{bertsekas1999nonlinear}
Bertsekas, D.~P.
\newblock \emph{Nonlinear programming}.
\newblock Athena Scientific, 1999.

\bibitem[Billio et~al.(2022)Billio, Casarin, and Iacopini]{billio2017bayesian}
Billio, M., Casarin, R., and Iacopini, M.
\newblock Bayesian markov-switching tensor regression for time-varying
  networks.
\newblock \emph{Journal of the American Statistical Association}, pp.\  1--13,
  2022.

\bibitem[Boyd \& Vandenberghe(2004)Boyd and Vandenberghe]{CVX2004}
Boyd, S. and Vandenberghe, L.
\newblock \emph{Convex Optimization}.
\newblock Cambridge University Press, 2004.

\bibitem[Cao \& Xie(2015)Cao and Xie]{cao2015poisson}
Cao, Y. and Xie, Y.
\newblock Poisson matrix recovery and completion.
\newblock \emph{IEEE Transactions on Signal Processing}, 64\penalty0
  (6):\penalty0 1609--1620, 2015.

\bibitem[Chacoff et~al.(2012)Chacoff, V{\'a}zquez, Lom{\'a}scolo, Stevani,
  Dorado, and Padr{\'o}n]{chacoff2012evaluating}
Chacoff, N.~P., V{\'a}zquez, D.~P., Lom{\'a}scolo, S.~B., Stevani, E.~L.,
  Dorado, J., and Padr{\'o}n, B.
\newblock Evaluating sampling completeness in a desert plant--pollinator
  network.
\newblock \emph{Journal of Animal Ecology}, 81\penalty0 (1):\penalty0 190--200,
  2012.

\bibitem[Chi \& Kolda(2012)Chi and Kolda]{chi2012tensors}
Chi, E.~C. and Kolda, T.~G.
\newblock On tensors, sparsity, and nonnegative factorizations.
\newblock \emph{SIAM Journal on Matrix Analysis and Applications}, 33\penalty0
  (4):\penalty0 1272--1299, 2012.

\bibitem[Daly et~al.(2019)Daly, Schulze, and McKee]{Daly1957}
Daly, C., Schulze, M.~D., and McKee, W.~A.
\newblock Meteorological data from benchmark stations at the {HJ} {A}ndrews
  experimental forest, 1957 to present, 2019.
\newblock URL
  \url{http://andlter.forestry.oregonstate.edu/data/abstract.aspx?dbcode=MS001.
  https://doi.org/10.6073/pasta/c021a2ebf1f91adf0ba3b5e53189c84f}.

\bibitem[Dennis et~al.(2015)Dennis, Morgan, and
  Ridout]{dennis2015computational}
Dennis, E.~B., Morgan, B.~J., and Ridout, M.~S.
\newblock Computational aspects of {N}-mixture models.
\newblock \emph{Biometrics}, 71\penalty0 (1):\penalty0 237--246, 2015.

\bibitem[Ding et~al.(2020)Ding, Fu, Huang, Wang, and
  Zhao]{ding2020hyperspectral}
Ding, M., Fu, X., Huang, T.-Z., Wang, J., and Zhao, X.-L.
\newblock Hyperspectral super-resolution via interpretable block-term tensor
  modeling.
\newblock \emph{IEEE Journal of Selected Topics in Signal Processing},
  15\penalty0 (3):\penalty0 641--656, 2020.

\bibitem[Dorado et~al.(2011)Dorado, V{\'a}zquez, Stevani, and
  Chacoff]{dorado2011rareness}
Dorado, J., V{\'a}zquez, D.~P., Stevani, E.~L., and Chacoff, N.~P.
\newblock Rareness and specialization in plant--pollinator networks.
\newblock \emph{Ecology}, 92\penalty0 (1):\penalty0 19--25, 2011.

\bibitem[Duchi et~al.(2011)Duchi, Hazan, and Singer]{duchi2011adaptive}
Duchi, J., Hazan, E., and Singer, Y.
\newblock Adaptive subgradient methods for online learning and stochastic
  optimization.
\newblock \emph{Journal of Machine Learning Research}, 12\penalty0
  (Jul):\penalty0 2121--2159, 2011.

\bibitem[Durif et~al.(2019)Durif, Modolo, Mold, Lambert-Lacroix, and
  Picard]{durif2019probabilistic}
Durif, G., Modolo, L., Mold, J.~E., Lambert-Lacroix, S., and Picard, F.
\newblock Probabilistic count matrix factorization for single cell expression
  data analysis.
\newblock \emph{Bioinformatics}, 35\penalty0 (20):\penalty0 4011--4019, 2019.

\bibitem[Fan et~al.(2020)Fan, Ding, Yang, and Udell]{fan2020low}
Fan, J., Ding, L., Yang, C., and Udell, M.
\newblock Low-rank tensor recovery with euclidean-norm-induced schatten-p
  quasi-norm regularization.
\newblock \emph{arXiv preprint arXiv:2012.03436}, 2020.

\bibitem[Fu et~al.(2020{\natexlab{a}})Fu, Ibrahim, Wai, Gao, and
  Huang]{fu2020block}
Fu, X., Ibrahim, S., Wai, H.-T., Gao, C., and Huang, K.
\newblock Block-randomized stochastic proximal gradient for low-rank tensor
  factorization.
\newblock \emph{IEEE Transactions on Signal Processing}, 68:\penalty0
  2170--2185, 2020{\natexlab{a}}.

\bibitem[Fu et~al.(2020{\natexlab{b}})Fu, Vervliet, De~Lathauwer, Huang, and
  Gillis]{fu2020computing}
Fu, X., Vervliet, N., De~Lathauwer, L., Huang, K., and Gillis, N.
\newblock Computing large-scale matrix and tensor decomposition with structured
  factors: A unified nonconvex optimization perspective.
\newblock \emph{IEEE Signal Processing Magazine}, 37\penalty0 (5):\penalty0
  78--94, 2020{\natexlab{b}}.

\bibitem[Fu et~al.(2021)Fu, Seo, Clarke, and Hutchinson]{fu2019link}
Fu, X., Seo, E., Clarke, J., and Hutchinson, R.~A.
\newblock Link prediction under imperfect detection: Collaborative filtering
  for ecological networks.
\newblock \emph{IEEE Transactions on Knowledge and Data Engineering},
  33\penalty0 (8):\penalty0 3117--3128, 2021.

\bibitem[Gandy et~al.(2011)Gandy, Recht, and Yamada]{gandy2011tensor}
Gandy, S., Recht, B., and Yamada, I.
\newblock Tensor completion and low-n-rank tensor recovery via convex
  optimization.
\newblock \emph{Inverse Problems}, 27:\penalty0 025010, 2011.

\bibitem[Ghadermarzy et~al.(2018)Ghadermarzy, Plan, and
  Yilmaz]{ghadermarzy2018learning}
Ghadermarzy, N., Plan, Y., and Yilmaz, O.
\newblock Learning tensors from partial binary measurements.
\newblock \emph{IEEE Transactions on Signal Processing}, 67\penalty0
  (1):\penalty0 29--40, 2018.

\bibitem[Harshman(1970)]{harshman1970foundations1}
Harshman, R.
\newblock Foundations of the {PARAFAC} procedure: Models and conditions for an
  ``explanatory" multi-modal factor analysis.
\newblock \emph{UCLA Working Papers in Phonetics}, 16, 1970.

\bibitem[Hu et~al.(2008)Hu, Koren, and Volinsky]{Hu2008}
Hu, Y., Koren, Y., and Volinsky, C.
\newblock {Collaborative Filtering for Implicit Feedback Datasets}.
\newblock \emph{IEEE International Conference on Data Mining}, 2008.

\bibitem[Hutchinson et~al.(2011)Hutchinson, Liu, and
  Dietterich]{hutchinson2011incorporating}
Hutchinson, R., Liu, L.-P., and Dietterich, T.
\newblock Incorporating boosted regression trees into ecological latent
  variable models.
\newblock In \emph{Proceedings of the AAAI Conference on Artificial
  Intelligence}, volume~25, pp.\  1343--1348, 2011.

\bibitem[Jain et~al.(2005)Jain, Nandakumar, and Ross]{jain2005score}
Jain, A., Nandakumar, K., and Ross, A.
\newblock Score normalization in multimodal biometric systems.
\newblock \emph{Pattern Recognition}, 38\penalty0 (12):\penalty0 2270--2285,
  2005.

\bibitem[Joseph et~al.(2009)Joseph, Elkin, Martin, and
  Possingham]{joseph2009modeling}
Joseph, L.~N., Elkin, C., Martin, T.~G., and Possingham, H.~P.
\newblock Modeling abundance using n-mixture models: the importance of
  considering ecological mechanisms.
\newblock \emph{Ecological Applications}, 19\penalty0 (3):\penalty0 631--642,
  2009.

\bibitem[Kanatsoulis et~al.(2020)Kanatsoulis, Fu, Sidiropoulos, and
  Akcakaya]{kanatsoulis2019regular}
Kanatsoulis, C.~I., Fu, X., Sidiropoulos, N.~D., and Akcakaya, M.
\newblock Tensor completion from regular sub-nyquist samples.
\newblock \emph{IEEE Transactions on Signal Processing}, 68:\penalty0 1--16,
  2020.

\bibitem[Karatzoglou et~al.(2010)Karatzoglou, Amatriain, Baltrunas, and
  Oliver]{karatzoglou2010multiverse}
Karatzoglou, A., Amatriain, X., Baltrunas, L., and Oliver, N.
\newblock Multiverse recommendation: N-dimensional tensor factorization for
  context-aware collaborative filtering.
\newblock In \emph{Proceedings of the ACM Conference on Recommender Systems},
  pp.\  79–86, 2010.

\bibitem[Kashima et~al.(2009)Kashima, Kato, Yamanishi, Sugiyama, and
  Tsuda]{kashima2009link}
Kashima, H., Kato, T., Yamanishi, Y., Sugiyama, M., and Tsuda, K.
\newblock Link propagation: A fast semi-supervised learning algorithm for link
  prediction.
\newblock In \emph{Proceedings of the SIAM international conference on data
  mining}, pp.\  1100--1111, 2009.

\bibitem[Kim \& Choi(2007)Kim and Choi]{kim2007nonnegative}
Kim, Y.-D. and Choi, S.
\newblock Nonnegative tucker decomposition.
\newblock In \emph{Proceedings of the IEEE Conference on Computer Vision and
  Pattern Recognition}, pp.\  1--8, 2007.

\bibitem[Kingma \& Ba(2015)Kingma and Ba]{kingma2015adam}
Kingma, D.~P. and Ba, J.
\newblock Adam: A method for stochastic optimization.
\newblock \emph{CoRR}, abs/1412.6980, 2015.

\bibitem[Lee \& Wang(2020)Lee and Wang]{lee2020tensor}
Lee, C. and Wang, M.
\newblock Tensor denoising and completion based on ordinal observations.
\newblock In \emph{International Conference on Machine Learning}, pp.\
  5778--5788. PMLR, 2020.

\bibitem[Liang et~al.(2016)Liang, Charlin, McInerney, and
  Blei]{liang2016modeling}
Liang, D., Charlin, L., McInerney, J., and Blei, D.~M.
\newblock Modeling user exposure in recommendation.
\newblock In \emph{Proceedings of the International Conference on World Wide
  Web}, pp.\  951--961, 2016.

\bibitem[Lin \& Zhang(2019)Lin and Zhang]{lin2019generalization}
Lin, S. and Zhang, J.
\newblock Generalization bounds for convolutional neural networks.
\newblock \emph{arXiv preprint arXiv:1910.01487}, 2019.

\bibitem[Liu et~al.(2019)Liu, Li, Tsang, and Liu]{liu2019costco}
Liu, H., Li, Y., Tsang, M., and Liu, Y.
\newblock Costco: A neural tensor completion model for sparse tensors.
\newblock In \emph{Proceedings of the ACM SIGKDD International Conference on
  Knowledge Discovery and Data Mining}, pp.\  324–334, 2019.

\bibitem[Liu et~al.(2013)Liu, Musialski, Wonka, and Ye]{liu2012tensor}
Liu, J., Musialski, P., Wonka, P., and Ye, J.
\newblock Tensor completion for estimating missing values in visual data.
\newblock \emph{IEEE Transactions on Pattern Analysis and Machine
  Intelligence}, 35\penalty0 (1):\penalty0 208--220, 2013.

\bibitem[MacKenzie et~al.(2006)MacKenzie, Nichols, Royle, Pollock, Bailey, and
  Hines]{MacKenzie2006}
MacKenzie, D.~I., Nichols, J.~D., Royle, J.~A., Pollock, K.~H., Bailey, L.~L.,
  and Hines, J.~E.
\newblock \emph{{Occupancy estimation and modeling: Inferring patterns and
  dynamics of species occurrence}}.
\newblock Elsevier, San Diego, USA, 2006.

\bibitem[Montanari \& Sun(2016)Montanari and Sun]{montanari2016spectral}
Montanari, A. and Sun, N.
\newblock Spectral algorithms for tensor completion.
\newblock \emph{Communications on Pure and Applied Mathematics}, 71, 2016.

\bibitem[Mu et~al.(2014)Mu, Huang, Wright, and Goldfarb]{mu2014square}
Mu, C., Huang, B., Wright, J., and Goldfarb, D.
\newblock Square deal: Lower bounds and improved relaxations for tensor
  recovery.
\newblock In \emph{Proceedings of the International Conference on Machine
  Learning}, volume~32, pp.\  73--81, 2014.

\bibitem[Papalexakis et~al.(2017)Papalexakis, Faloutsos, and
  Sidiropoulos]{papalexakis2016tensor}
Papalexakis, E.~E., Faloutsos, C., and Sidiropoulos, N.~D.
\newblock Tensors for data mining and data fusion: Models, applications, and
  scalable algorithms.
\newblock \emph{ACM Transactions on Intelligent Systems and Technology},
  8\penalty0 (2):\penalty0 16, 2017.

\bibitem[Pu et~al.(2022)Pu, Ibrahim, Fu, and Hong]{pu2022stochastic}
Pu, W., Ibrahim, S., Fu, X., and Hong, M.
\newblock Stochastic mirror descent for low-rank tensor decomposition under
  non-euclidean losses.
\newblock \emph{IEEE Transactions on Signal Processing}, 70:\penalty0
  1803--1818, 2022.

\bibitem[Razaviyayn et~al.(2013)Razaviyayn, Hong, and
  Luo]{razaviyayn2013unified}
Razaviyayn, M., Hong, M., and Luo, Z.-Q.
\newblock A unified convergence analysis of block successive minimization
  methods for nonsmooth optimization.
\newblock \emph{SIAM Journal on Optimization}, 23\penalty0 (2):\penalty0
  1126--1153, 2013.

\bibitem[Royle(2004)]{royle2004n}
Royle, J.~A.
\newblock N-mixture models for estimating population size from spatially
  replicated counts.
\newblock \emph{Biometrics}, 60\penalty0 (1):\penalty0 108--115, 2004.

\bibitem[Schein et~al.(2015)Schein, Paisley, Blei, and
  Wallach]{schein2015bayesian}
Schein, A., Paisley, J., Blei, D.~M., and Wallach, H.
\newblock Bayesian poisson tensor factorization for inferring multilateral
  relations from sparse dyadic event counts.
\newblock In \emph{Proceedings of the ACM SIGKDD International Conference on
  Knowledge Discovery and Data Mining}, pp.\  1045--1054, 2015.

\bibitem[Schneider(2020)]{eric2020failing}
Schneider, E.~C.
\newblock Failing the test — the tragic data gap undermining the {U.S.}
  pandemic response.
\newblock \emph{New England Journal of Medicine}, 383\penalty0 (4):\penalty0
  299--302, 2020.

\bibitem[Seo \& Hutchinson(2018)Seo and Hutchinson]{seo2018predicting}
Seo, E. and Hutchinson, R.
\newblock Predicting links in plant-pollinator interaction networks using
  latent factor models with implicit feedback.
\newblock \emph{Proceedings of the AAAI Conference on Artificial Intelligence},
  32\penalty0 (1), Apr. 2018.

\bibitem[Seo et~al.(2022)Seo, Jones, Hutchinson, and Pfeiffer]{Jones2017}
Seo, E., Jones, J.~A., Hutchinson, R.~A., and Pfeiffer, V.~W.
\newblock Plant pollinator data at {HJ} {A}ndrews experimental forest, 2011 to
  2021, 2022.
\newblock URL
  \url{http://andlter.forestry.oregonstate.edu/data/abstract.aspx?dbcode=SA026.
  https://doi.org/10.6073/pasta/eec55cbc3dbfc56428629773737ab3e5}.

\bibitem[Serfling(1974)]{serfling1974prob}
Serfling, R.~J.
\newblock {Probability Inequalities for the Sum in Sampling without
  Replacement}.
\newblock \emph{The Annals of Statistics}, 2\penalty0 (1):\penalty0 39 -- 48,
  1974.

\bibitem[Shalev-Shwartz \& Ben-David(2014)Shalev-Shwartz and
  Ben-David]{shalev2014understanding}
Shalev-Shwartz, S. and Ben-David, S.
\newblock \emph{Understanding machine learning: From theory to algorithms}.
\newblock Cambridge university press, 2014.

\bibitem[Shashua \& Hazan(2005)Shashua and Hazan]{shashua2005nonnegative}
Shashua, A. and Hazan, T.
\newblock Non-negative tensor factorization with applications to statistics and
  computer vision.
\newblock In \emph{Proceedings of the International Conference on Machine
  Learning}, pp.\  792–799, 2005.

\bibitem[Simchowitz(2013)]{simchowitz2013zero}
Simchowitz, M.
\newblock Zero-inflated poisson factorization for recommendation systems.
\newblock \emph{Junior Independent Work (advised by D. Blei), Princeton
  University, Department of Mathematics}, 2013.

\bibitem[S{\o}rensen \& De~Lathauwer(2019)S{\o}rensen and
  De~Lathauwer]{sorensen2019fiber}
S{\o}rensen, M. and De~Lathauwer, L.
\newblock Fiber sampling approach to canonical polyadic decomposition and
  application to tensor completion.
\newblock \emph{SIAM Journal on Matrix Analysis and Applications}, 40\penalty0
  (3):\penalty0 888--917, 2019.

\bibitem[Tan et~al.(2016)Tan, Wu, Shen, Jin, and Ran]{tan2016shortterm}
Tan, H., Wu, Y., Shen, B., Jin, P.~J., and Ran, B.
\newblock Short-term traffic prediction based on dynamic tensor completion.
\newblock \emph{IEEE Transactions on Intelligent Transportation Systems},
  17\penalty0 (8):\penalty0 2123--2133, 2016.

\bibitem[Tillinghast et~al.(2020)Tillinghast, Fang, Zhang, and
  Zhe]{tillinghast2020prob}
Tillinghast, C., Fang, S., Zhang, K., and Zhe, S.
\newblock Probabilistic neural-kernel tensor decomposition.
\newblock In \emph{IEEE International Conference on Data Mining}, pp.\
  531--540, 2020.

\bibitem[Tylianakis et~al.(2010)Tylianakis, Lalibert{\'{e}}, Nielsen, and
  Bascompte]{Tylianakis2010}
Tylianakis, J.~M., Lalibert{\'{e}}, E., Nielsen, A., and Bascompte, J.
\newblock {Conservation of species interaction networks}.
\newblock \emph{Biological Conservation}, 143\penalty0 (10):\penalty0
  2270--2279, 2010.

\bibitem[Vershynin(2012)]{vershynin2012introduction}
Vershynin, R.
\newblock Introduction to the non-asymptotic analysis of random matrices.
\newblock In \emph{Compressed Sensing: Theory and Applications}, pp.\
  210–268. Cambridge University Press, 2012.

\bibitem[Wainwright(2019)]{wainwright2019high}
Wainwright, M.~J.
\newblock \emph{High-dimensional statistics: A non-asymptotic viewpoint},
  volume~48.
\newblock Cambridge University Press, 2019.

\bibitem[Wang et~al.(2018)Wang, Peng, Zhao, Leung, Zhao, and
  Meng]{wang2018hyperspectral}
Wang, Y., Peng, J., Zhao, Q., Leung, Y., Zhao, X.-L., and Meng, D.
\newblock Hyperspectral image restoration via total variation regularized
  low-rank tensor decomposition.
\newblock \emph{IEEE Journal of Selected Topics in Applied Earth Observations
  and Remote Sensing}, 11\penalty0 (4):\penalty0 1227--1243, 2018.

\bibitem[Yuan \& Zhang(2016)Yuan and Zhang]{yuan2016tensor}
Yuan, M. and Zhang, C.-H.
\newblock On tensor completion via nuclear norm minimization.
\newblock \emph{Foundations of Computational Mathematics}, 16\penalty0
  (4):\penalty0 1031--1068, 2016.

\bibitem[Zhang et~al.(2020{\natexlab{a}})Zhang, Fu, Wang, Zhao, and
  Hong]{zhang2020spectrum}
Zhang, G., Fu, X., Wang, J., Zhao, X.-L., and Hong, M.
\newblock Spectrum cartography via coupled block-term tensor decomposition.
\newblock \emph{IEEE Transactions on Signal Processing}, 68:\penalty0
  3660--3675, 2020{\natexlab{a}}.

\bibitem[Zhang et~al.(2020{\natexlab{b}})Zhang, Ghader, Pack, Xiong, Darzi,
  Yang, Sun, Kabiri, and Hu]{zhang2020interactive}
Zhang, L., Ghader, S., Pack, M.~L., Xiong, C., Darzi, A., Yang, M., Sun, Q.,
  Kabiri, A., and Hu, S.
\newblock An interactive {COVID}-19 mobility impact and social distancing
  analysis platform.
\newblock \emph{medRxiv}, 2020{\natexlab{b}}.

\bibitem[Zhang \& Aeron(2017)Zhang and Aeron]{zhang2017exact}
Zhang, Z. and Aeron, S.
\newblock Exact tensor completion using t-{SVD}.
\newblock \emph{IEEE Transactions on Signal Processing}, 65\penalty0
  (6):\penalty0 1511--1526, 2017.

\end{thebibliography}
\end{document}